\documentclass{article} 
\usepackage{iclr2025_conference,times}

\iclrfinalcopy

\usepackage{amsmath,amsfonts,bm}

\def\1{\bm{1}}

\DeclareMathAlphabet{\mathsfit}{\encodingdefault}{\sfdefault}{m}{sl}
\SetMathAlphabet{\mathsfit}{bold}{\encodingdefault}{\sfdefault}{bx}{n}

\newcommand{\E}{\mathbb{E}}

\newcommand{\R}{\mathbb{R}}

\DeclareMathOperator*{\argmin}{arg\,min}

\usepackage[utf8]{inputenc} 
\usepackage[T1]{fontenc}    
\usepackage{hyperref}       
\usepackage{url}            
\usepackage{booktabs}       
\usepackage{amsfonts} 
\usepackage{amsmath} 
\usepackage{nicefrac}       
\usepackage{microtype}      
\usepackage{xcolor}         
\usepackage{graphicx}
\usepackage{subcaption}
\usepackage{booktabs, makecell, tabularx}
\usepackage{multirow}
\usepackage{algorithm}
\usepackage{algpseudocode}
\usepackage[normalem]{ulem}
\usepackage{rotating}
\usepackage{array}
\usepackage{arydshln}
\title{Optimal Continuous Optimization for Structure Learning  }

\usepackage{amsthm}
\usepackage{amssymb}
\usepackage{cleveref}

\newcommand{\algorithmstyle}[1]{\sf{#1}}

\newcommand{\dagma}{{\algorithmstyle{DAGMA}}}
\newcommand{\golem}{{\algorithmstyle{GOLEM}}}
\newcommand{\notears}{{\algorithmstyle{NOTEARS}}}

\newcommand{\eqdef}{\stackrel{def}=}

\newcommand{\card}[1]{\left \vert #1 \right \vert }

\newtheorem{theorem}{Theorem}

\newtheorem{assumption}[theorem]{Assumption}

\newcolumntype{?}{!{\vrule width 1pt}}
\newcommand{\CommentNewline}[1]{\newline \vspace{0mm} \Comment {#1}}

\usepackage{hyperref}
\usepackage{url}

\usepackage{amsmath,enumerate}
\usepackage{comment}
\usepackage{cleveref}

\usepackage{graphicx}     
\usepackage{subcaption}

\usepackage{threeparttable}

\newcommand{\model}{\mathbf W}
\newcommand{\weights}{\mathbf L}
\newcommand{\modelrd}{\model^{(1/3)}}
\newcommand{\modeldprojected}{\model^{(2/3)}}
\newcommand{\pspace}{\mathbb R^{d\times d}}
\newcommand{\dspace}{\mathbb {D}}
\newcommand{\ospace}{\Pi}
\newcommand{\proj}{\psi}
\newcommand{\ordering}{\pi}
\newcommand{\optalg}[2]{\mathcal A_{#1}(#2)}

\newcommand{\algstyle}{\sf}

\newcommand{\frameworkours}{{$\proj{}${\algstyle DAG}}}
\newcommand{\algours}{{$\proj{}${\algstyle DAG}}}
\newcommand{\ours}{{\algours{}}}
\newcommand{\sgd}{{\algstyle SGD}}
\newcommand{\golemev}{{\algstyle GOLEM-EV}}

\newcommand{\thirdwidth}{0.325\textwidth}
\newcommand{\threesubfigwidth}{0.3\textwidth}

\newcommand{\Edist}[2]{\mathbb{E}_{#1}\left[#2\right] } 
\newcommand{\cD}{\mathcal D}

\newcommand{\ls}{\left(}
\newcommand{\rs}{\right)}
\newcommand{\lp}{\left[}
\newcommand{\rp}{\right]}
\newcommand{\lb}{\left\lbrace}
\newcommand{\rb}{\right\rbrace}

\title{$\psi$DAG: Projected Stochastic Approximation Iteration for DAG Structure Learning}

\author{%
Klea Ziu\textsuperscript{1}\thanks{Correspondence to: Klea Ziu <klea.ziu@mbzuai.ac.ae>, Dmitry Kamzolov <kamzolov.opt@gmail.com>}, Slavom\'ir Hanzely\textsuperscript{1}, Loka Li\textsuperscript{1},
  \\
  \textbf{Kun Zhang\textsuperscript{1,2}, Martin Tak\'a\v{c}\textsuperscript{1}, Dmitry Kamzolov\textsuperscript{1}}  \\
    \textbf{\textsuperscript{1}} Mohamed bin Zayed University of Artificial Intelligence, Abu Dhabi, UAE\\
    \textbf{\textsuperscript{2}} Carnegie Mellon University, Pittsburgh, USA
}

\begin{document}

\maketitle

\begin{abstract}
Learning the structure of Directed Acyclic Graphs (DAGs) presents a significant challenge due to the vast combinatorial search space of possible graphs, which scales exponentially with the number of nodes. Recent advancements have redefined this problem as a continuous optimization task by incorporating differentiable acyclicity constraints. These methods commonly rely on algebraic characterizations of DAGs, such as matrix exponentials, to enable the use of gradient-based optimization techniques. Despite these innovations, existing methods often face optimization difficulties due to the highly non-convex nature of DAG constraints and the per-iteration computational complexity. In this work, we present a novel framework for learning DAGs, employing a Stochastic Approximation approach integrated with Stochastic Gradient Descent (SGD)-based optimization techniques. Our framework introduces new projection methods tailored to efficiently enforce DAG constraints, ensuring that the algorithm converges to a feasible local minimum. With its low iteration complexity, the proposed method is well-suited for handling large-scale problems with improved computational efficiency. We demonstrate the effectiveness and scalability of our framework through comprehensive experimental evaluations, which confirm its superior performance across various settings.
\end{abstract}

\section{Introduction} \label{intro}

Learning graphical structures from data using Directed Acyclic Graphs (DAGs) is a fundamental challenge in machine learning \citep{koller2009probabilistic,peters2016causal,arjovsky2019invariant,sauer2021counterfactual}. This task has a wide range of practical applications across fields such as economics, genome research \citep{zhang2013integrated,stephens2009bayesian}, social sciences \citep{morgan2015counterfactuals}, biology \citep{PMID:15845847}, causal inference \citep{pearl2009causality, spirtes2000causation}. 
Learning the graphical structure is essential because the resulting models can often be given causal interpretations or transformed into representations with causal significance, such as Markov equivalence classes.
When graphical models cannot be interpreted causally \citep{pearl2009causality, spirtes2000causation}, they can still offer a compact and flexible representation for decomposing the joint distribution.

Structure learning methods are typically categorized into two approaches: score-based algorithms searching for a DAG minimizing a particular loss function and constraint-based algorithms relying on conditional independence tests. Constraint-based methods, such as the {\algstyle{PC}} algorithm \citep{spirtes1991algorithm} and {\algstyle{FCI}} \citep{Spirtes1995CausalII,colombo2012learning}, use conditional independence tests to recover the Markov equivalence class under the assumption of faithfulness. Other approaches, like those described in \citet{margaritis1999bayesian} and \citet{ tsamardinos2003algorithms}, employ local Markov boundary search. On the other hand, score-based methods frame the problem as an optimization of a specific scoring function, with typical choices including BGe \citep{kuipers2014addendum}, BIC \citep{maxwell1997efficient}, BDe(u) \citep{heckerman1995learning}, and MDL \citep{bouckaert1993probabilistic}.  Given the vast search space of potential graphs, many score-based methods employ local heuristics, such as Greedy Equivalence Search (GES) \citep{chickering2002optimal}, to efficiently navigate this complexity. Additionally, \citet{tsamardinos2006max}, \citet{gamez2011learning} propose hybrid methods combining elements of both constraint-based and score-based learning.

Recently, \citet{notears} introduced a smooth formulation for enforcing acyclicity, transforming the structure learning problem from its inherently discrete nature into a continuous, non-convex optimization task. This formulation allows for the use of gradient-based optimization techniques, enabling various extensions and adaptations to various domains, including nonlinear models \citep{yu2019dag,ng2022masked, kalainathan2022structural}, interventional datasets \citep{brouillard2020differentiable,faria2022differentiable}, unobserved confounders \citep{bhattacharya2021differentiable, bellot2021deconfounded}, incomplete datasets \citep{gao2022missdag,wang2020causal}, time series analysis \citep{sun2021nts,pamfil2020dynotears}, multi-task learning \citep{chen2021multi}, multi-domain settings \citep{zeng2021causal}, federated learning \citep{ng2022towards,gao2023feddag}, and representation learning \citep{yang2021causalvae}. 
With the growing interest in continuous structure learning methods \citep{vowels2022d}, a variety of theoretical and empirical studies have emerged. For instance, \citet{golem} investigated the optimality conditions and convergence properties of continuously constrained approaches such as \citet{notears}. In the bivariate case, \citet{deng2023global} demonstrated that a suitable optimization strategy converges to the global minimum of the least squares objective. Additionally, \citet{zhang2022truncated} and \citet{dagma} identified potential gradient vanishing issues with existing DAG constraints \citep{notears} and proposed adjustments to overcome these challenges.

\paragraph{Contributions.} In this work, we focus on the graphical models represented as Directed Acyclic Graphs (DAGs). Our main contributions can be summarized as follows:
\begin{enumerate}
    \item \textbf{Problem reformulation:} We introduce a new reformulation \eqref{eq:objective} of the discrete optimization problem for finding DAG as a stochastic optimization problem and we discuss its properties in detail in \Cref{sec:reformulation}.
    We demonstrate that the solution of this reformulated problem recovers the true DAG (\Cref{sec:reformulation}). 

    \item \textbf{Novel algorithm:} Leveraging insights from stochastic optimization, we present a new framework (\Cref{alg:fr}) for DAG learning (\Cref{sec:framework}) and present a simple yet effective algorithm \algours{} (\Cref{alg:ours}) within the framework. We proved that Algorithm \algours{}  converges to a local minimum of problem \eqref{eq:objective} at the rate
    $$\E \lp F(\model_T) - F(\model^{\ast})\rp \leq \mathcal O \ls \tfrac{\sigma_1 R}{\sqrt{T}} + \tfrac{L_1R^2}{T}\rs,$$
    where $T$ is a number of SGD-type steps, as detailed in (\Cref{sec:optimization_ordering}).
    
    \item \textbf{Experimental comparison:} In \Cref{sec:exp}, we demonstrate that the method \ours{} scales very well with graph size, handling up to $10000$ nodes. At that scale, the primary limitation is not computation complexity but the memory required to store the DAG itself.
    As a baseline, we compare \ours{} with established DAG learning methods, including \notears{} \citep{notears}, \golem{} \citep{golem} and \dagma{} \citep{dagma}.
    We show a significant improvement in scalability, as baseline methods struggle with larger graphs. Specifically, \notears{} \citep{notears}, \golem{} \citep{golem} and \dagma{} \citep{dagma} require more than $100$ hours for graphs with over $3000$ nodes, exceeding the allotted time.

\end{enumerate}

\section{Background}
\subsection{Graph Notation}
Before discussing the connection to the most relevant literature, we formalize the graph notation associated with DAGs.

Let \(\mathcal{G} \eqdef (V, E, w)\) represent a weighted directed graph, where \(V\) denotes the set of vertices with cardinality \(d \eqdef \card{V}\), \(E \in 2^{V \times V}\) is the set of edges, and \(w: V \times V \to \mathbb{R} \setminus \{0\}\) assigns weights to the edges. The \textit{adjacency matrix} \(\mathbf{A}(\mathcal{G}): \mathbb{R}^{d \times d}\) is defined such that \([\mathbf{A}(\mathcal{G})]_{ij} = 1\) if \((i,j) \in E\) and \(0\) otherwise. Similarly, the \textit{weighted adjacency matrix} \(\mathbf{W}(\mathcal{G})\) is defined by \([\mathbf{W}(\mathcal{G})]_{ij} = w(i,j)\) if \((i,j) \in E\) and \(0\) otherwise. 

When the weight function $w$ is binary, we simplify the notation to \(\mathcal{G} \eqdef (V, E)\). Similarly, when the graph \(\mathcal{G}\) is clear from context, we shorthand the notation to \(\mathbf{A} \eqdef \mathbf{A}(\mathcal{G})\) and \(\mathbf{W} \eqdef \mathbf{W}(\mathcal{G})\).

We denote the space of DAGs as $\mathbb{D}$. Since we will be utilizing topological sorting of DAGs\footnote{Topologial sorting of a graph \(\mathcal{G} \eqdef (V, E, w)\) refers to vertex ordering $V_1, V_2, \dots, V_d$ such that $E$ contains no edges of the form $V_i \to V_j,$ where $i\leq j$. Importantly, every DAG has at least one topological sorting.}, we also denote the space of vertex permutations $\ospace$.

\subsection{Linear DAG and SEM}

A Directed Acyclic Graph (DAG) model, defined on a set of $n$ random vectors $X\in \R^{n \times d}$, where \( X = (X_1, \dots, X_n) \) and $X_i \in \R^d$, consists of two components:
\begin{enumerate}
    \item A DAG \( \mathcal{G} = (V, E) \), which encodes a set of conditional independence relationships among the variables.

    \item The joint distribution \( P(X) \) with density \( p(x) \), which is Markov with respect to the DAG \( \mathcal{G} \) and factors as $p(x) = \prod_{i=1}^{d} p(x_i \mid x_{\text{PA}_{\mathcal{G}}(i)}),$ where \( \text{PA}_{\mathcal{G}}(i) = \{j \in V : X_j \to X_i \in E\} \) represents the set of parents of \( X_i \) in \( \mathcal{G} \). 
\end{enumerate}

This work focuses on the linear DAG model, which can be equivalently represented by a set of linear Structural Equation Models (SEMs). In matrix notation, the linear DAG model can be expressed as
\begin{equation}\label{eq:linear_SEM}
    X = X\mathbf{W} + N,
\end{equation}
where \( \mathbf{W} = [\mathbf{W}_1 | \cdots | \mathbf{W}_d] \) is a weighted adjacency matrix, and \( N = (N_1, \dots, N_n) \) is a matrix where each $N_i \in \R^d$ represents a noise vector with independent components. The structure of graph \( \mathcal{G} \) is determined by the non-zero coefficients in \( \mathbf{W} \); specifically \( X_j \to X_i \in E \) if and only if the corresponding coefficient in \( \mathbf{W}_i \) for \( X_j \) is non-zero. The classical objective function is based on the least squares loss applied to the linear DAG model, 
\begin{equation} \label{eq:quadratic_loss}
    l(\model; X) \eqdef \frac{1}{2n} \|X - X\model\|_F^2.
\end{equation}

\subsection{Most Related Literature} \label{sec:literature_related}
A body of research most connected to our work is based on the non-convex continuous framework of \citet{notears}, \citet{golem}, and \citet{dagma}, which we will discuss in detail.

\citet{notears} addressed the constrained optimization problem 
\begin{equation} \label{eq:lnotears}
\min_{\model \in \mathbb{R}^{d \times d}} 
\ell(\model; X)_{\notears{}} \eqdef
\frac{1}{2n} \|X - X\model\|_F^2 + \lambda \|\model\|_1 \quad \text{subject to} \quad h(\model) = 0, 
\end{equation}
where \( \ell(\model; X) \) represents the least squares objective and \( h(\model) := \text{tr}(e^{\model \odot \model}) - d \) enforces the DAG constraint. Additionally, an \( \ell_1 \) regularization term \( \lambda \|\model\|_1 \), where \( \| \cdot \|_1 \) is the element-wise \( \ell_1 \)-norm and \( \lambda \) is a hyperparameter incorporated into the objective function. This formulation addresses the linear case with equal noise variances, as discussed in \citet{loh2014high} and \citet{peters2014identifiability}. This constrained optimization problem is solved using the augmented Lagrangian method \citep{bertsekas1999nonlinear}, followed by thresholding the obtained edge weights. 
However, since this approach computes the acyclicity function via the matrix exponential, each iteration incurs a computational complexity of $\mathcal O(d^3)$, which significantly limits the scalability of the method.

\citet{golem} introduced the \golem{} method, which enhances the scoring function by incorporating an additional log-determinant term, \(\log|\det(I - \model)|\) to align with the Gaussian log-likelihood,

\begin{equation} \label{eq:lgolem}
\min_{\model \in \mathbb{R}^{d \times d}} \ell(\model; X)_{\golem{}} \eqdef \frac{d}{2} \log\|X - X\model\|_F^2 - \log|\det(I - \model)| + \lambda_1 \|\model\|_1 + \lambda_2 h(\model),
\end{equation}
where \(\lambda_1\) and \(\lambda_2\) serve as regularization hyperparameters within the objective function. Although the newly added log-determinant term is zero when the current model \(\model\) is a DAG, this score function does not provide an exact characterization of acyclicity. Specifically, the condition \(\log|\det(I - \model)| = 0\) does not imply that \(\model\) represents a DAG. 

\citet{dagma} introduces a novel acyclicity characterization for DAGs using a log-determinant function,  

\begin{equation} \label{eq:ldagma}
\min_{\model \in \mathbb{R}^{d \times d}}
\ell(\model; X)_{\dagma{}} \eqdef
\frac{1}{2n} \|X - X\model\|_F^2 + \lambda_1 \|\model\|_1 \quad \text{subject to} \quad h^s_{ldet}(\model) = 0,
\end{equation}
where 
\(
h^s_{ldet}(\model) \overset{\text{def}}{=} -\log\det(sI - \model \circ \model) + d \log s
\), and it is both exact and differentiable.

In practice, the augmented Lagrangian method enforces the hard DAG constraint by increasing the penalty coefficient towards infinity, which requires careful parameter fine-tuning and can lead to numerical difficulties and ill-conditioning \citep{birgin2005numerical, ng2022convergence}. As a result, existing methods face challenges across several aspects of optimization, including the careful selection of constraints, high computational complexity, and scalability issues.

To overcome these challenges, we propose a novel framework for enforcing the acyclicity constraint, utilizing a low-cost projection method. This approach significantly reduces iteration complexity and eliminates the need for expensive hyperparameter tuning.

\section{Stochastic Approximation for DAGs}
Our framework is built on a reformulation of the objective function as a stochastic optimization problem, aiming to minimize the stochastic function $F(w)$,
\begin{equation}
    \label{eq:stoh_min}
    \min_{w\in \R^d} \lb F(w) \eqdef \E_{\xi} \lp f(w,\xi) \rp \rb,
\end{equation}
where $\xi \in \Xi$ is a random variable drawn from the distribution $\Xi$. This formulation is common in stochastic optimization where computing the exact expectation is infeasible, but the values of  $f(w,\xi)$ and its stochastic gradients $g(w,\xi)$ can be computed. Linear and logistic regressions are classical examples of such problems. 

To address this problem, two main approaches exist: Stochastic Approximation (SA) and Sample Average Approximation (SAA). The SAA approach involves sampling a fixed number $n$ of random variables or data points $\xi_i$ and then minimizing their average $\tilde{F}(w)$:
\begin{equation}
    \label{eq:SAA}
     \min_{w\in \R^d} \lb \tilde{F}(w) \eqdef \tfrac{1}{n} \sum_{i=1}^n f(w,\xi_i)\rb.
\end{equation}
Now, the problem \eqref{eq:SAA} becomes deterministic and can be solved using various optimization methods, such as gradient descent. However, the main drawback of this approach is that the solution of \eqref{eq:SAA} $\tilde{w}^{\ast}$ is not necessarily equal to the solution of the original problem \eqref{eq:stoh_min}. Even with a perfect solution of \eqref{eq:SAA}, there will still be a gap $\|\tilde{w}^{\ast}-w^{\ast}\|=\delta_x$ and $F(\tilde{w}^{\ast}) - F^{\ast} = \delta_F $ between approximate and true solution. These gaps are dependent on the sample size $n$.
\\
Stochastic Approximation (SA) minimizes the true function $F(w)$ by utilizing the stochastic gradient $g(w,\xi)$. Below, we provide the formal definition of a stochastic gradient.
\begin{assumption} \label{as:1ord_stoch}
        For all $w \in \R^d$, we assume that stochastic gradients $g(w, \xi)\in \R^d$ satisfy
        \begin{equation}
        \label{eq:stoch_grad_def}
                \mathbb{E}[g(w, \xi) \mid w] = \nabla F(w), \quad
                \mathbb{E}\left[\|g(w, \xi)-  \nabla F(w)\|^2 \mid w\right] \leq \sigma_1^2.
        \end{equation}
\end{assumption}
We use these stochastic gradients in \sgd{}-type methods:
\begin{equation} \label{eq:sgd}
    w_{t+1} = w_t - h_t g(w_t,\xi_i), 
\end{equation}
where $h_t$ is a step-size schedule.
SA originated with the pioneering paper by \cite{robbins1951stochastic}. For convex and $L$-smooth function $F(w)$, \citet{polyak1990new, polyak1992acceleration, nemirovski2009robust, nemirovski1983problem} developed significant improvements to SA method in the form of longer step-sizes with iterate averaging, and obtained the convergence guarantee 
$$\E \lp F(w_T) - F(x^{\ast})\rp \leq \mathcal O \ls \tfrac{\sigma_1 R}{\sqrt{T}} + \tfrac{L_1R^2}{T}\rs.$$ 
\citet{lan2012optimal} developed an optimal method with a guaranteed convergence rate $\mathcal O \ls \tfrac{\sigma_1 R}{\sqrt{T}} + \tfrac{L_1R^2}{T^2}\rs$, matching the worst-case lower bounds. The key advantage of SA is that it provides convergence guarantees for the original problem \eqref{eq:stoh_min}. Additionally, methods effective for the SA approach tend to perform well for the SAA approach as well. 

\subsection{Stochastic Reformulation} \label{sec:reformulation}
Using the perspective of Stochastic Approximation, we can rewrite the linear DAG \eqref{eq:linear_SEM} as
\begin{equation} \label{eq:x}
    x = X_i = \lp I - \model_{\ast}^{\top}\rp^{-1}N_i,
\end{equation}
where $\model^{\ast}$ is a true DAG that corresponds to the full distribution, and our goal is to find DAG $\model$ that is close to $\model^{\ast}$. If we assume that $x = X_i$ is a random vector sampled from a distribution $\cD$, we can express the objective function as an expectation,
\begin{equation} \label{eq:objective}
    \min_{\model \in \dspace} \Edist{x\sim \cD}{l(\model; x) \eqdef \tfrac 12 \Vert x -\model^{\top} x \Vert^2=\tfrac 12 \Vert x^{\top} - x^{\top} \model\Vert^2}.
\end{equation}
For $x$ from \eqref{eq:x} we can calculate  $\| x -\model^{\top} x \|=\| (I -\model^{\top}) x \|=\| (I -\model)\lp I - \model_{\ast}^{\top}\rp^{-1}N_i\|$, which implies that the minimizer of \eqref{eq:objective} recovers the true DAG. Conversely, this is not the case for methods such as \citet{notears}, \citet{golem}, and \citet {dagma}, which are based on SAA approaches with losses \eqref{eq:quadratic_loss}, \eqref{eq:lnotears}, \eqref{eq:lgolem}, \eqref{eq:ldagma}. 

\section{Scalable Framework} \label{sec:framework}

Instead of strictly enforcing DAG constraints throughout the entire iteration process, we propose a novel, scalable optimization framework that consists of three main steps:
\begin{enumerate}
    \item Running an optimization algorithm $\mathcal A_1$ without any DAG constraints, $\mathcal A_1: \pspace \to \pspace.$
    \item Finding a DAG that is ``closest'' to the current iterate using a projection $\proj: \pspace \to (\dspace, \ospace)$, which also returns its  topological sorting.
    \item Running the optimization algorithm $\mathcal A_2$ while preserving the vertex order, $\mathcal A_2: (\dspace; \ospace) \to \dspace.$
\end{enumerate}

\begin{algorithm}
    \caption{\frameworkours{} framework} \label{alg:fr}
    \begin{algorithmic}[1]
        \State \textbf{Requires:} Initial model $\model_0 \in \pspace$.
        \For {$k=0,1,2\dots, K-1$}
            \State $\modelrd_k = \optalg 1 {\model_k}$ \Comment{$\modelrd_k \in \pspace$}.
            \State $(\modeldprojected_k, \ordering_k) = \proj(\modelrd_k)$ \Comment{$\modeldprojected_k \in \dspace$}
            \State $\model_{k+1} = \optalg 2 {\modeldprojected_k; \ordering_k}$ \Comment{$\model_{k+1} \in \dspace \subset \pspace$}.
        \EndFor
        \State \textbf{Output:} $\model_K$. 
    \end{algorithmic}
\end{algorithm}

\subsection{Optimization for the fixed vertex ordering}
\label{sec:optimization_ordering}
Let us clarify how to optimize while preserving the vertex order in step 3 of the framework. 
Given a DAG $\mathcal{G}$, we can construct its topological ordering, denoted as $ord(\mathcal{G})$. In this ordering, for every edge, the start vertex appears earlier in the sequence than the end vertex. In general, this ordering is not unique. In the space of DAGs with $d$ vertices $\dspace$, there are $d!$ possible topological orderings. 

Once we have a topological ordering of the DAG, we can construct a larger DAG, $\hat{\mathcal{G}}$, by performing the transitive closure of $\mathcal{G}$. This new DAG $\hat{\mathcal{G}}$ contains all the edges of the original DAG, and additionally, it includes an edge between vertices $V_i$ and $V_j$ if there exists the path from $V_i$ to $V_j$ in $\mathcal{G}$. Thus, $\hat{\mathcal{G}}$ is an expanded version of $\mathcal{G}$. 

Now, the question arises: is it possible to construct an even larger DAG that contains both $\mathcal{G}$ and $\hat{\mathcal{G}}$? The answer is yes! We call this graph the Full DAG, denoted by $\tilde{\mathcal{G}}$, which is constructed via full transitive closure. In $\tilde{\mathcal{G}}$, there is an edge from vertex $V_i$ to vertex $V_j$ if $i < j$ in topological ordering $ord(\mathcal{G})$. This makes $\tilde{\mathcal{G}}$ the maximal DAG that includes $\mathcal{G}$. Note that for every topological sort, there is a corresponding full DAG. So, there are a total of $d!$ different full DAGs in the space of DAGs with $d$ vertices $\dspace$. 

We are now ready to discuss the optimization part. Let us formulate the following optimization problem
\begin{equation}
\label{eq:problem_masked}
    \min_{\model\in \R^{d\times d}} \Edist{x\sim \cD}{l(\model \cdot \mathbf{A}; x) = \tfrac 12 \Vert x -(\model\cdot \mathbf{A})^{\top} x \Vert^2},
\end{equation}
where ($\cdot$) denotes elementwise matrix multiplication. In this formulation, $\mathbf{A}$ acts as a mask, specifying coordinates that do not require gradient computation. The problem \eqref{eq:problem_masked} is a quadratic convex stochastic optimization problem, which can be efficiently solved using stochastic gradient descent (SGD)-type methods. These methods guarantee convergence to the global minimum, with a rate of $ \mathcal O \ls \tfrac{\sigma_1 R}{\sqrt{T}} + \tfrac{L_1R^2}{T}\rs.$

\begin{figure}
    \centering
    \includegraphics[width=0.45\textwidth]{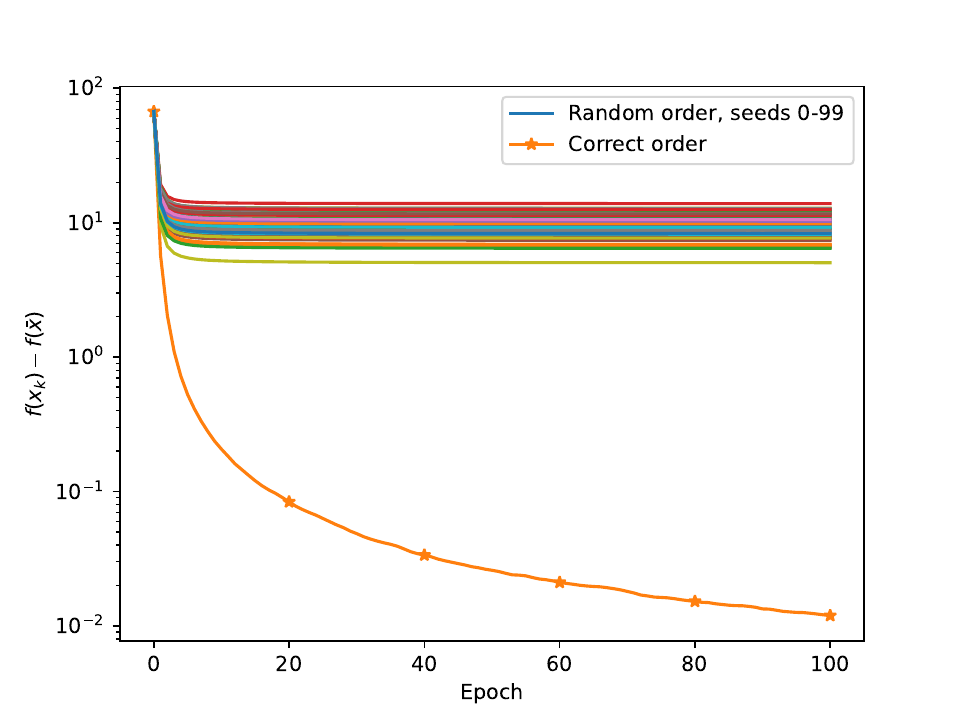}
    \includegraphics[width=0.45\textwidth]{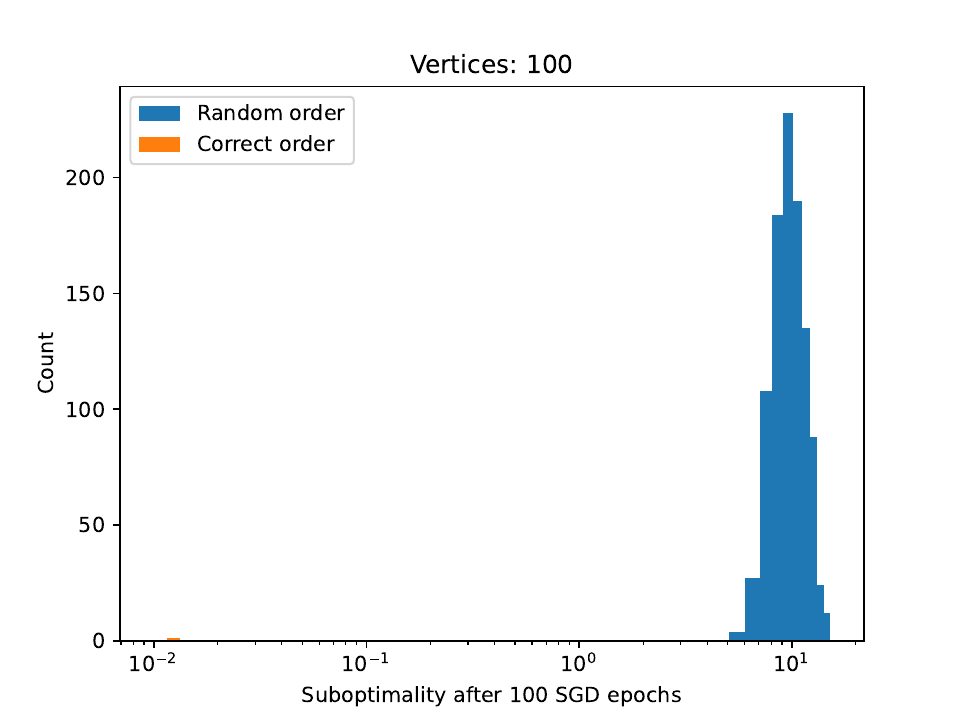}
    \caption{Minimization of \eqref{eq:objective} using SGD over a fixed topological ordering of vertices on graph type ER4 with $d=100$ vertices with Gaussian noise.
    Plots demonstrate that minimizing \eqref{eq:objective} over a fixed random vertex ordering does not approach the true solution of \eqref{eq:objective}.}
    \label{fig:objective_random}
\end{figure}

Assume that $\mathcal{G}^{\ast}$ is the true DAG with a weighted adjacency matrix $\mathbf{W}^{\ast}$, which is the solution we aim to find. Next, we can have the true ordering $ord(\mathcal{G}^{\ast})$ and the true full DAG $\tilde{\mathcal{G}}^{\ast}$ with its adjacency matrix $\mathbf{A}(\tilde{\mathcal{G}}^{\ast})$. The optimization problem \eqref{eq:objective}, with the solution $\mathbf{W}^{\ast}$, can be addressed by solving the optimization problem \eqref{eq:problem_masked} with $\mathbf{A} = \mathbf{A}(\mathcal{G}^{\ast})$. This result indicates that, if we know the true topological ordering $ord(\mathcal{G}^{\ast})$, then we can recover the true DAG $\mathbf{W}^{\ast}$ with high accuracy. From a discrete optimization perspective, this approach significantly reduces the space of constraints from $2^{d^2-d}$ to $d!$. 
To illustrate the specificity of the minimizer of the proposed problem, \Cref{fig:objective_random} demonstrates that minimizing \eqref{eq:objective} over a fixed random vertex ordering does not approach the true solution of \eqref{eq:objective}. "Correct order" curve demonstrates the convergence of \eqref{eq:problem_masked} when the true ordering $ord(\mathcal{G}^{\ast})$ is known.

In Algorithm \eqref{alg:fr}, in step 4, we find the closest ordering to the current point, and next in step 5 solve \eqref{eq:problem_masked} for this ordering and its full DAG. The solution obtained for the given ordering is an approximate local minimum for the original problem \eqref{eq:objective}, which leads us to the next theorem

\begin{theorem}
    For an $L_1$-smooth function $F(\model) = \E_{x\sim \cD} \lp l(\model;x)\rp $, Algorithm \eqref{alg:fr}, with access to $\sigma_1$-stochastic gradients, converges to a local minimum of problem \eqref{eq:objective} at the rate
    $$\E \lp F(\model_T) - F(\model^{\ast})\rp \leq \mathcal O \ls \tfrac{\sigma_1 R}{\sqrt{T}} + \tfrac{L_1R^2}{T}\rs,$$
    where $T$ is a number of SGD-type steps.
\end{theorem}

Note that for a fixed vertex ordering and fixed adjacency matrix $\mathbf{A}$, the objective \eqref{eq:problem_masked} becomes separable, enabling parallel computation for large-scale problems. In this work, we solved the minimization problem \eqref{eq:problem_masked} for the number of nodes up to $d=10^4$, at which point the limiting factor was the memory to store $\model \in \pspace$. Through parallelization and efficient memory management, it is possible to solve even larger problems.

\subsection{Method} \label{sec:method} 
We now introduce the method \algours{}, which implements the framework outlined in \Cref{alg:fr}.

For simplicity, we select algorithm $\mathcal A_1$ as $\tau_1$ steps of Stochastic Gradient Descent (\sgd{}). Similarly, $\mathcal A_2$ consists of $\tau_2$ steps \sgd{}, where gradients are projected onto the space spanned by DAG's topological sorting, thus preserving the vertex order. It is important to reiterate that \sgd{} is guaranteed to converge to the neighborhood of the solution. In the implementation, we employed an advanced version of SGD, Universal Stochastic Gradient Method from      \citep{rodomanov2024universal}.

The implementation of the projection method is simple as well. We compute a ``closest'' topological sorting and remove all edges not permitted by this ordering.
The topological sorting is computed by a heuristic that calculates norms of all rows and columns to find the lowest value $v_i$. The corresponding vertex $i$ is then assigned to the ordering based on the following rule:
\begin{itemize}
    \item If $v_i$ was the column norm, $i$ is assigned to the beginning of the ordering.
    \item If $v_i$ was the row norm, $i$ is assigned to the end of the ordering. 
\end{itemize}
This step reduces the number of vertices, and the remaining vertices are topologically sorted through by a recursive call. We formalize this procedure in \Cref{alg:proj}. Note that this procedure can be efficiently implemented without recursion and with the computation cost $\mathcal O(d^2)$. 

\begin{algorithm}[h]
    \caption{Projection $\proj(\model)$ computing the ``closest'' vertex ordering (recursive form)} \label{alg:proj}
    \begin{algorithmic}[1]
        \State \textbf{Requires:} Model $\model \in \pspace$, (optional) weights $\weights \in \pspace$ with default value $\weights=\mathbf 1 \mathbf 1^\top$.
        \For {$k=1, \dots, d$}
            \State Set $r_k = \Vert \left(\model \circ \weights \right) [k][:] \Vert^2$
            \State Set $c_k = \Vert \left(\model \circ \weights \right) [:][k] \Vert^2$
        \EndFor
        
        \State Set $i_c = \argmin_{k\in \{1, \dots, d\}} c_k$
        \State Set $i_r = \argmin_{k\in \{1, \dots, d\}} r_k$
        \If {$r_{i_r}<= c_{i_c}$}
            \State \textbf{Output:} $[\proj(\model(i_c, i_c), \weights(i_c, i_c)), i_r]$
        \Else
            \State \textbf{Output:} $[i_c, \proj(\model(i_c, i_c), \weights(i_c, i_c))]$
        \EndIf
            \CommentNewline{By $A(i,j)$ we denote the submatrix $A[1, \dots, i-1, i+1, \dots, d][1, \dots, j-1, j+1, \dots, d]$}
    \end{algorithmic}
\end{algorithm}

\begin{algorithm} 
    \caption{\algours{}} \label{alg:ours}
    \begin{algorithmic}[1]
        \State \textbf{Requires:} initial model $\model_0 \in \pspace$, numbers or iterations $\tau_1, \tau_2$.
        \For {$k=0,1,2\dots, K-1$}
            \State $\modelrd_k =\sgd{}(\model_k)$  \Comment{$\tau_1$ iterations over $\pspace$.}
            \State $(\modeldprojected_k, \ordering_k) = \text{\Cref{alg:proj} }(\modelrd_k)$ 
            \State $\model_{k+1} =\sgd{}_{\ordering_k}(\model_k)$  \Comment{$\tau_2$ iterations preserving ordering $\ordering_k$.}
        \EndFor
        \State \textbf{Output:} $\model_K$
    \end{algorithmic}
\end{algorithm}

\section{Experiments} \label{sec:exp}
We experimentally compare our newly proposed algorithm \algours{}\footnote{Code implementing the proposed algorithm is available at \url{https://github.com/kz29/-psiDAG}. We use the Universal Stochastic Gradient Method from      \citep{rodomanov2024universal} as the inner optimizer. The code is available in the OPTAMI package \url{https://github.com/OPTAMI/OPTAMI} \citep{kamzolov2024optami}.} to other score-based methods for computing linear DAGs, \notears{} \citep{notears}, \golem{}\footnote{In all experiments, we consider \golemev, where the noise variances are equal.} \citep{golem} and \dagma{} \citep{dagma}.
As it is established that \dagma{} \cite{dagma} is an improvement over \notears{} \cite{notears}, we use mostly the former one in our experiments. 
As the baseline algorithms were implemented without extensive hyperparameter tuning, we avoided hyperparameter tuning as much as possible. 
In particular, we apply the same threshold as the one in \citet{notears}, \citet{golem}, \citet{dagma} across all scenarios.

\begin{figure}
    \centering
   \begin{subfigure}{\textwidth}
        \centering
        \includegraphics[width=\thirdwidth]{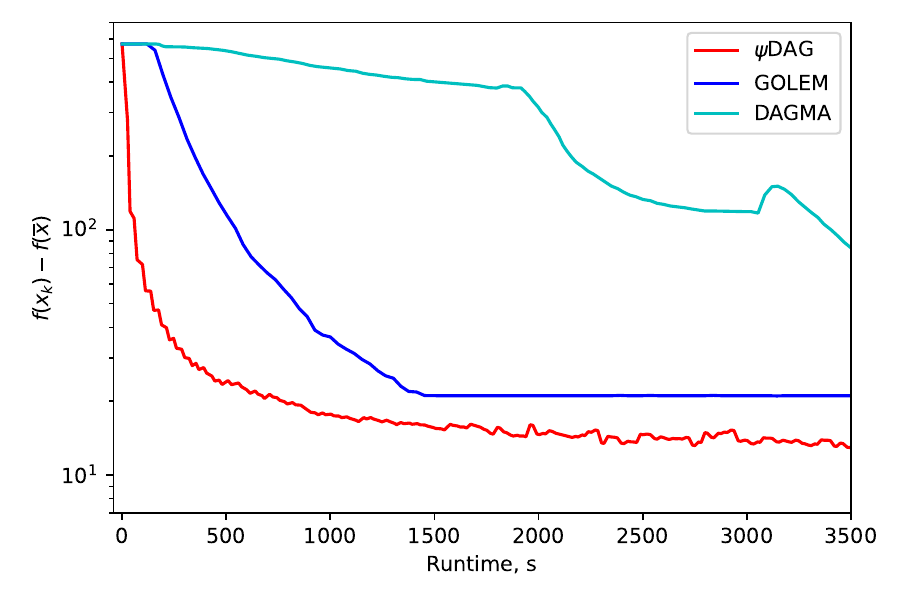}
        \includegraphics[width=\thirdwidth]{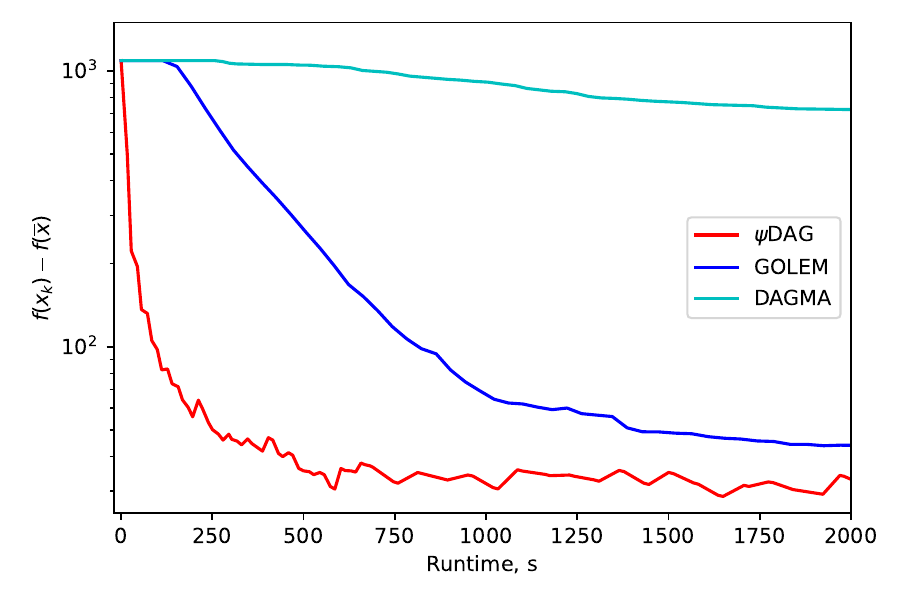}
        \includegraphics[width=\thirdwidth]{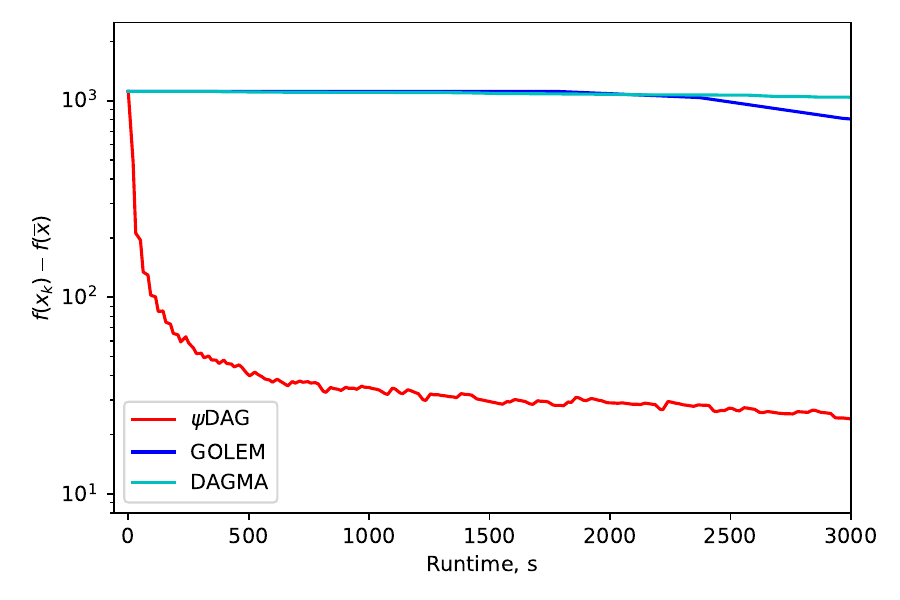}
        \includegraphics[width=\thirdwidth]{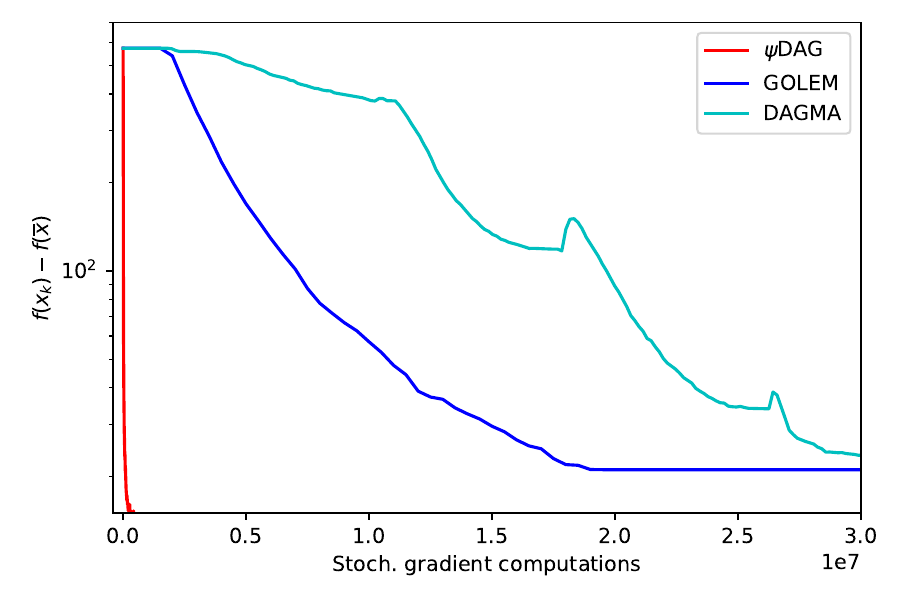}
        \includegraphics[width=\thirdwidth]{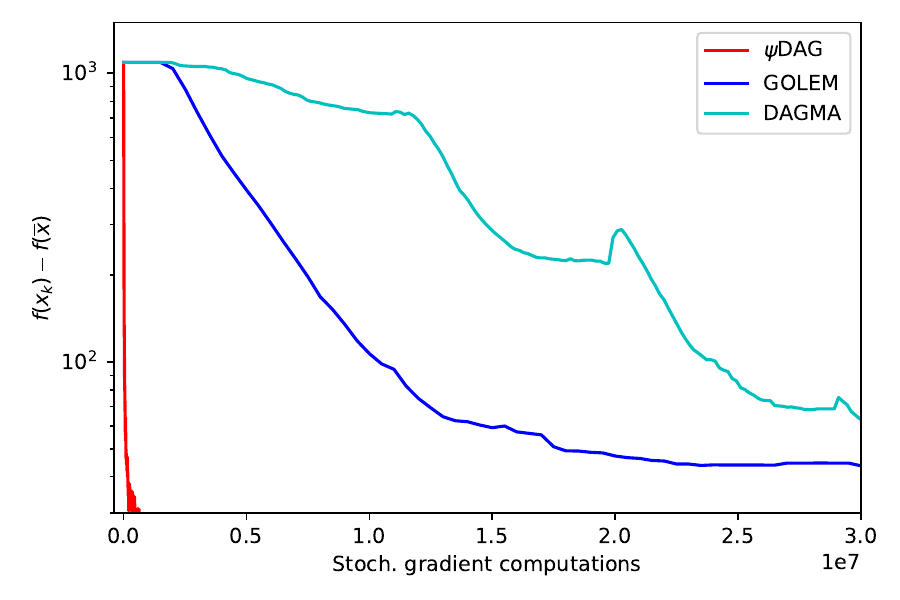}
        \includegraphics[width=\thirdwidth]{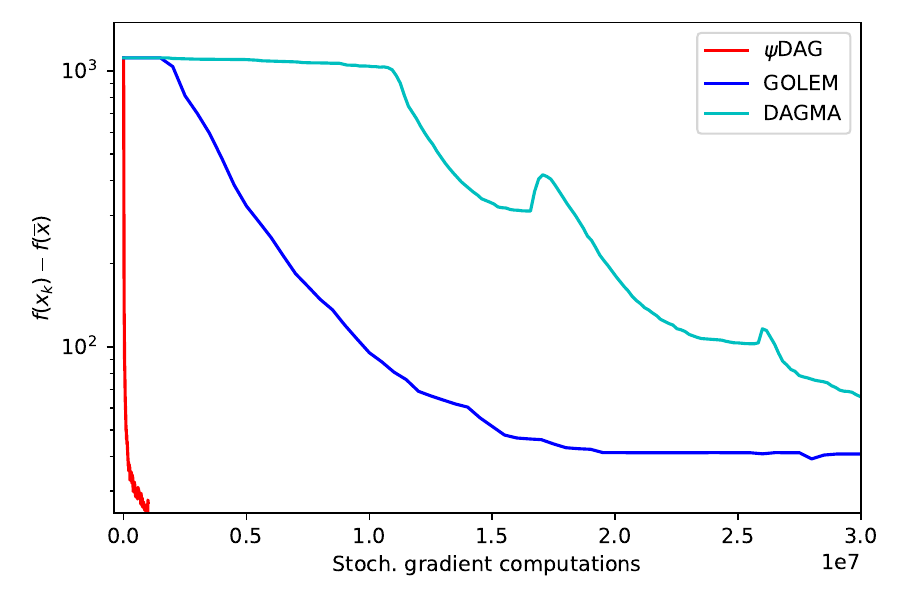}
    \end{subfigure}
    \caption{Linear SEM methods of \ours{}, \golem{} and \dagma{} on graphs of type ER4 with $d =1000$ number of nodes and with different noise distributions: Gaussian (first), exponential (second), and Gumbel (third).}
    \label{fig:er4_main}
\end{figure}

\Cref{fig:er4_main} shows that \ours{} consistently exhibits faster convergence across different noise distributions. 
\Cref{exp_fig} extends this result across different graph sizes and graph types.

\subsection{Synthetic Data Generation}
We generate ground truth DAGs to have \(d\) nodes and an average of \( k \times d \) edges, where \( k \in \{2, 4, 6\} \) is a sparsity parameter.
The graph structure is determined by the choice of the graph models to be either Erdős-Rényi (ER) or Scale Free (SF), and together with sparsity parameter $k$, we refer to them as \(\text{ER}k\) or \(\text{SF}k\). 
Each of the edges has assigned a random weight uniformly sampled from the interval \([-1, -0.05] \cup [0.05, 1]\). 

Following the linear Structural Equation Model (SEM), the observed data \(X\) has form \( X = N(I - W)^{-1} \), where \(N \in \mathbb{R}^{n \times d}\) represents \(n\) $d$-dimensional independent and identically distributed (i.i.d.) noise samples drawn from either Gaussian, exponential or Gumbel distributions.
In this study, we focus on an equal variance (EV) noise setting, with a scale factor of 1.0 applied to all variables. Unless otherwise specified, we generate the same number of samples \(n \in \{5000, 10000\}\) for training and validation datasets, respectively. 
A more detailed description can be found in Appendix \ref{app_exp}.

\subsection{Scalability Comparison}
\begin{figure}[ht]
  \centering
  \begin{subfigure}{\textwidth}
      \centering
      \includegraphics[width=\thirdwidth]{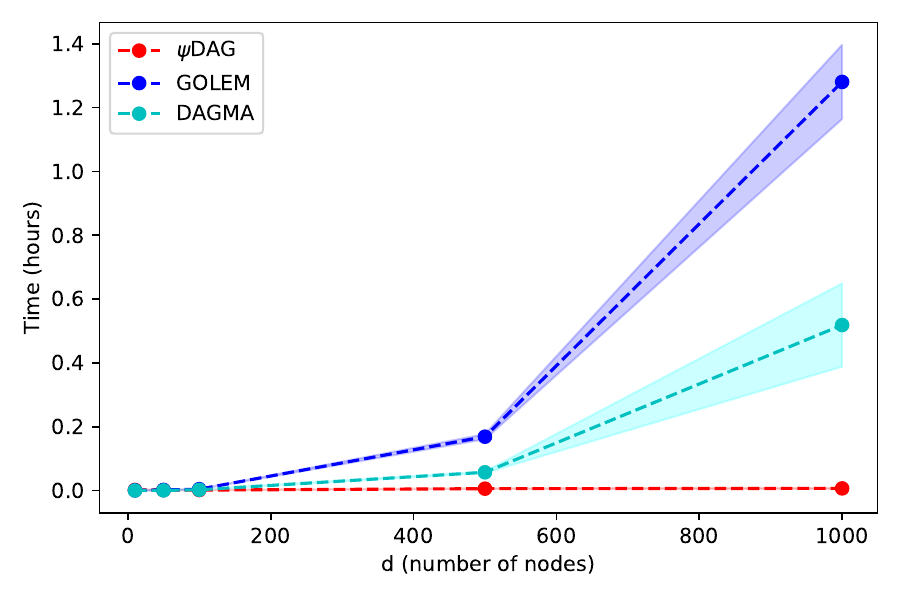}
      \hfill
      \includegraphics[width=\thirdwidth]{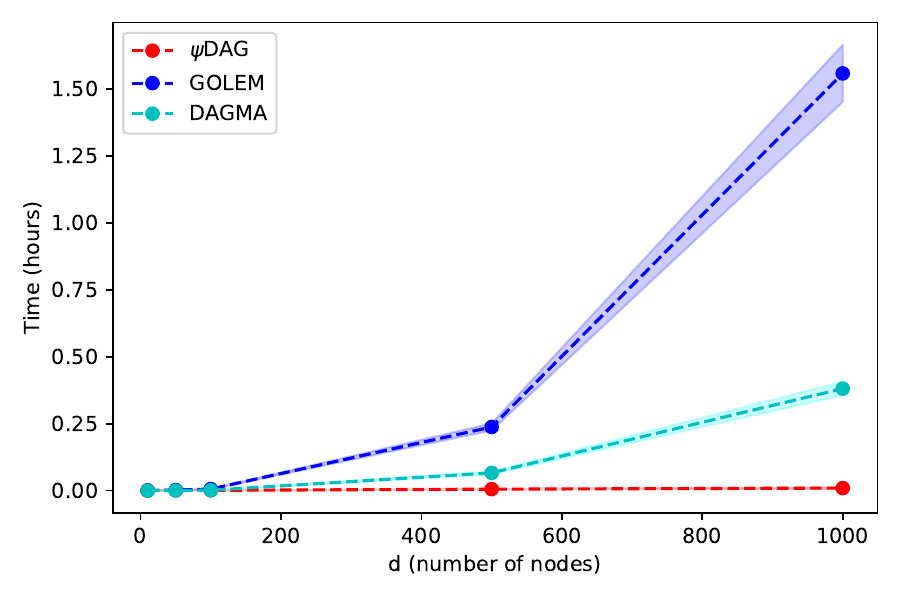}
      \hfill
      \includegraphics[width=\thirdwidth]{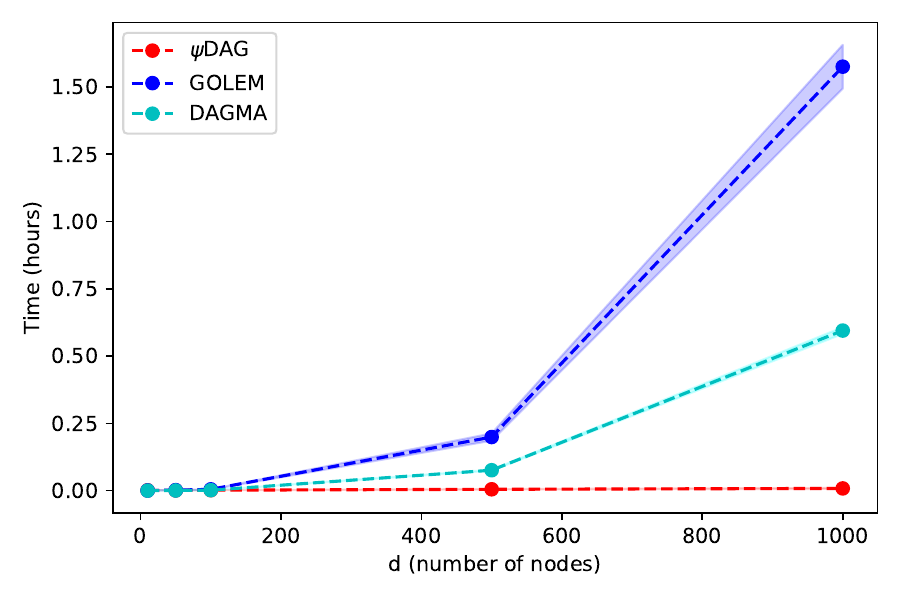}
      
  \caption{ER2 graph type} 
  \label{fig:gumb_exp_er2}
  \end{subfigure}
  \begin{subfigure}{\textwidth}
      \centering
      \includegraphics[width=\thirdwidth]{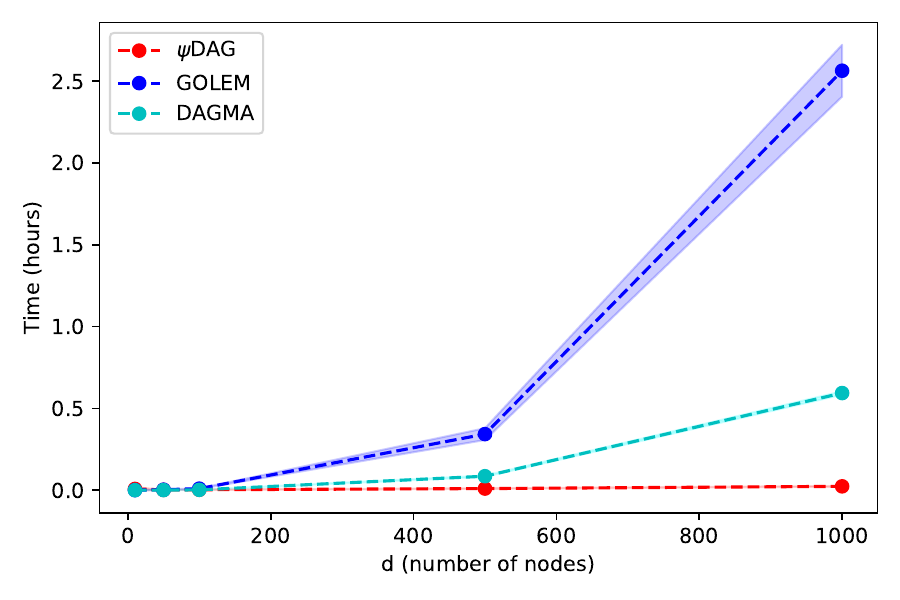}
      \hfill
        \includegraphics[width=\thirdwidth]{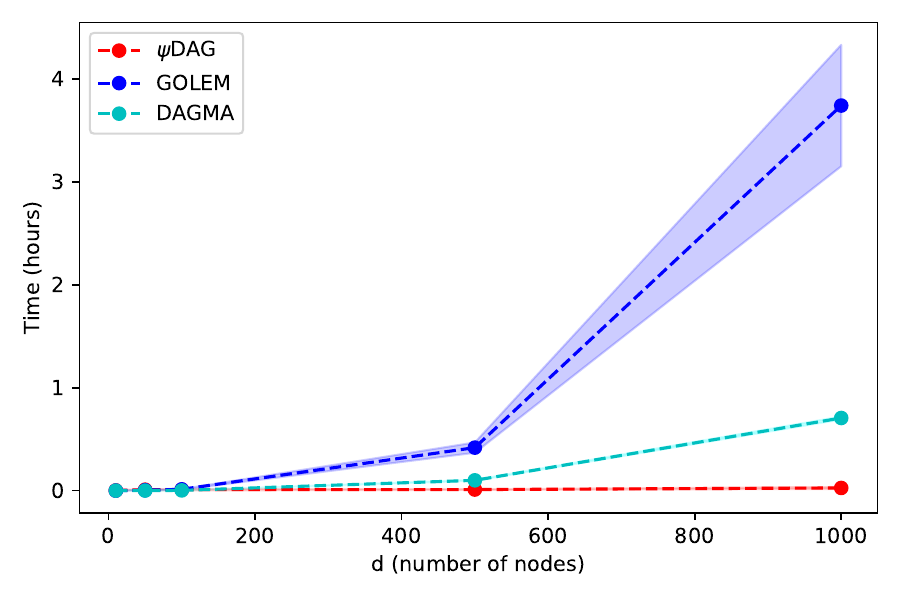}
      \hfill
            \includegraphics[width=\thirdwidth]{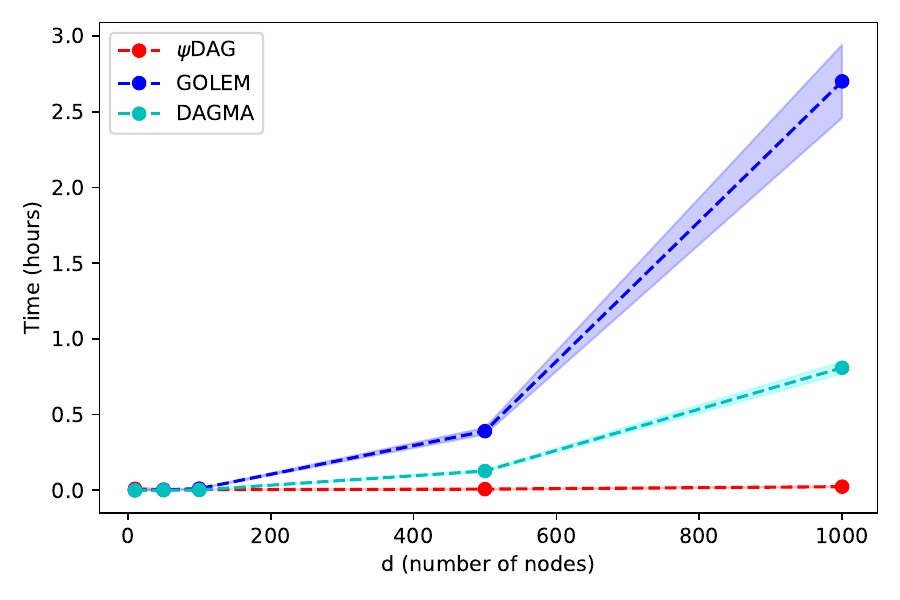}
  \caption{ER4 graph type} 
  \label{fig:gumb_exp_er4}
  \end{subfigure}
  \caption{Runtime (hours) of \ours{}, \golem{} and \dagma{} for ER2 and ER4 graph types with small number of nodes \(d = \{10,50,100, 500, 1000\}\). Noise distributions vary in different columns: Gaussian (first), exponential (second), and Gumbel (third).
   Method \ours{} showcases much better scalability when the number of nodes increases.}
   \label{fig:small}
\end{figure}

\begin{figure}
  \centering
  \begin{subfigure}{0.32\textwidth}
    \centering
    \includegraphics[width=\textwidth]{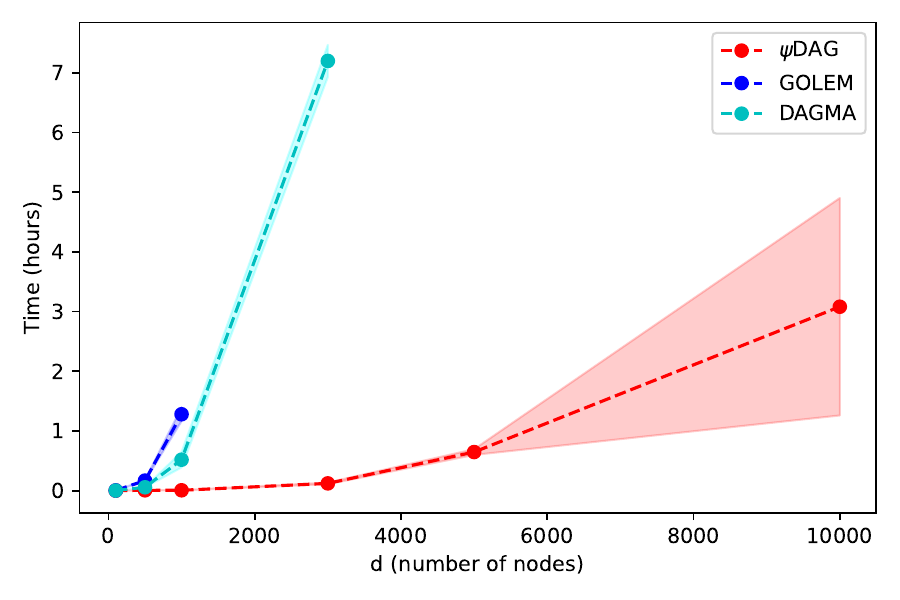}
    \caption{ER2}
    \label{sfig_er2}
  \end{subfigure}
  \hfill 
  \begin{subfigure}{0.32\textwidth}  
    \centering
    \includegraphics[width=\textwidth]{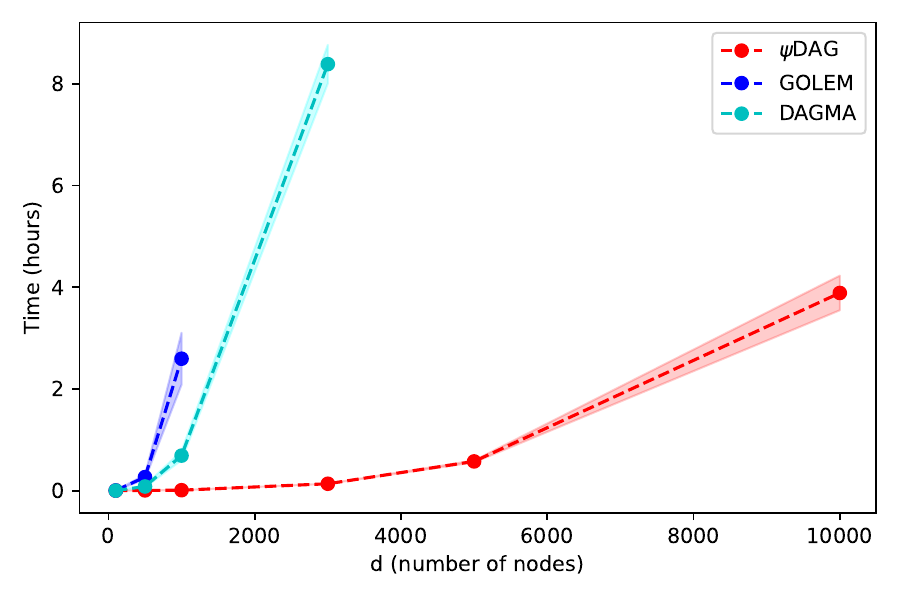}
    \caption{SF2}
    \label{sfig_sf2}
  \end{subfigure}
  \hfill
  \begin{subfigure}{0.32\textwidth}
    \centering
    \includegraphics[width=\textwidth]{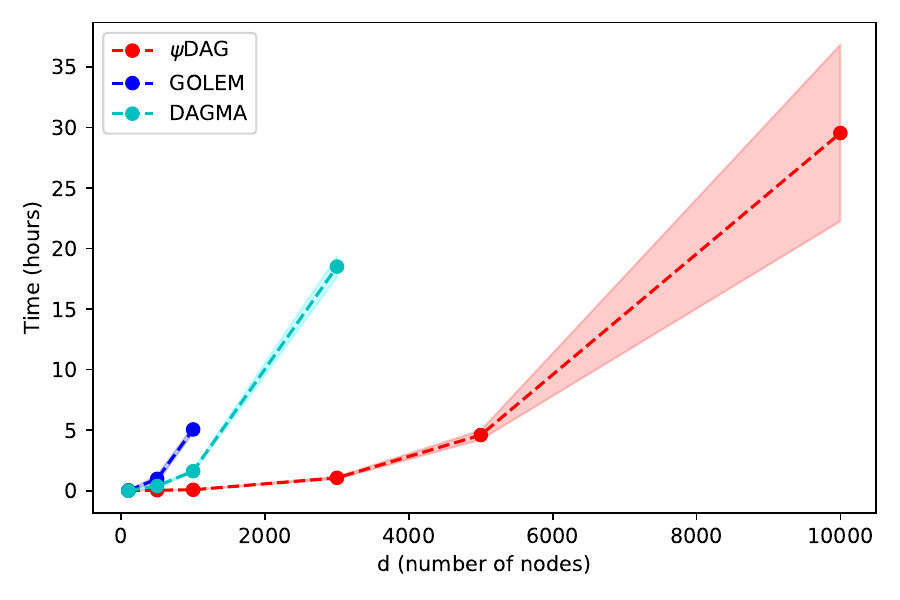}
    \caption{ER6}
    \label{sfig_er6}
  \end{subfigure}

  \caption{
    Runtime (hours) of \algours{}, \golem{}, and \dagma{} for different graph types as the graph size increases. The noise distribution is always Gaussian. 
    \Cref{sfig_er2} extends \Cref{fig:gumb_exp_er2} to a large number of nodes $d\in \{ 3000, 5000, 10000\}$,  \Cref{sfig_sf2} presents graph type SF2 and \Cref{sfig_er6} showcases a more dense graph structure.
    Method \ours{} demonstrates much better scalability as the number of nodes increases. In several scenarios, both \golem{} and \dagma{} failed to consistently meet the stopping criterion. For ER6 graphs with $d=100$ nodes, \golem{} failed to converge in two out of three runs, while \dagma{} failed once. Additionally, \dagma{} failed to converge in one out of three runs for $d=1000$. All non-converging runs were excluded from the figures.
  }
  \label{fig:runtime}
\end{figure}

In this section, we discuss the runtime speed of the proposed \ours{} algorithm. We run the proposed algorithm and baselines until the function value converges close to the solution, 
\(f(x_k) - f(\overline{x}) \leq 0.1 \cdot f(\overline{x})\).

Figures~\ref{fig:gumb_exp_er2} and \ref{fig:gumb_exp_er4} compare the performance of \ours{} against \golem{} and \dagma{} on smaller graphs with various structures and noise distributions. Meanwhile, \Cref{fig:runtime} illustrates the scalability of \ours{} on the large graphs.

Both Figures \ref{fig:small} and \ref{fig:runtime} clearly demonstrate that \ours{} significantly outperforms \golem{} and \dagma{} in terms of runtime across both sparse and dense ER graphs in nearly all considered scenarios. The only exception occurs with very small graphs \(d<100\) and high sparsity (ER2, SF2), where \dagma{} is marginally faster than \ours{}. 
However, as graph size and density increase, \ours{} scale efficiently across all scenarios even up until $d = 10000$ nodes. In the case of sparse graphs (Figures \ref{sfig_er2}, \ref{sfig_sf2}), \ours{} consistently converges within a few hours, even for $d=10000$ nodes.

In contrast, increasing graph size causes the computational cost of both \golem{} and \dagma{} to skyrocket. Notably, across all tested graphs (\ref{sfig_er2}), \golem{} exceeds allocated runtime of $36$ hours for $d\geq3000$ nodes, while \dagma{} exceeds for $d\geq 5000$ nodes.

In several experimental scenarios, we also observed that both \golem{} and \dagma{} occasionally failed to meet the stopping criterion, even for smaller graphs. For very small ER6 graphs ($d=100$, \Cref{sfig_er6}), neither method consistently achieved the stopping criterion -- \dagma{} failed to converge once, and \golem{} failed twice out of three random seeds. With \(d = 1000\) nodes, \dagma{} again failed to converge in one out of the three runs. 
All non-converging runs were excluded from the figures.

\subsection{Real Data}
We also evaluate the proposed method against baselines \notears{} \citep{notears}, \golem{} \citep{golem}, and \dagma{} \citep{dagma} on a real-world dataset, \emph {causal protein signaling network data}, provided by \citet{sachs2005causal} that captures the expression levels of proteins and phospholipids in human cells. 
This dataset is widely used in the literature on probabilistic graphical models, with experimental annotations that are well-established in the biological research community. 

The dataset comprises 7,466 samples, of which we utilize the first 853, corresponding to a network with 11 nodes representing proteins and 17 edges denoting their interactions. Despite its relatively small size, it is considered to be a challenging benchmark in recent studies \citep{notears, golem, gao2021dag}. 
For all experiments, we used the first 853 samples for training and the subsequent 902 samples for testing. After the training phase, we employed the same default threshold of 0.3 as was used by the other baseline approaches \notears{}, \golem{}, \dagma{}.

As shown in \Cref{real_data}, our method outperforms both baselines  \golem{} \citep{golem} and \notears{} \citep{notears} in all metrics, the SHD (lower is better), TPR (higher is better) and FPR (smaller is better). More detailed description can be found in \Cref{app_exp}.
We report the total number of edges of the output DAG. We do not report the performance of \dagma{} because it fails to optimize the problem (its iterate \(\mathbf{W}\) diverges from the feasible domain during the first iteration).

\begin{table}
    \centering
    \begin{threeparttable}
        \centering
        \caption{Performance of the top-performing methods on the causal protein signaling network dataset \cite{sachs2005causal}. The threshold for all methods is 0.3. }
        \begin{tabular}{lccccc}
            \label{real_data}
            & SHD$(\downarrow)$& TPR $(\uparrow)$ &FPR $(\downarrow)$&  Total edges & Reference \\ \toprule
            \golem{}&  26 &     0.294 & 0.47 &23 & \citet{golem} \\ \midrule
            \notears{} & 15 &  0.294 & 0.26&15 & \citet{notears}\\ \midrule
            \algours{} &   \textbf{14}     & \textbf{0.411}   & \textbf{0.18} &14 & \Cref{alg:ours}   \\
            \bottomrule
        \end{tabular}
    \end{threeparttable} 
    \label{tab:my_label}
\end{table}

\section{Conclusion}

We introduce a novel framework for learning Directed Acyclic Graphs (DAGs) that addresses the scalability and computational challenges of existing methods. Our approach leverages Stochastic Approximation techniques in combination with Stochastic Gradient Descent (\sgd{})-based methods, allowing for efficient optimization even in high-dimensional settings. A key contribution of our framework is the introduction of new projection techniques that effectively enforce DAG constraints, ensuring that the learned structure adheres to the acyclicity requirement without the need for computationally expensive penalties or constraints seen in prior works.

The proposed framework is theoretically grounded and guarantees convergence to a feasible local minimum. One of its main advantages is its low iteration complexity, making it highly suitable for large-scale structure learning problems, where traditional methods often struggle with runtime and memory limitations. By significantly reducing the per-iteration cost and improving convergence behavior, our framework demonstrates superior scalability when applied to larger datasets and more complex graph structures.

We validate the effectiveness of our method through extensive experimental evaluations across a variety of settings, including both synthetic and real-world datasets. These experiments show that our framework consistently outperforms existing methods such as \golem{} \citep{golem}, \notears{} \citep{notears}, and \dagma{} \citep{dagma}, both in terms of objective loss and runtime, particularly in scenarios involving large and dense graphs. Furthermore, our method exhibits robust performance across different types of graph structures, highlighting its potential applicability to various practical fields such as biology, finance, and causal inference.

\newpage
\bibliography{dag_bib,kamzolov_bib}
\bibliographystyle{iclr2025_conference}

\clearpage

\appendix

\tableofcontents

\section{Detailed Experiment Description} \label{app_exp}
\paragraph{Computing.} Our experiments were carried out on a machine equipped with 80 CPUs and one NVIDIA Quadro RTX A6000 48GB GPU. Each experiment was allotted a maximum wall time of 36 hours as in \dagma{} \cite{dagma}.

\paragraph{Graph Models.} In our experimental simulations, we generate graphs using two established random graph models: 
\begin{itemize}
    \item \textbf{Erdős-Rényi (ER) graphs:} These graphs are constructed by independently adding edges between nodes with a uniform probability. We denote these graphs as ER\(_k\), where \(k d\) represents the expected number of edges. 

    \item \textbf{Scale-Free (SF) graphs}: These graphs follow the preferential attachment process as described in \citet{barabasi1999emergence}. We use the notation SF\(_k\) to indicate a scale-free graph with expected \(k d\) edges and an attachment exponent of \(\beta = 1\), consistent with the preferential attachment process. Since we focus on directed graphs, this model corresponds to Price's model, a traditional framework used to model the growth of citation networks.
\end{itemize}

It is important to note that ER graphs are inherently undirected. To transform them into Directed Acyclic Graphs (DAGs), we generate a random permutation of the vertex labels from 1 to \(d\), then orient the edges according to this ordering. For SF graphs, edges are directed as new nodes are added, ensuring that the resulting graph is a DAG. After generating the ground-truth DAG, we simulate the structural equation model (SEM) for linear cases, conducting experiments accordingly.

\paragraph{Metrics.} The performance of each algorithm is assessed using the following four key metrics:
\begin{itemize}
    \item \textbf{Structural Hamming Distance (SHD):} A widely used metric in structure learning that quantifies the number of edge modifications (additions, deletions, and reversals) required to transform the estimated graph into the true graph.
    \item \textbf{True Positive Rate (TPR):} This metric calculates the proportion of correctly identified edges relative to the total number of edges in the ground-truth DAG.
    \item \textbf{False Positive Rate (FPR):} This measures the proportion of incorrectly identified edges relative to the total number of absent edges in the ground-truth DAG.

    \item  \textbf{Runtime:} The time taken by each algorithm to complete its execution provides a direct measure of the algorithm's computational efficiency.
\item  \textbf{Stochastic gradient computations:} Number of gradient computed.

\end{itemize}

\paragraph{Linear SEM.} In the linear case, the functions are directly parameterized by the weighted adjacency matrix \(W\). Specifically, the system of equations is given by \(X_i = X\mathbf{W}_i + N_i,\) where \(\mathbf{W} = [\mathbf{W}_1 | \cdots | \mathbf{W}_d] \in \mathbb{R}^{d \times d}\), and \(N_i \in \mathbb{R}\) represents the noise. The matrix \(\mathbf{W}\) encodes the graphical structure, meaning there is an edge \(X_j \to X_i\) if and only if \(W_{j,i} \neq 0\). Starting with a ground-truth DAG \(B \in \{0,1\}^{d \times d}\) obtained from one of the two graph models, either ER or SF, edge weights were sampled independently from \(\text{Unif}[-1, -0.05] \cup [0.05, 1]\) to produce a weight matrix \(\mathbf{W} \in \mathbb{R}^{d \times d}\). Using this matrix \(\mathbf{W}\), the data \(X = X\mathbf{W} + N \) was sampled under the following three noise models:

\begin{itemize}
    \item \textbf{Gaussian noise:} \(N_i \sim N(0, 1)\) for all \(i \in [d]\),
    \item \textbf{Exponential noise:} \(N_i \sim \text{Exp}(1)\) for all \(i \in [d]\),
    \item \textbf{Gumbel noise:} \(N_i \sim \text{Gumbel}(0, 1)\) for all \(i \in [d]\).
\end{itemize}

Using these noise models, random datasets \(X \in \mathbb{R}^{n \times d}\) were generated by independently sampling the rows according to one of the models described above. Unless specified otherwise, each simulation generated \(n = 5000\) training samples and a validation set of $10,000$ samples.

The implementation details of the baseline methods are as follows:
\begin{itemize}
    \item \textbf{\notears{} \cite{notears}}was implemented using the authors' publicly available Python code, which can be found at \url{https://github.com/xunzheng/notears}. This method employs a least squares score function, and we used their default set of hyperparameters without modification.

    \item \textbf{\golem{} \cite{golem}} was implemented using the authors' Python code, available at \url{https://github.com/ignavierng/golem}, along with their PyTorch version at \url{https://github.com/huawei-noah/trustworthyAI/blob/master/gcastle/castle/algorithms/gradient/notears/torch/golem_utils/golem_model.py}. We adopted the default hyperparameter settings provided by the authors, specifically \( \lambda_1 = 0.02 \) and \( \lambda_2 = 5 \). For additional details of their method, we refer to Appendix F in \cite{golem}
    \item \textbf{\dagma{} \cite{dagma}} was implemented using the authors' Python code, which is available at \url{https://github.com/kevinsbello/dagma}. We used the default hyperparameters provided in their implementation.

\end{itemize}

\paragraph{Thresholding.}
Following the approach taken in previous studies, including the baseline methods \citep{notears, golem, dagma}, for all the methods, we apply a final thresholding step of 0.3 to effectively reduce the number of false discoveries.

\section{Experiments} \label{exp_fig}
Plots show performance of the algorithms \algours{}, \golem{}, \dagma{} for combinations of number of vertices $d\in \{10, 50, 100, 500, 1000, 3000, 5000, 10000\}$, graph types $\in \{ER, SF\}$, average density of graphs $k\in \{2,4,6\}$, and noise types to be either Gaussian, Exponential, or Gumbel. 

Plots are grouped by the noise type and the number of vertices of the graph and arranged into figures.

\begin{figure*}
    \centering
    \begin{subfigure}{\textwidth}
        \centering
        \includegraphics[width=\thirdwidth]{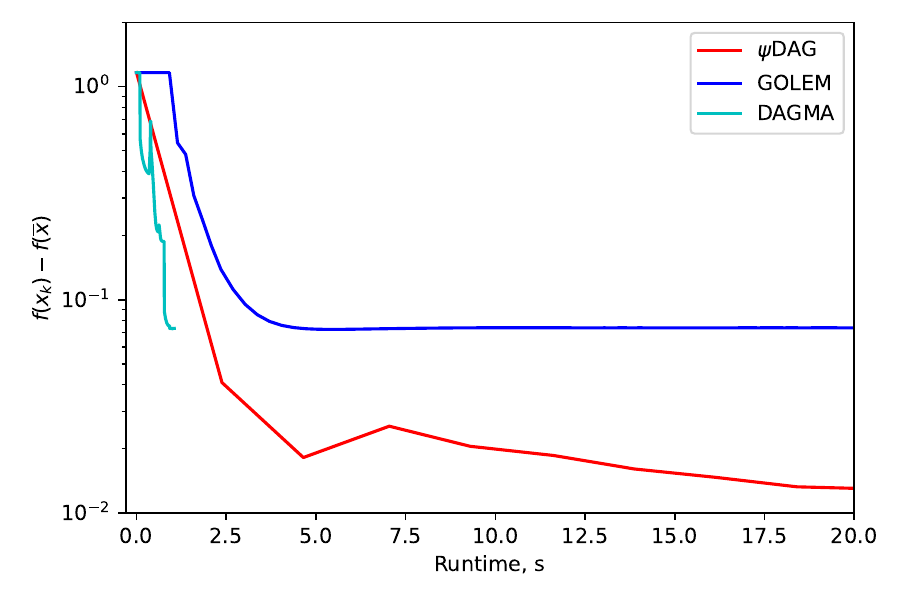}
        \includegraphics[width=\thirdwidth]{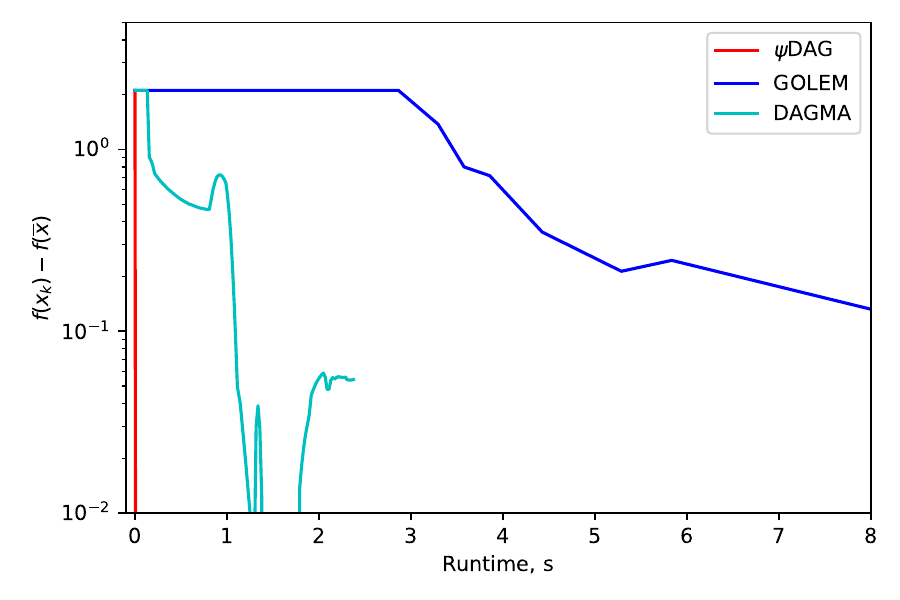}
        \includegraphics[width=\thirdwidth]{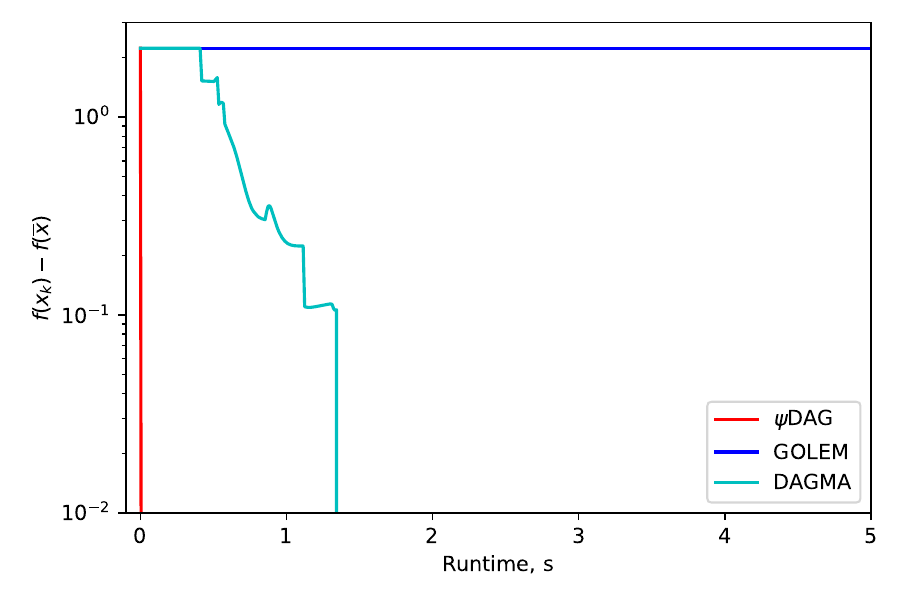}
        \includegraphics[width=\thirdwidth]{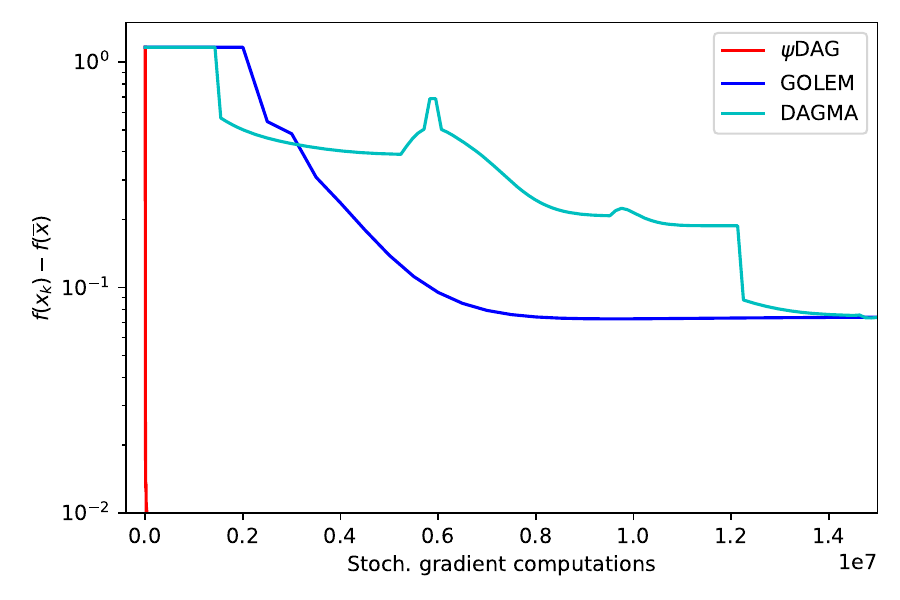}
        \includegraphics[width=\thirdwidth]{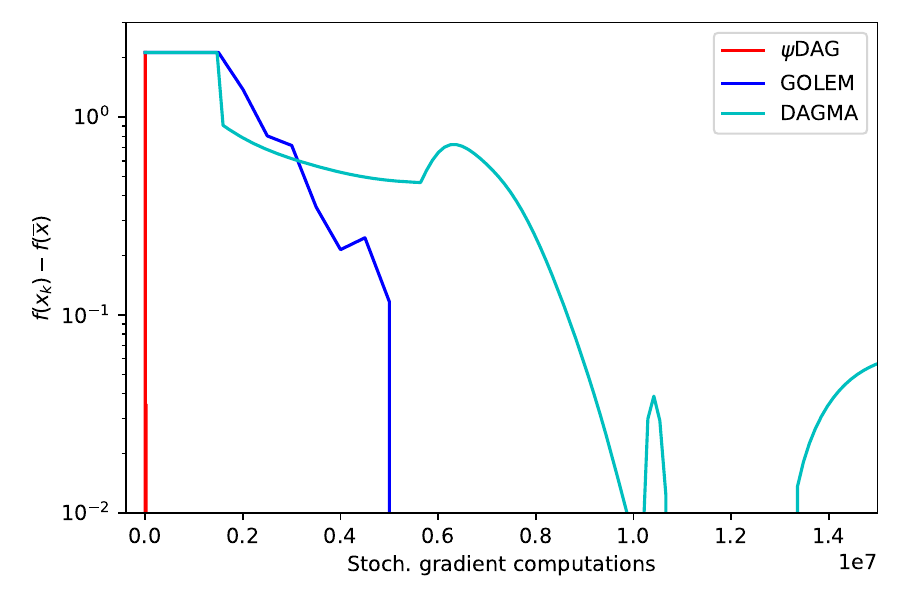}
        \includegraphics[width=\thirdwidth]{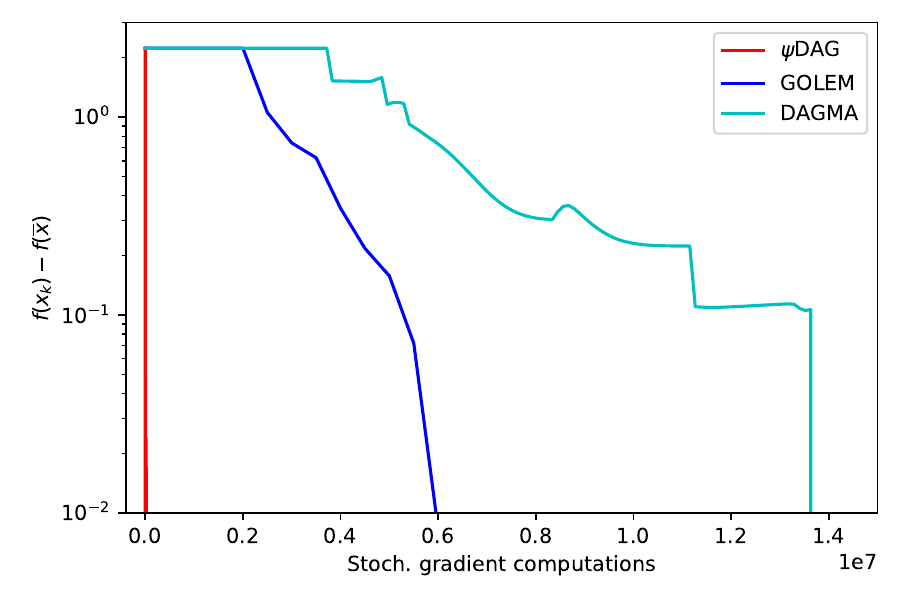}
        \caption{$d=10$ vertices}
    \end{subfigure}

    \begin{subfigure}{\textwidth}
        \centering
        \includegraphics[width=\thirdwidth]{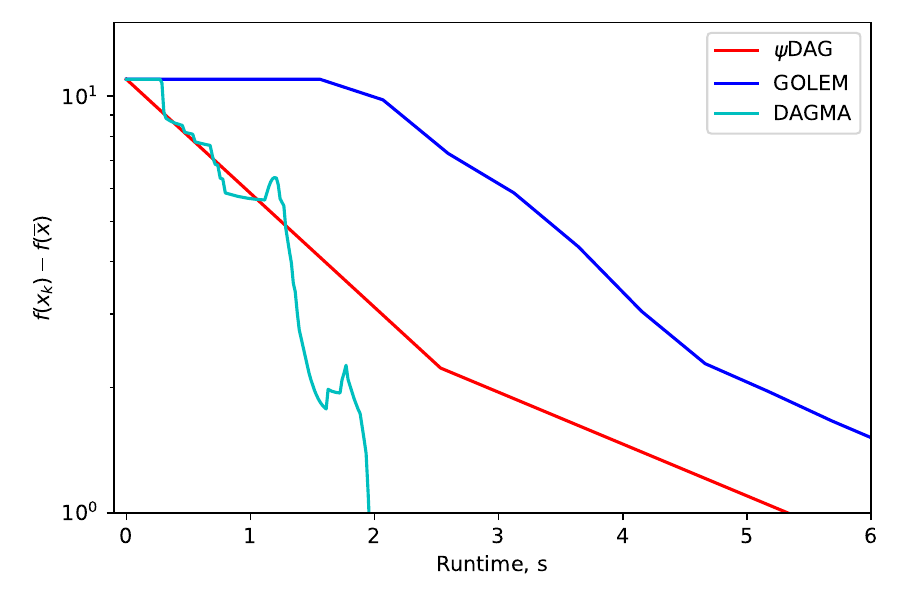}
        \includegraphics[width=\thirdwidth]{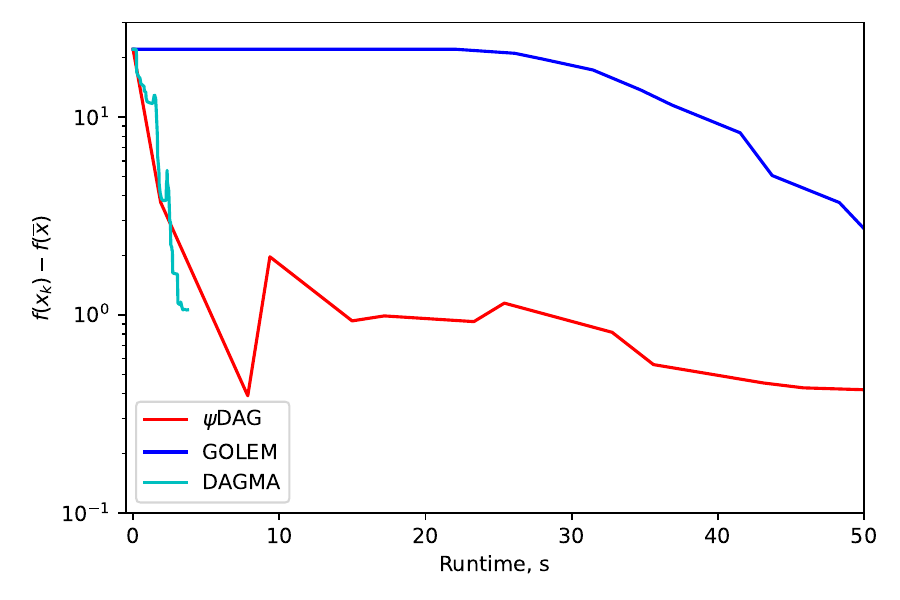}
        \includegraphics[width=\thirdwidth]{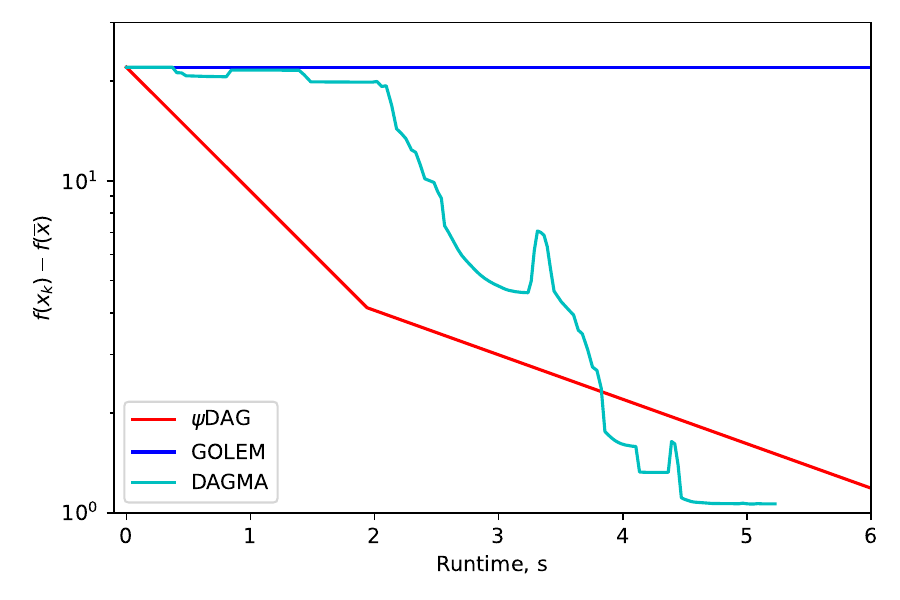}
        \includegraphics[width=\thirdwidth]{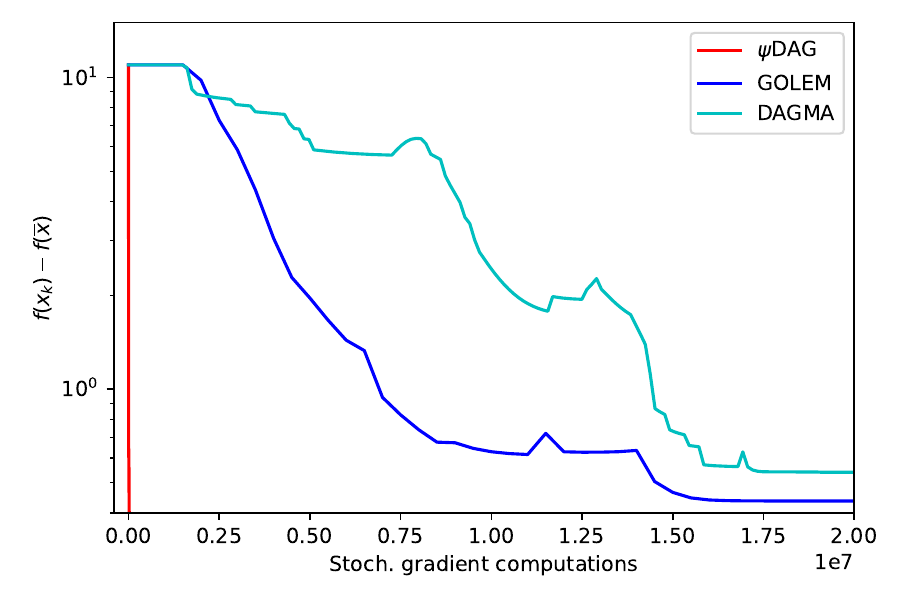}
        \includegraphics[width=\thirdwidth]{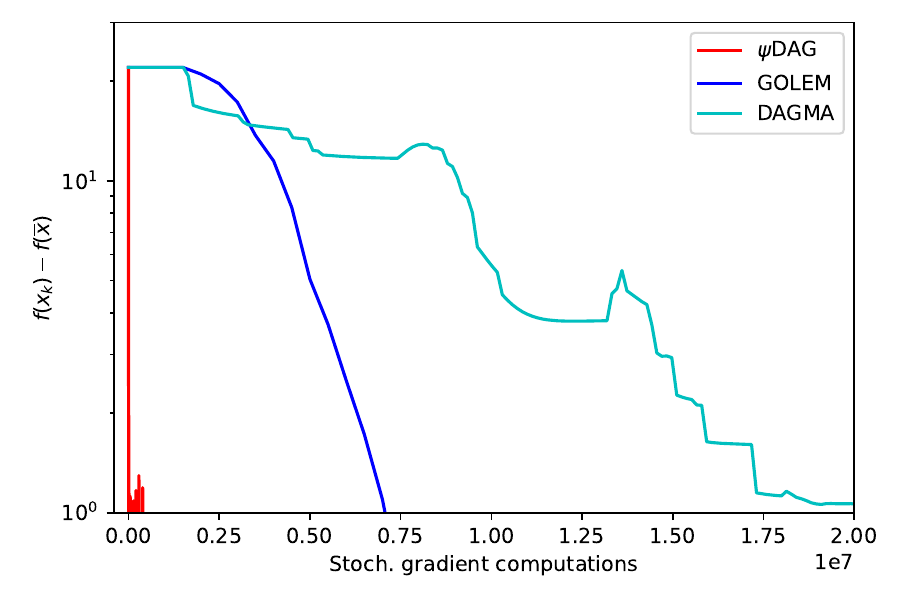}
        \includegraphics[width=\thirdwidth]{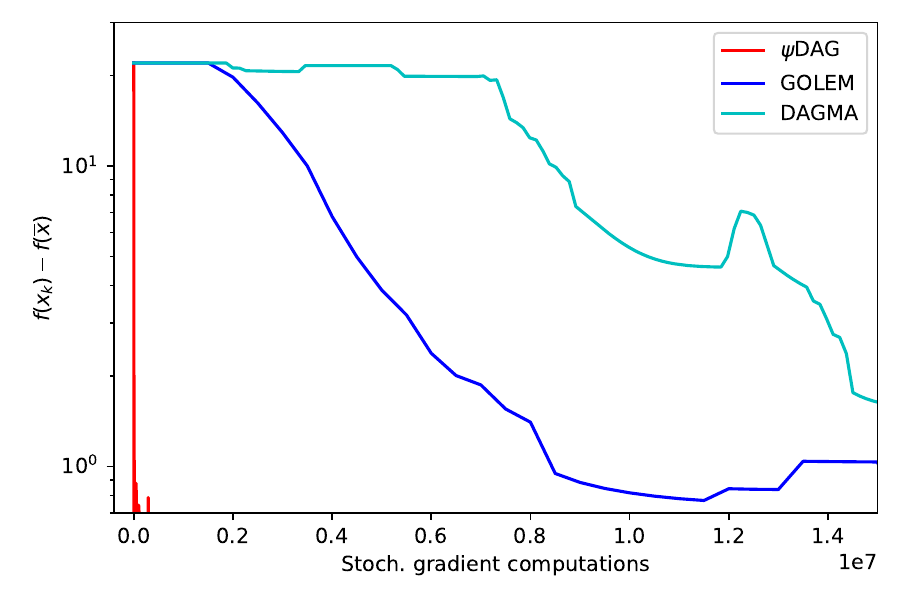}
        \caption{$d=50$ vertices}
    \end{subfigure}

    \begin{subfigure}{\textwidth}
        \centering
        \includegraphics[width=\thirdwidth]{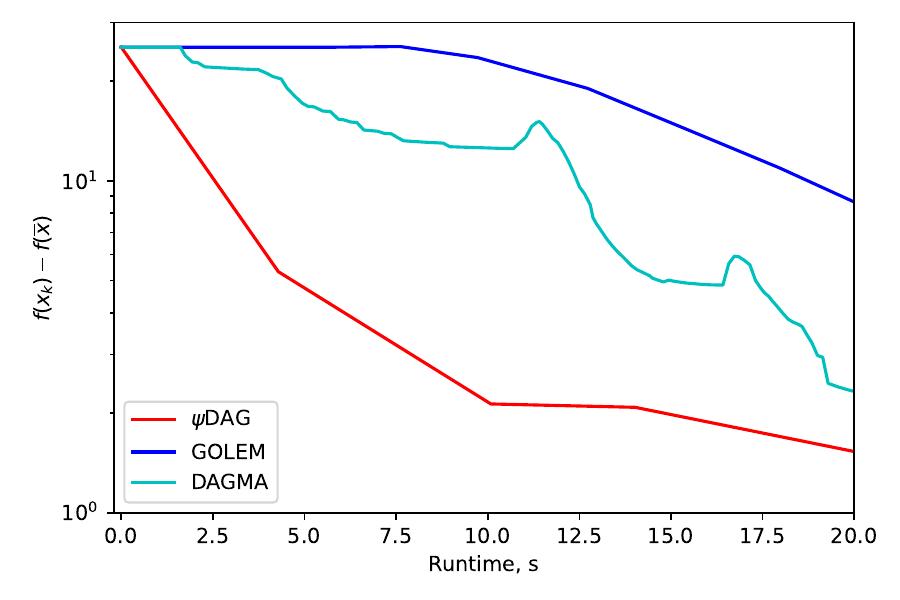}
        \includegraphics[width=\thirdwidth]{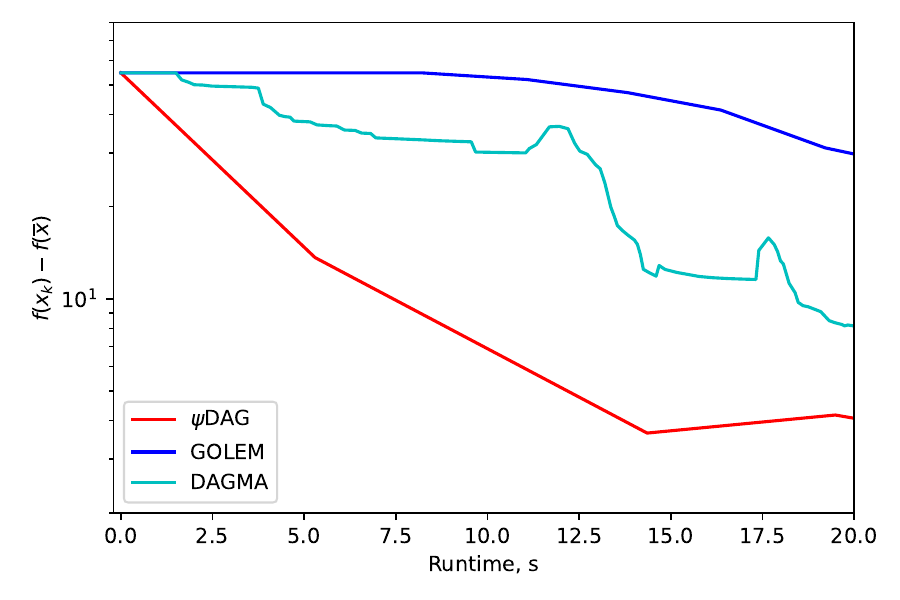}
        \includegraphics[width=\thirdwidth]{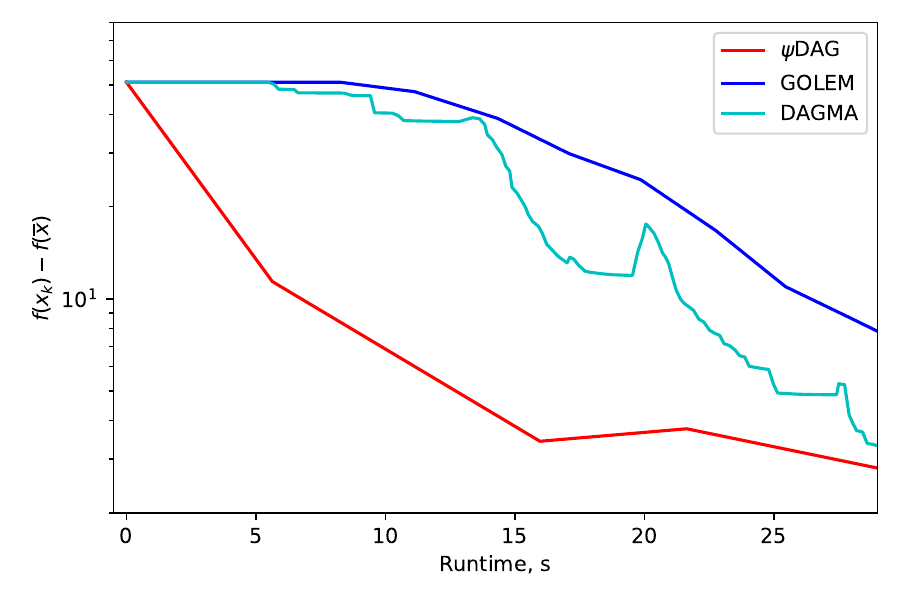}
        \includegraphics[width=\thirdwidth]{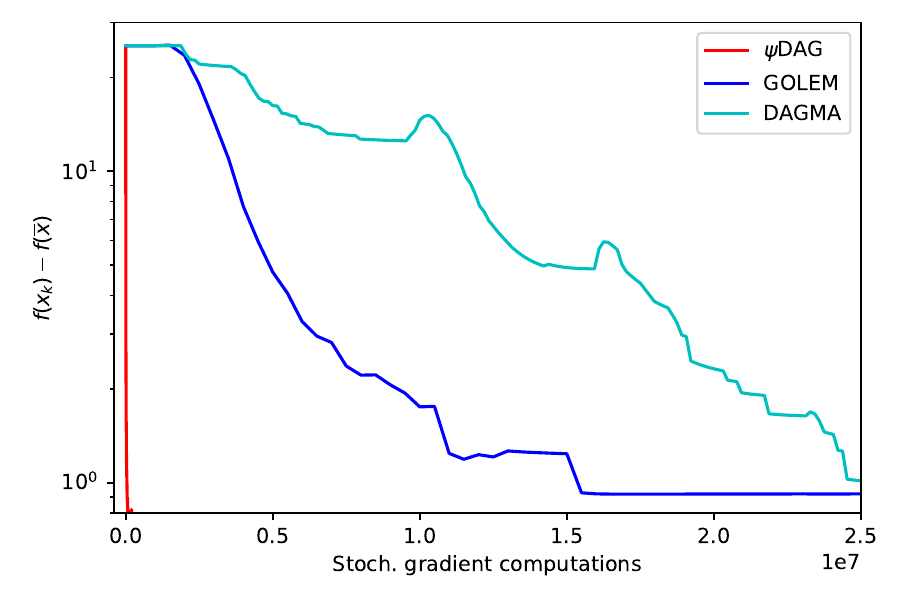}
        \includegraphics[width=\thirdwidth]{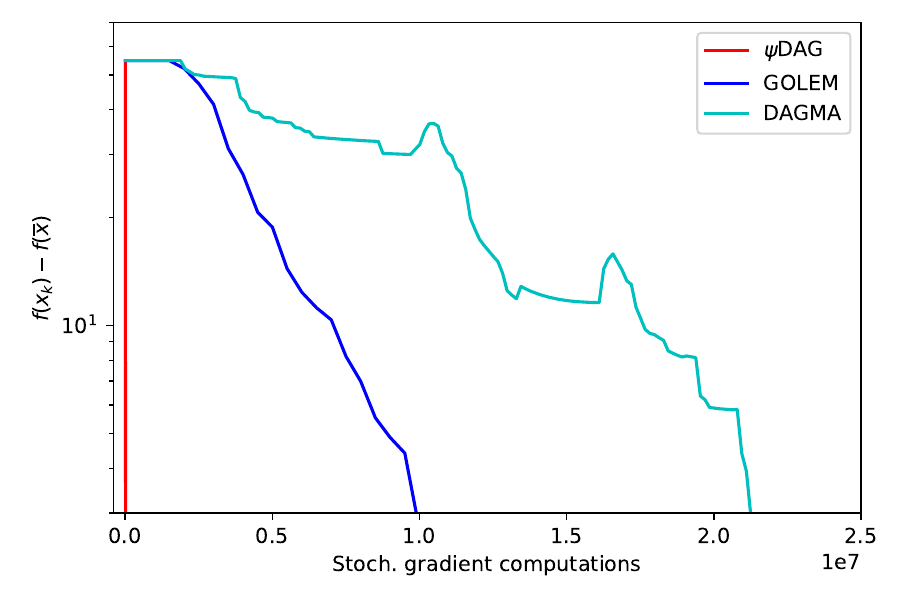}
        \includegraphics[width=\thirdwidth]{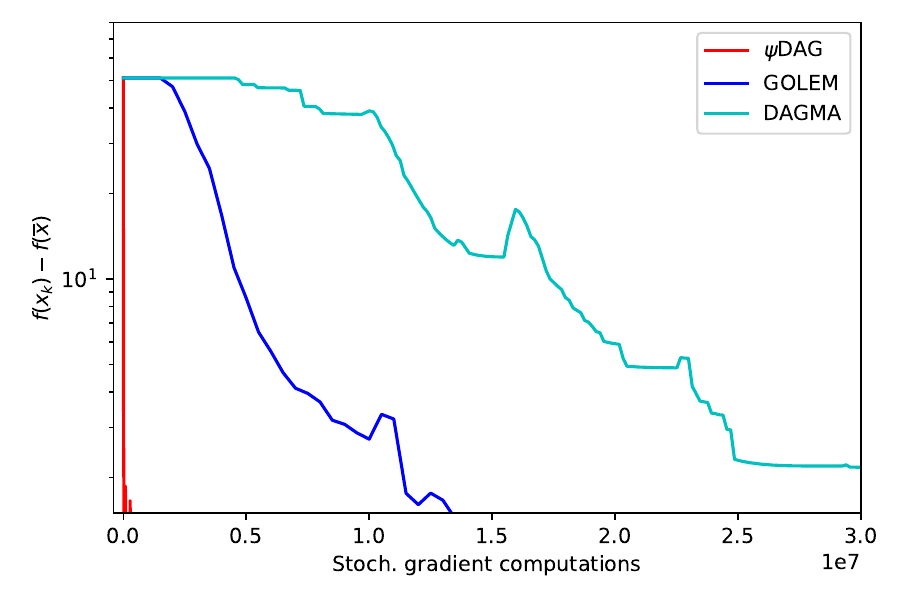}
        \caption{$d=100$ vertices}
    \end{subfigure}

    \caption{Linear SEM methods on graphs of type ER2 with different noise distributions: Gaussian (first), exponential (second), Gumbel (third).}
    \label{fig:er2_small}
\end{figure*}

\begin{figure*}
    \centering
    
    \begin{subfigure}{\textwidth}
        \centering
        \includegraphics[width=\thirdwidth]{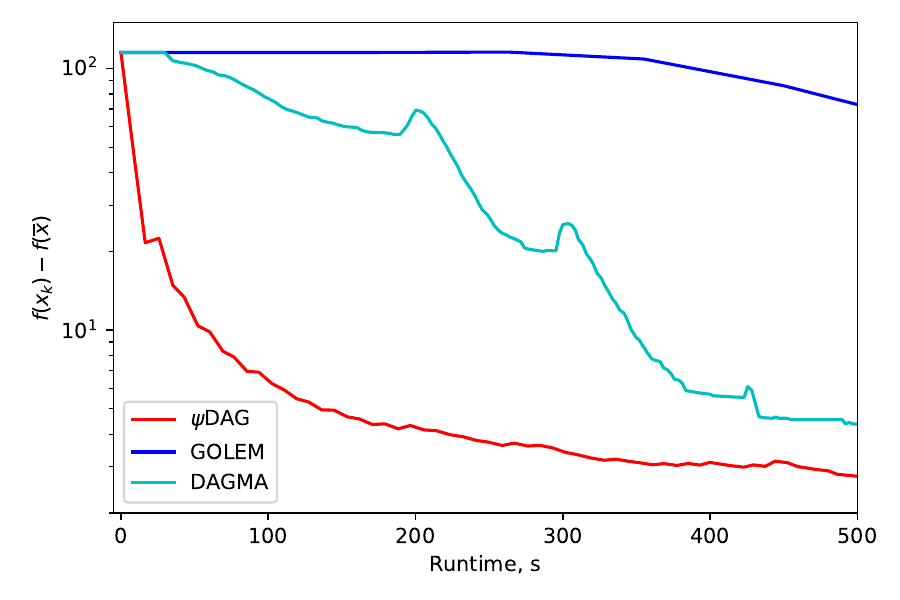}
        \includegraphics[width=\thirdwidth]{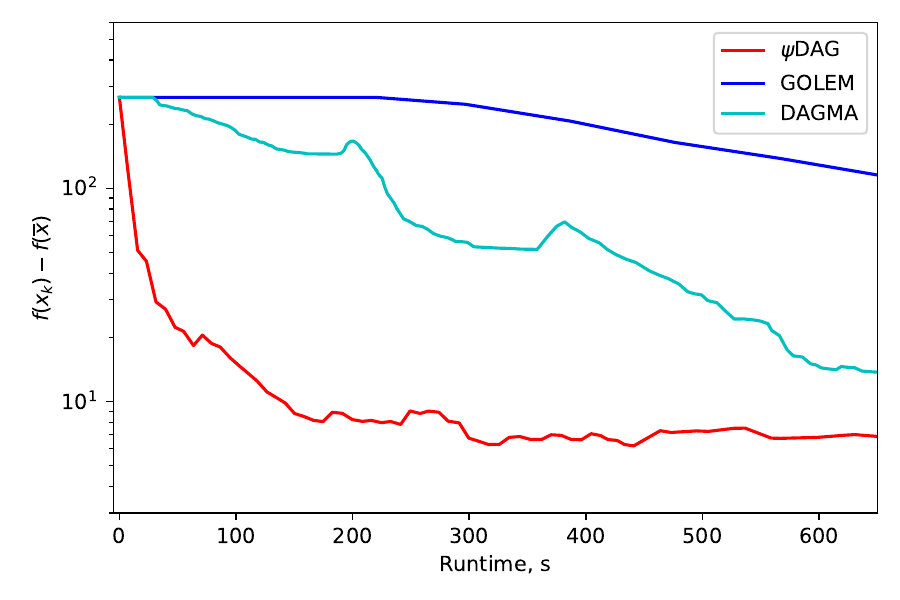}
        \includegraphics[width=\thirdwidth]{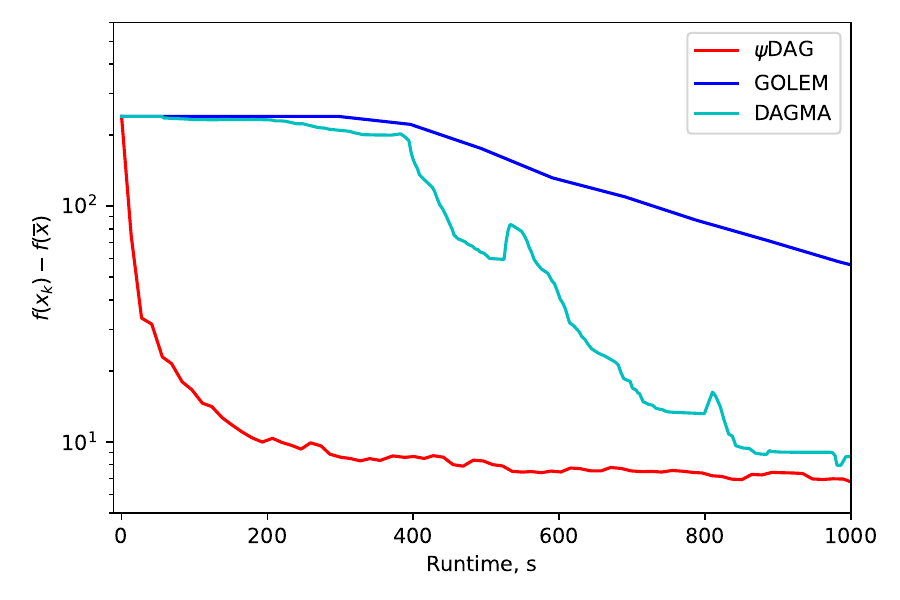}
        \includegraphics[width=\thirdwidth]{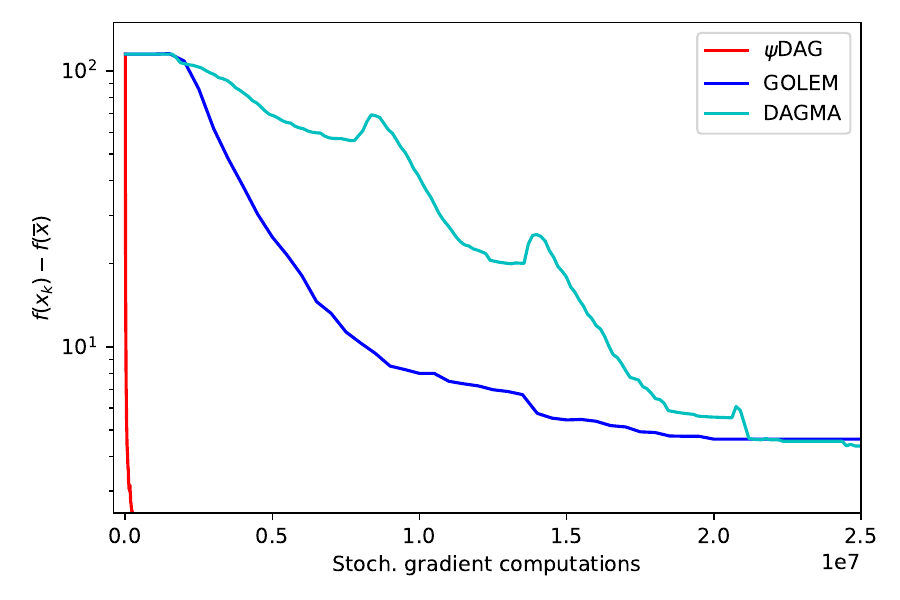}
        \includegraphics[width=\thirdwidth]{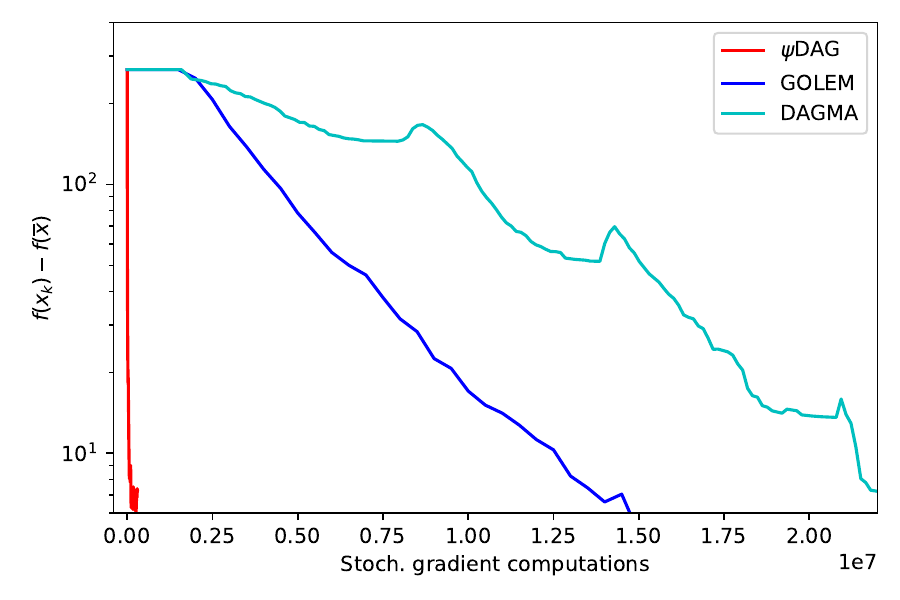}
        \includegraphics[width=\thirdwidth]{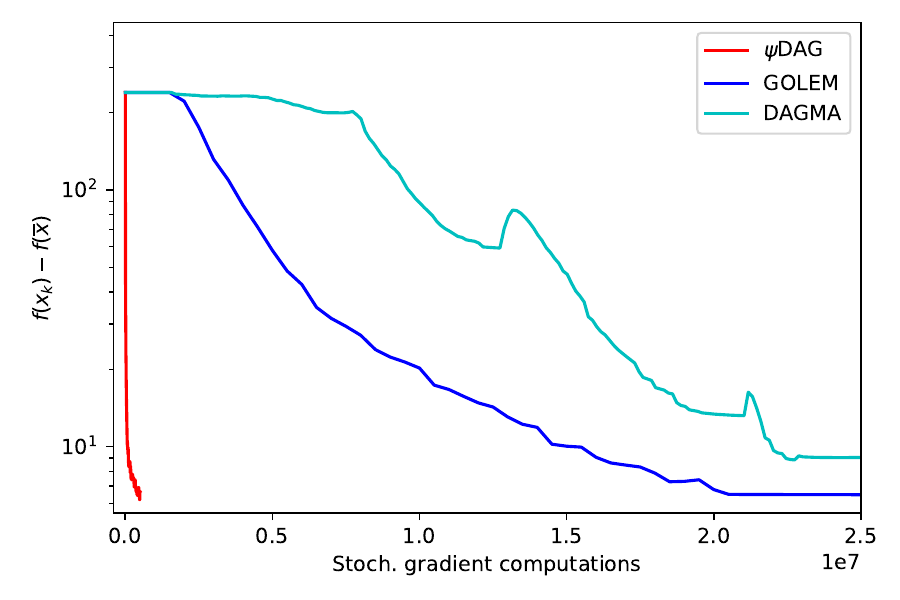}
        \caption{$d=500$ vertices}
    \end{subfigure}

    \begin{subfigure}{\textwidth}
        \centering
        \includegraphics[width=\thirdwidth]{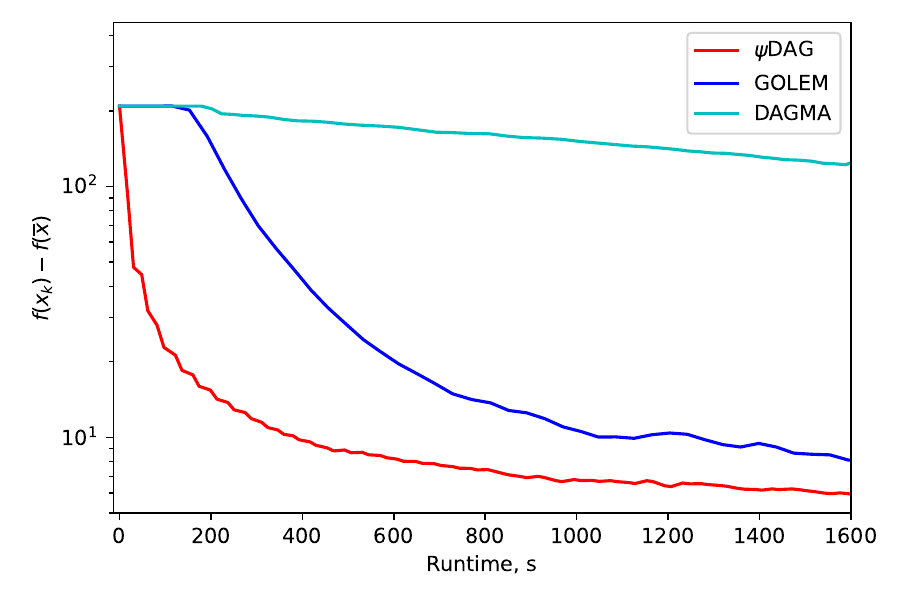}
        \includegraphics[width=\thirdwidth]{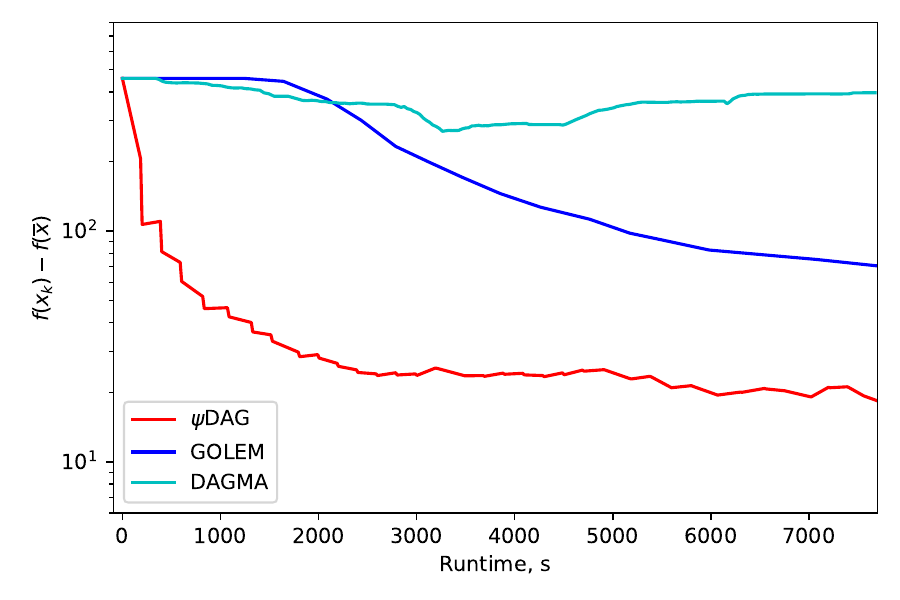}
        \includegraphics[width=\thirdwidth]{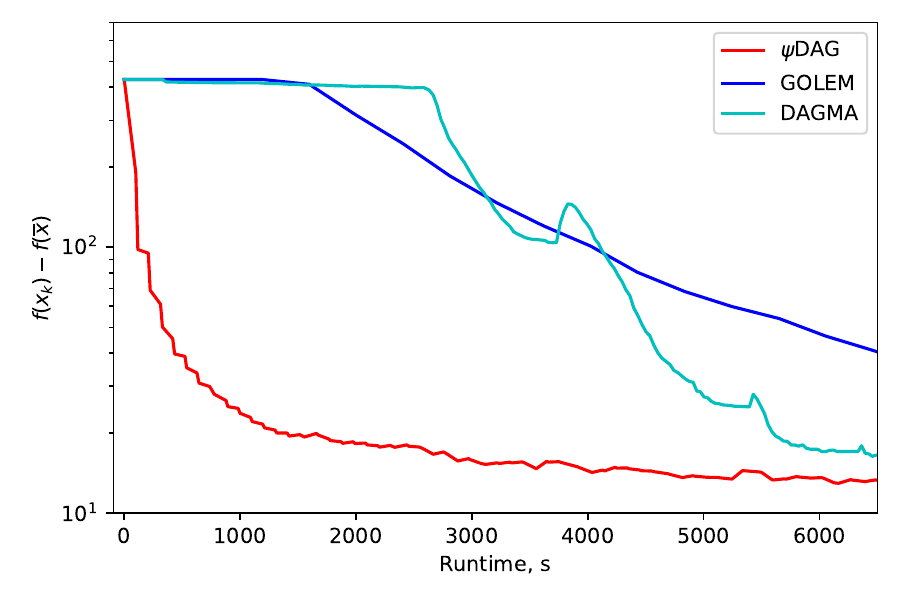}
        \includegraphics[width=\thirdwidth]{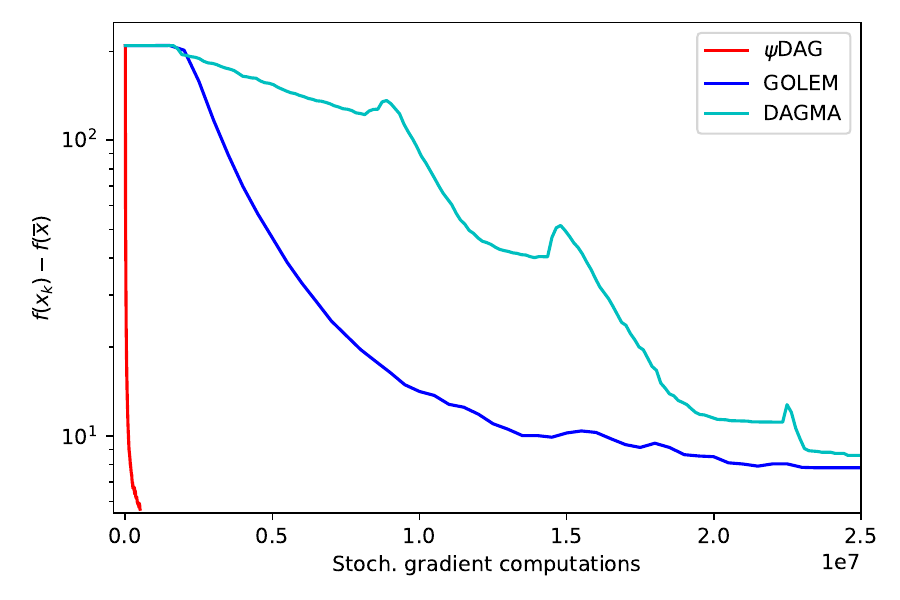}
        \includegraphics[width=\thirdwidth]{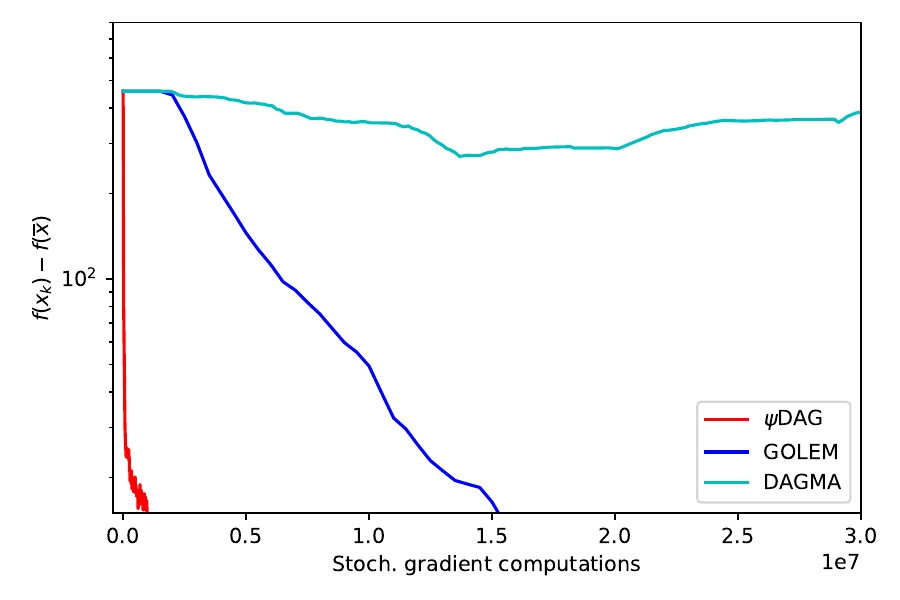}
        \includegraphics[width=\thirdwidth]{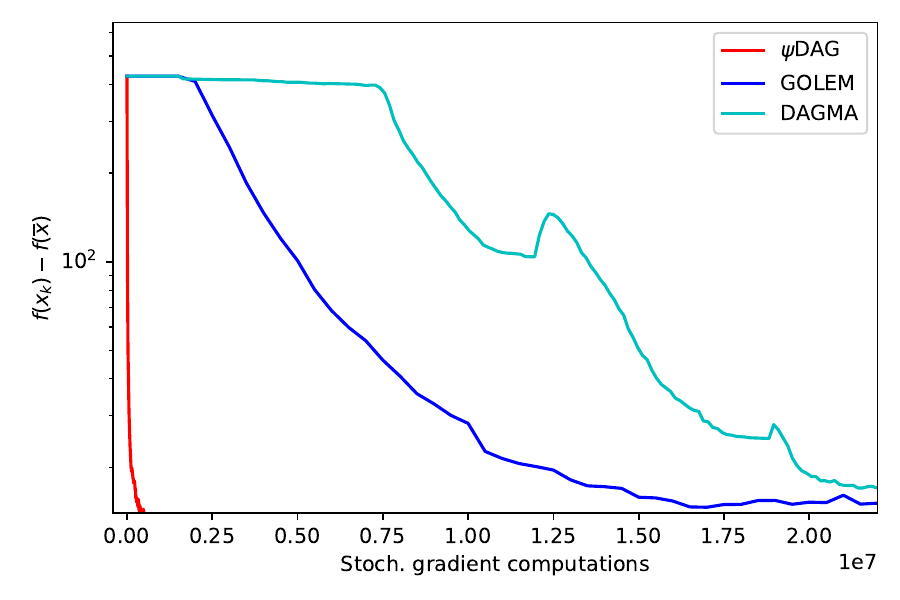}
        \caption{$d=1000$ vertices}\label{fig:1000_er2}
    \end{subfigure}

    \begin{subfigure}{\textwidth}
        \includegraphics[width=\thirdwidth]{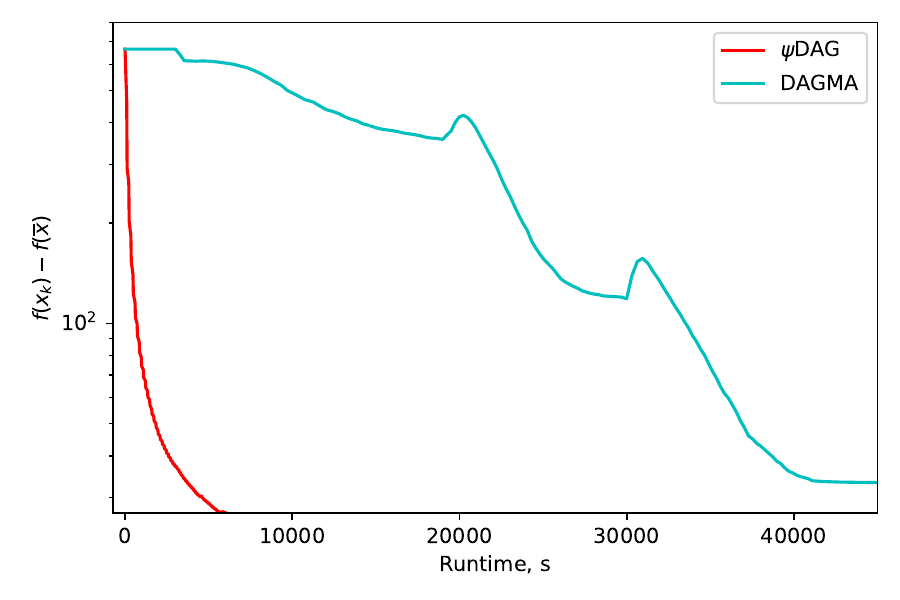}
        \includegraphics[width=\thirdwidth]{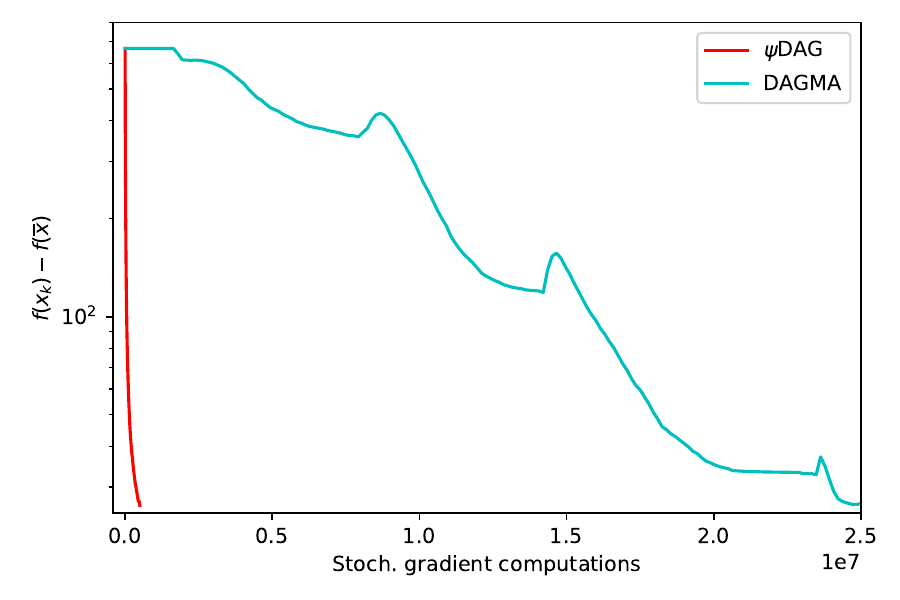}
        \caption{$d=3000$ vertices}
    \end{subfigure}

    \caption{Linear SEM methods on graphs of type ER2 with different noise distributions: Gaussian (first), exponential (second), Gumbel (third).}
    \label{fig:er2_medium}
\end{figure*}

\begin{figure*}
    \centering
    \begin{subfigure}{\textwidth}
        \centering
        \includegraphics[width=\thirdwidth]{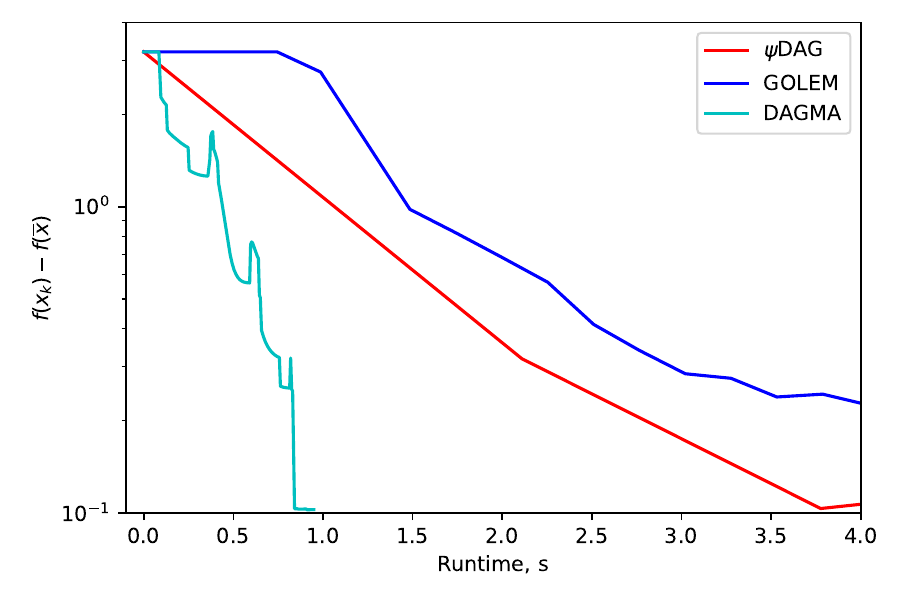}
        \includegraphics[width=\thirdwidth]{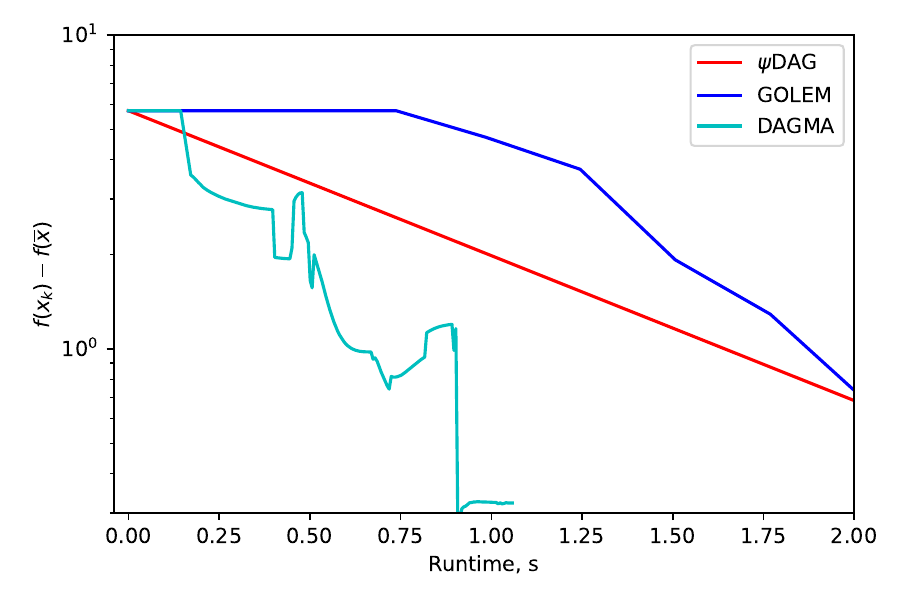}
        \includegraphics[width=\thirdwidth]{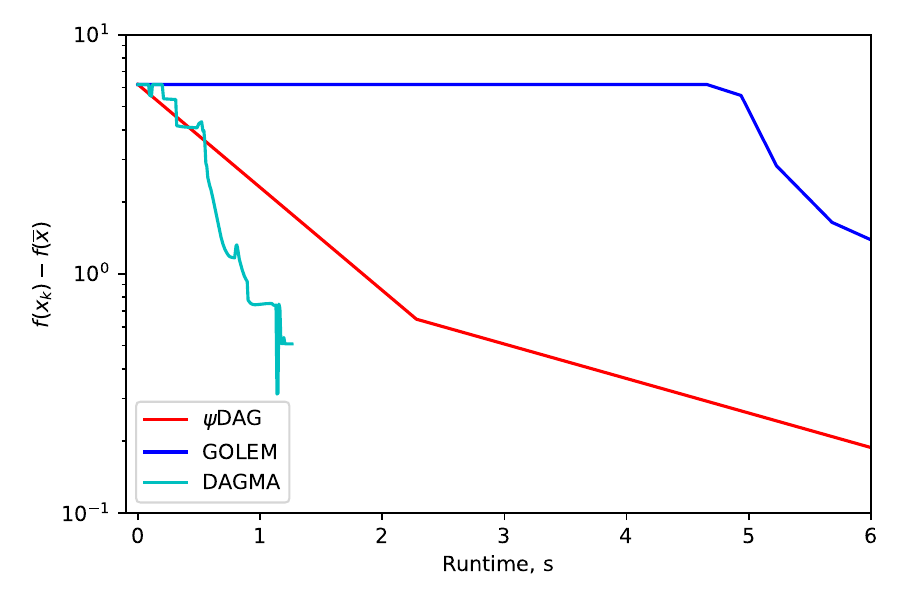}
        \includegraphics[width=\thirdwidth]{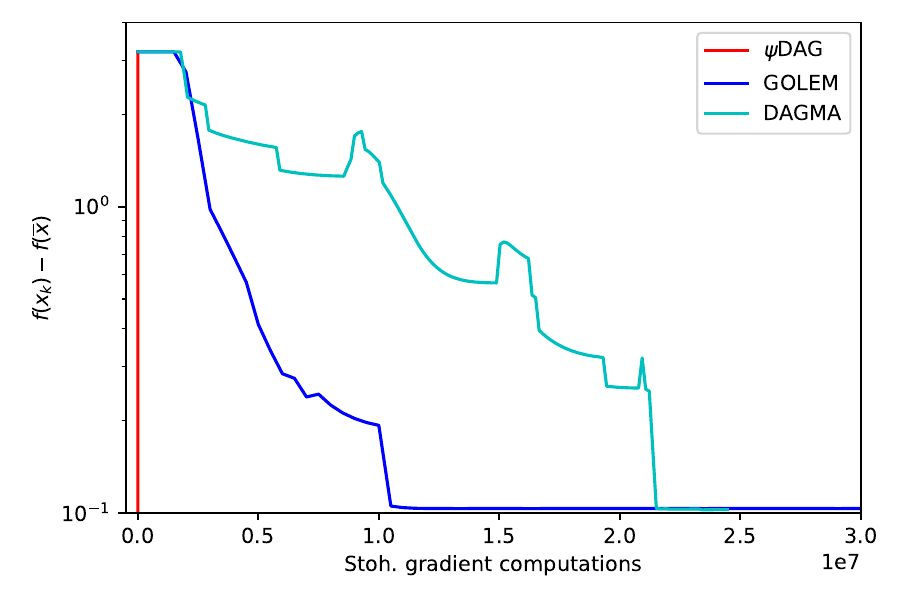}
        \includegraphics[width=\thirdwidth]{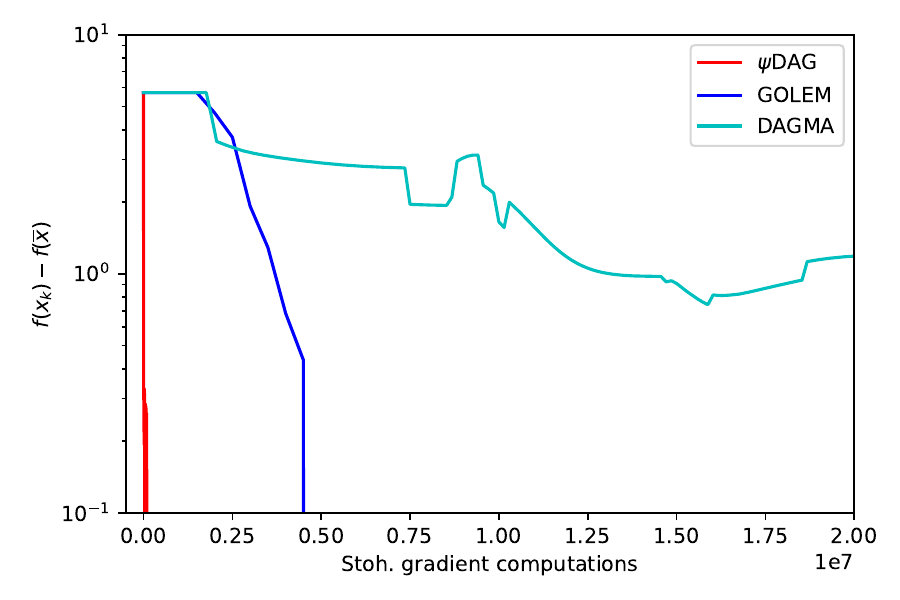}
        \includegraphics[width=\thirdwidth]{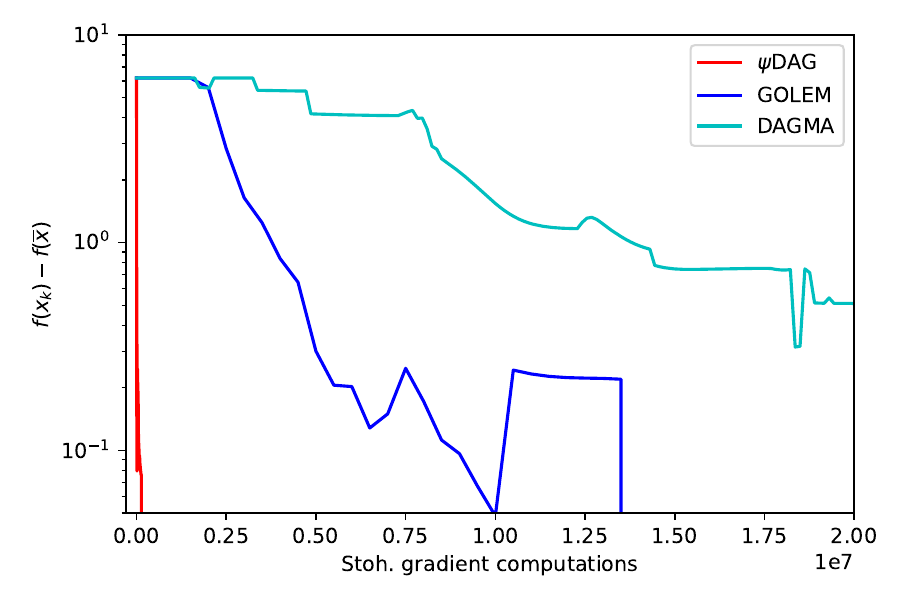}
        \caption{$d=10$ vertices}
    \end{subfigure}

    \begin{subfigure}{\textwidth}
        \centering
        \includegraphics[width=\thirdwidth]{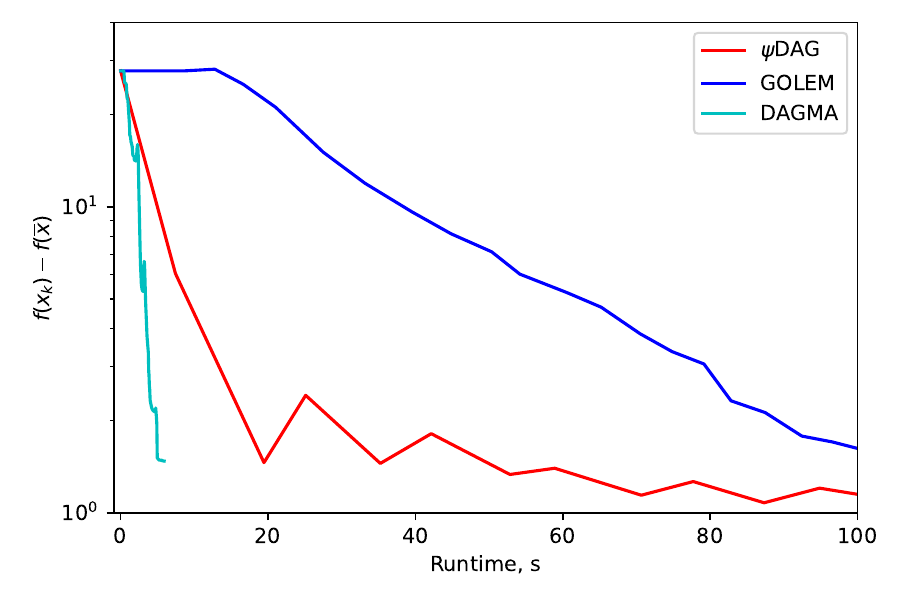}
        \includegraphics[width=\thirdwidth]{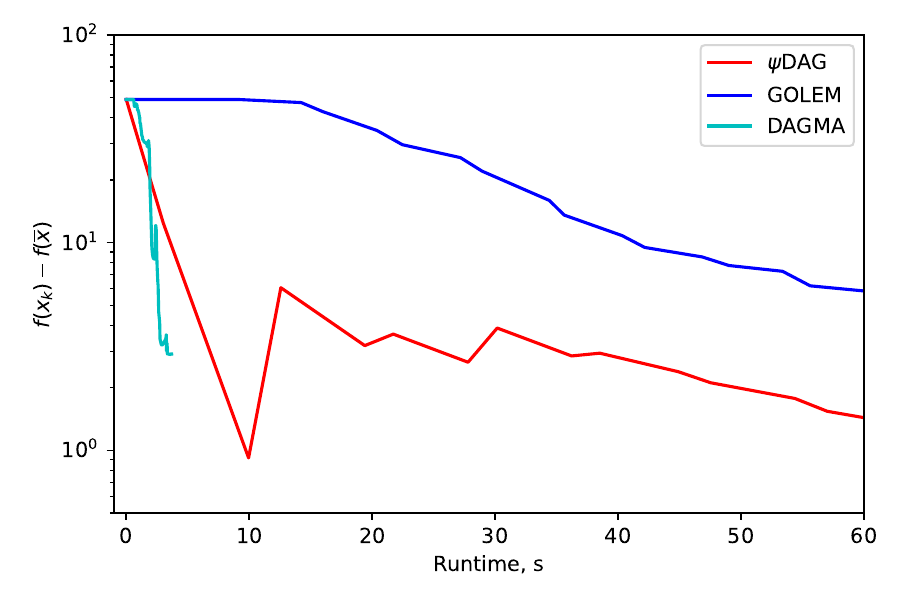}
        \includegraphics[width=\thirdwidth]{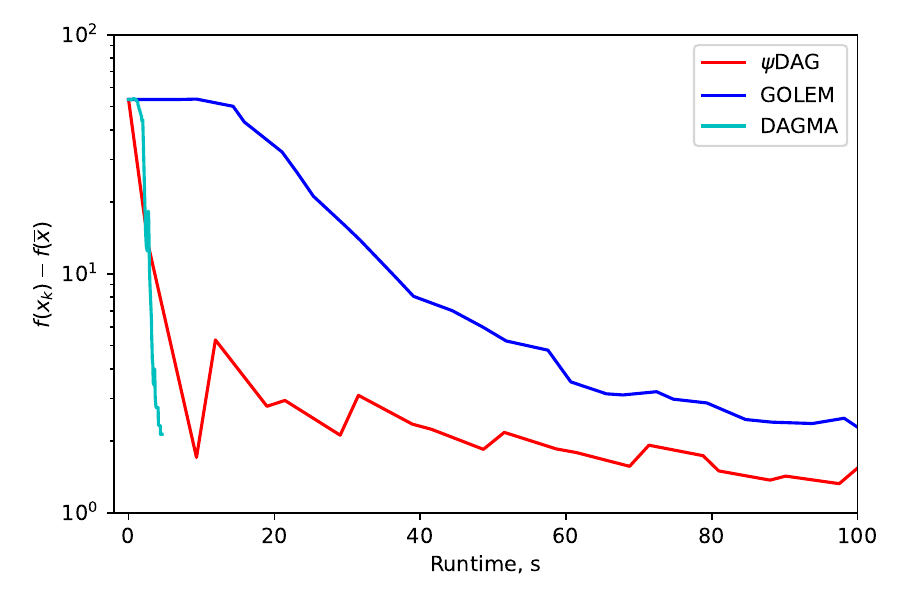}
        \includegraphics[width=\thirdwidth]{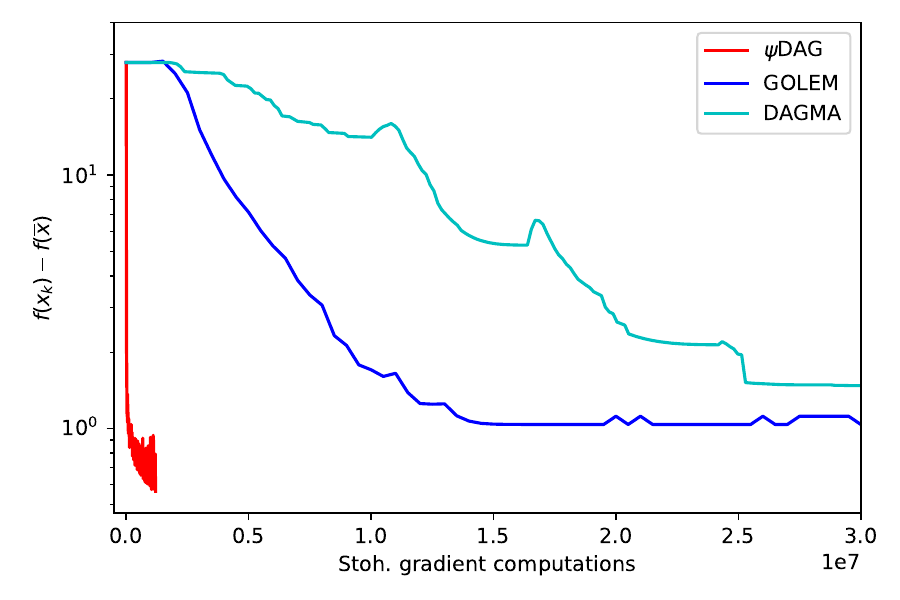}
        \includegraphics[width=\thirdwidth]{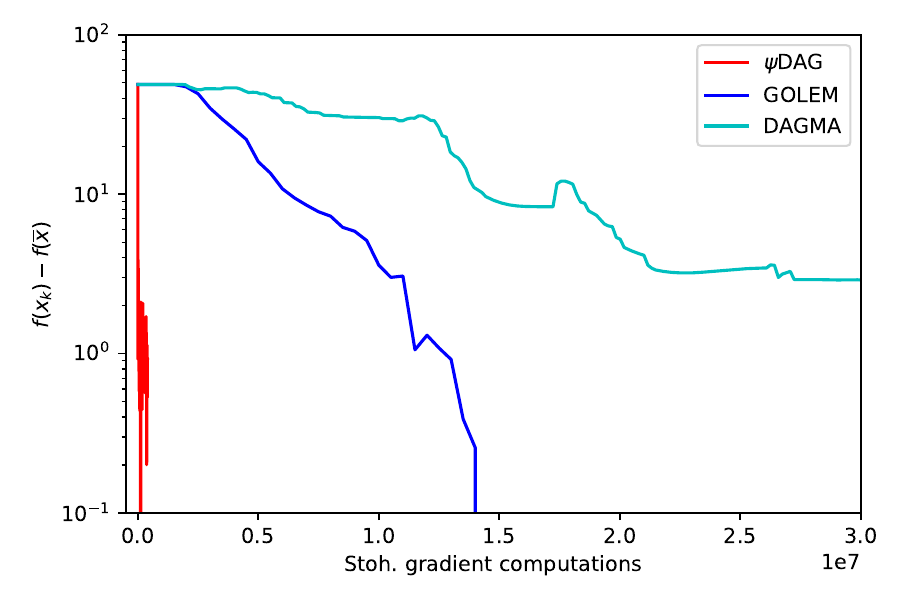}
        \includegraphics[width=\thirdwidth]{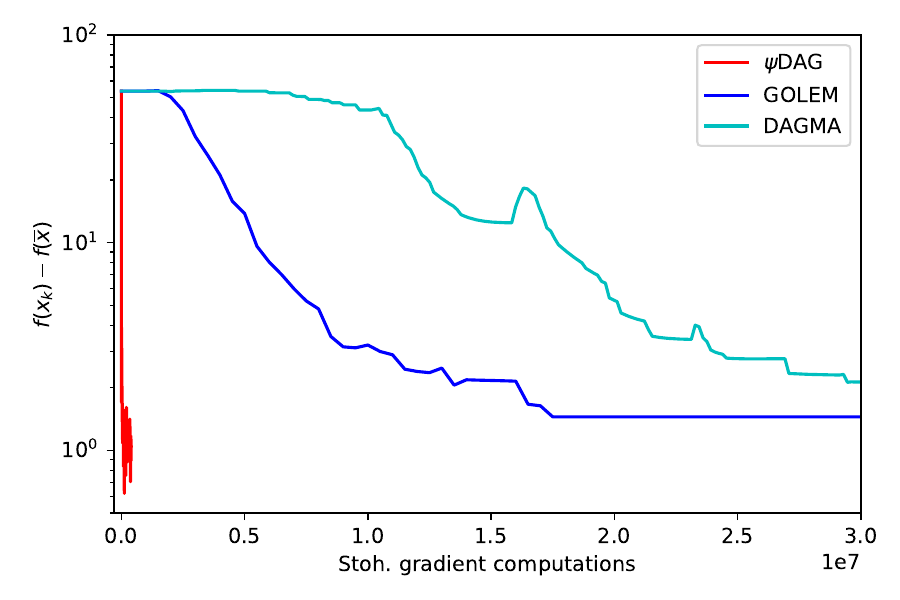}
        \caption{$d=50$ vertices}
    \end{subfigure}

    \begin{subfigure}{\textwidth}
        \centering
        \includegraphics[width=\thirdwidth]{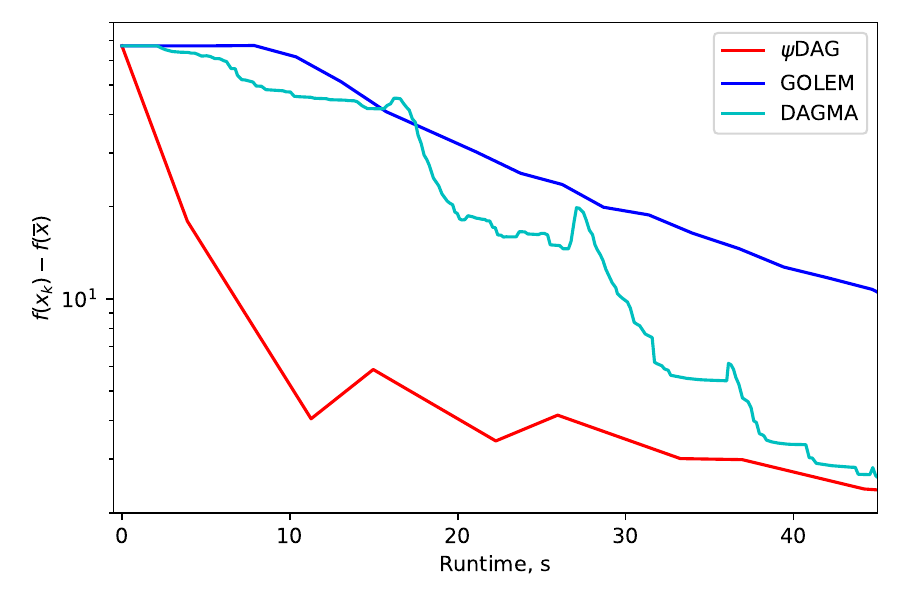}
        \includegraphics[width=\thirdwidth]{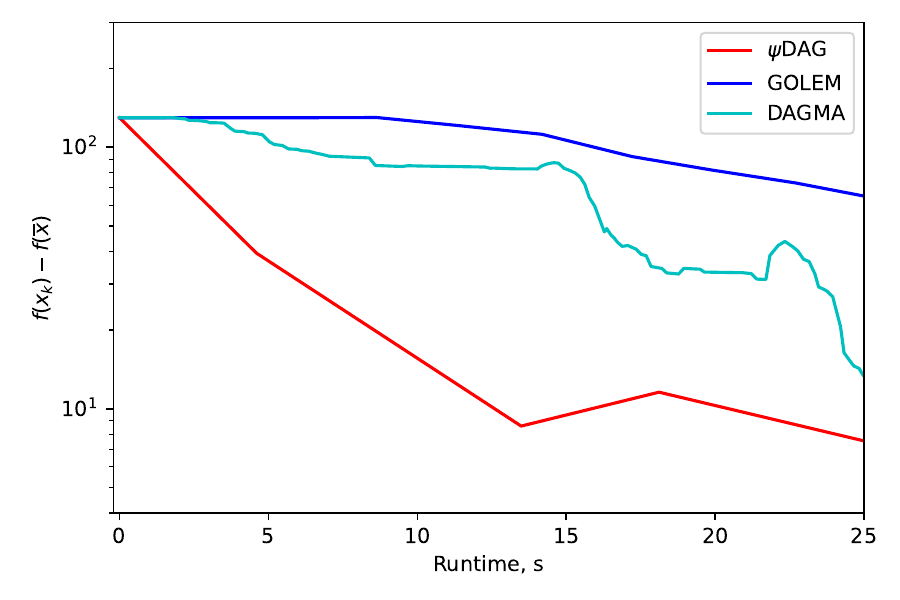}
        \includegraphics[width=\thirdwidth]{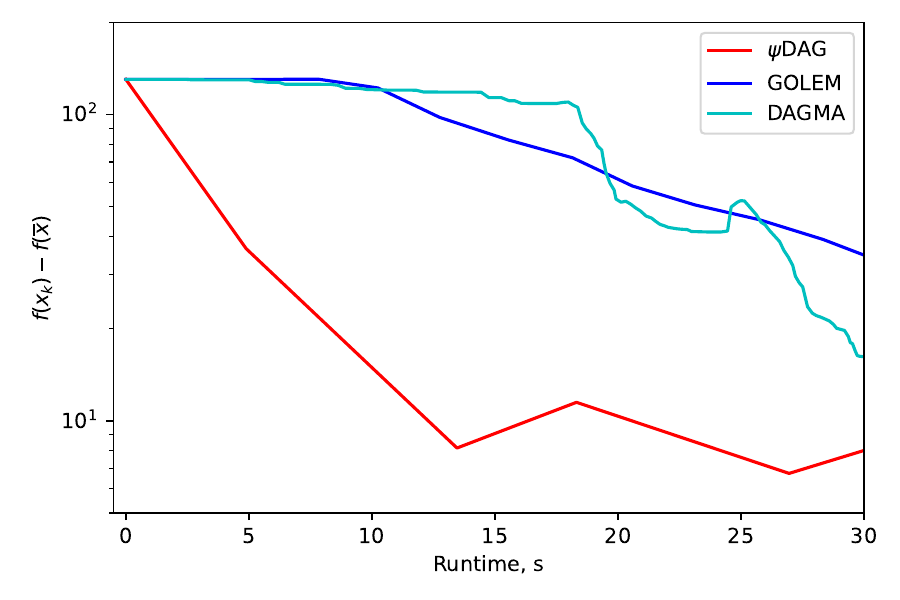}
        \includegraphics[width=\thirdwidth]{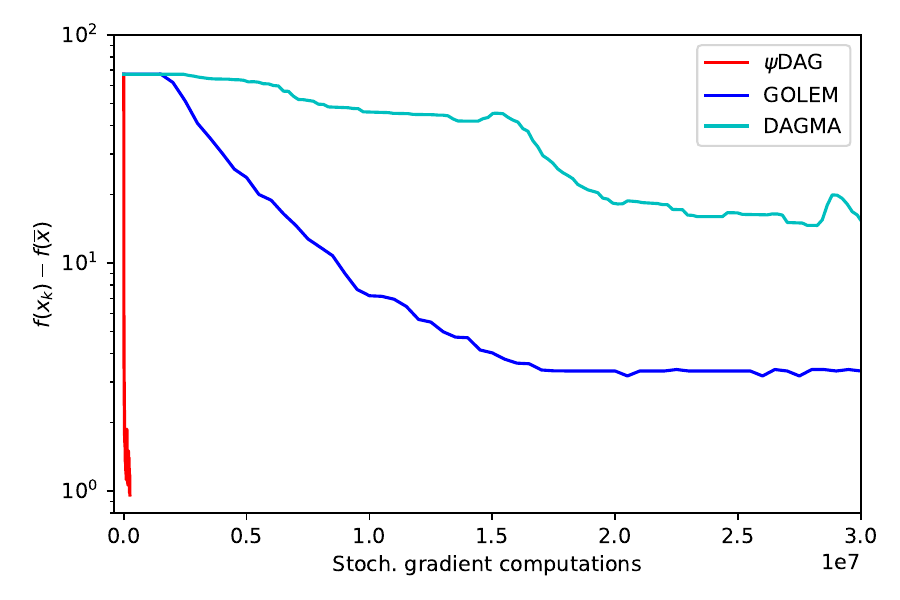}
        \includegraphics[width=\thirdwidth]{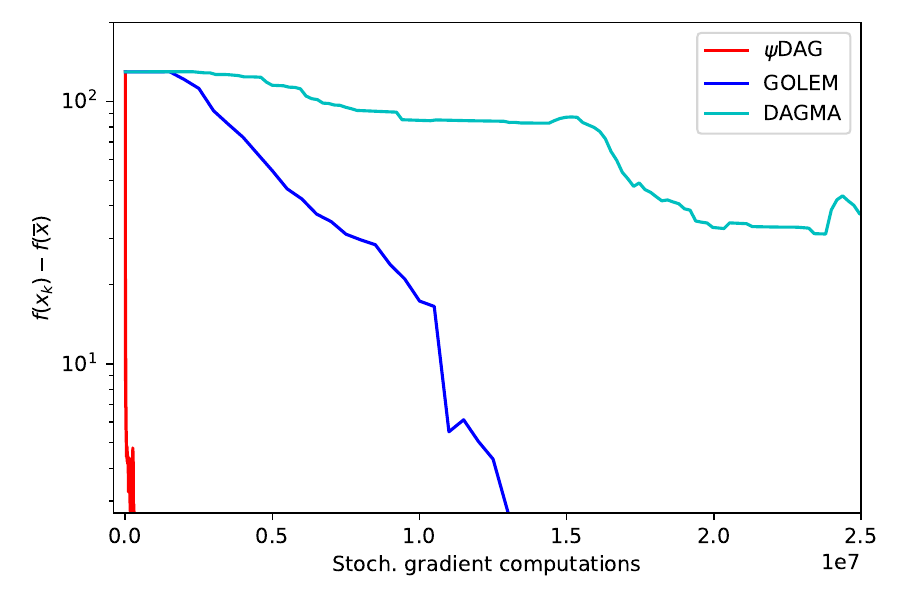}
        \includegraphics[width=\thirdwidth]{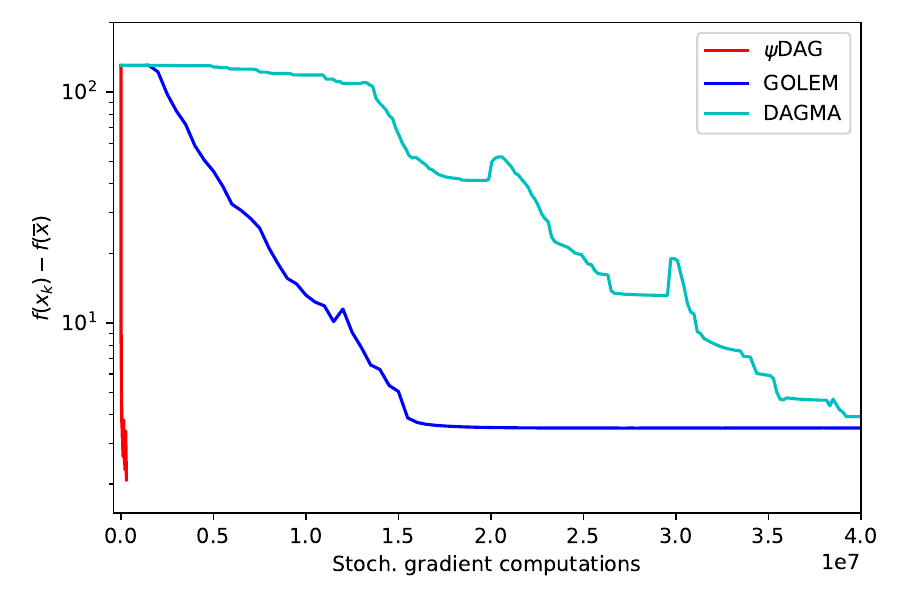}
        \caption{$d=100$ vertices}
    \end{subfigure}

    \caption{Linear SEM methods on graphs of type ER4 with different noise distributions: Gaussian (first), exponential (second), Gumbel (third).}
    \label{fig:er4_small}
\end{figure*}

\begin{figure*}
    \centering
    
    \begin{subfigure}{\textwidth}
        \centering
        \includegraphics[width=\thirdwidth]{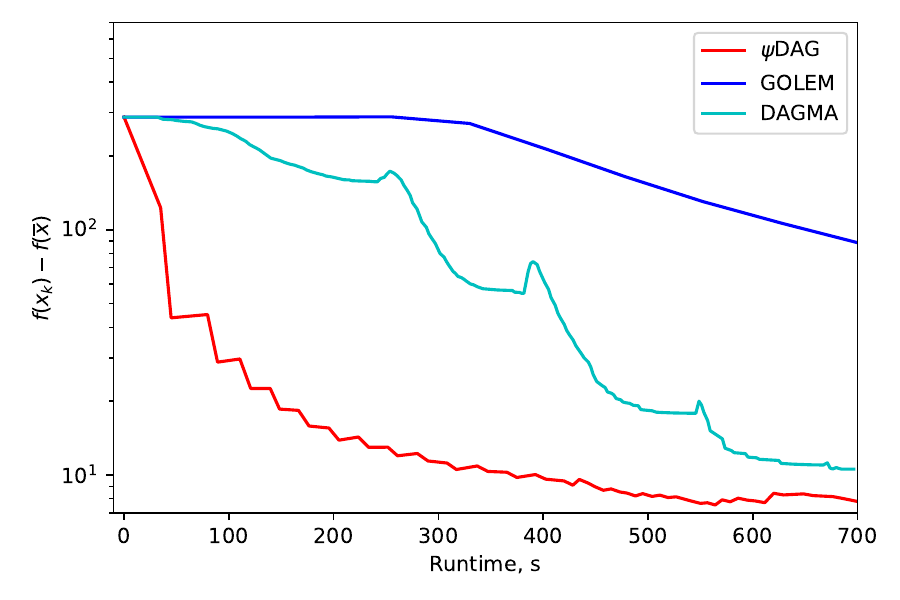}
        \includegraphics[width=\thirdwidth]{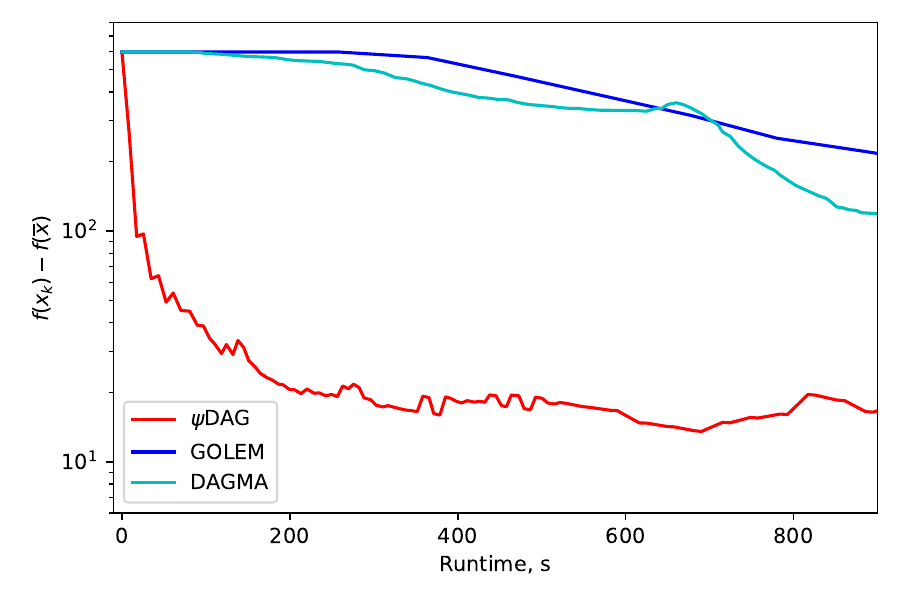}
        \includegraphics[width=\thirdwidth]{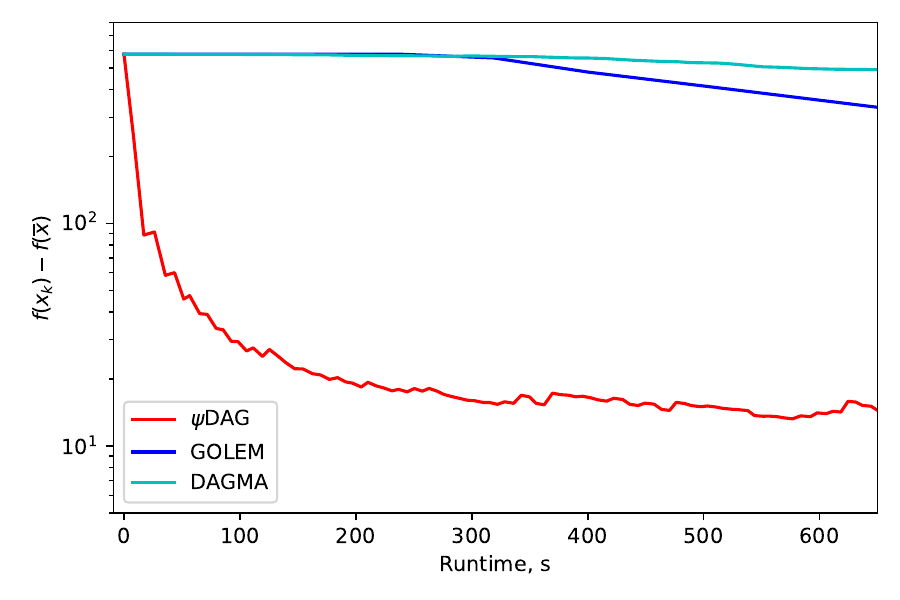}
        \includegraphics[width=\thirdwidth]{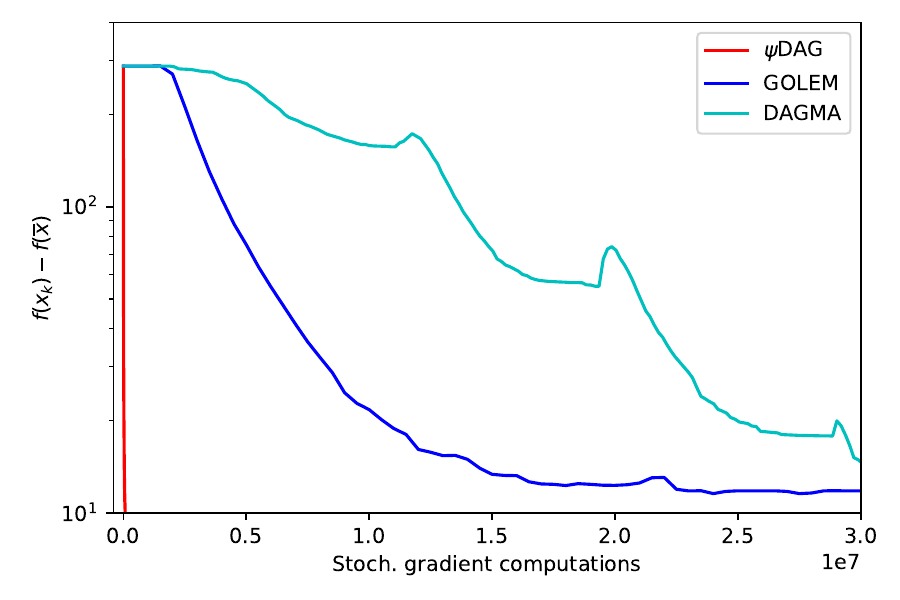}
        \includegraphics[width=\thirdwidth]{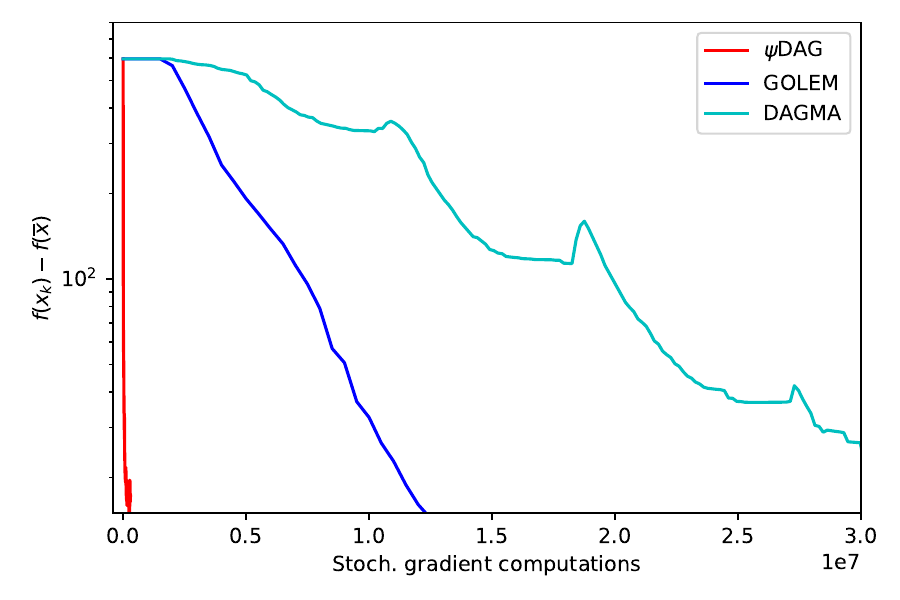}
        \includegraphics[width=\thirdwidth]{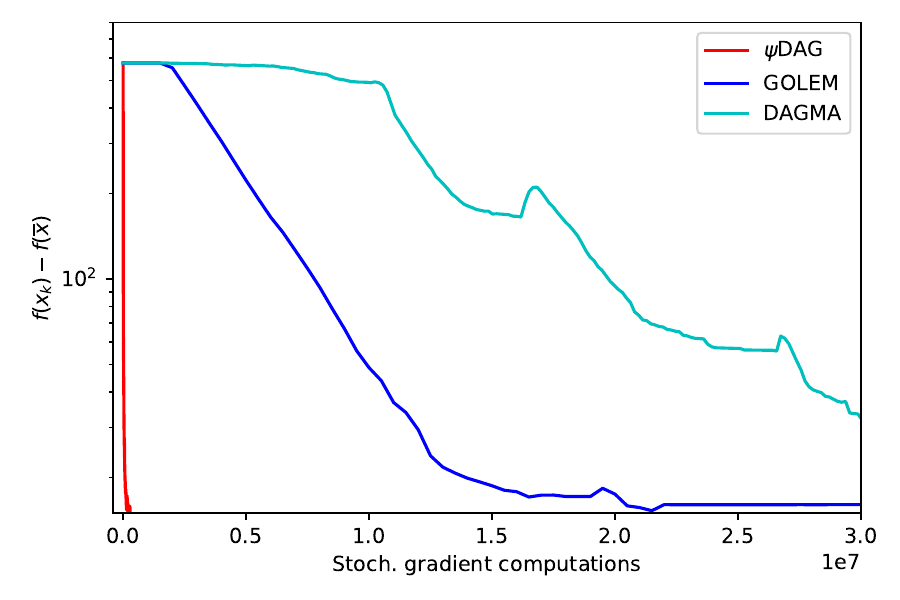}
        \caption{$d=500$ vertices}
    \end{subfigure}

    \begin{subfigure}{\textwidth}
        \centering
        \includegraphics[width=\thirdwidth]{figures/dagma_img/d=1000_ER4_5000_gaussian_ev_time.pdf}
        \includegraphics[width=\thirdwidth]{figures/dagma_img/d=1000_ER4_5000_exponential_time.pdf}
        \includegraphics[width=\thirdwidth]{figures/dagma_img/d=1000_ER4_5000_gumbel_time.pdf}
        \includegraphics[width=\thirdwidth]{figures/dagma_img/d=1000_ER4_5000_gaussian_ev_iters_dagma.pdf}
        \includegraphics[width=\thirdwidth]{figures/dagma_img/d=1000_ER4_5000_exponential_iters_dagma.pdf}
        \includegraphics[width=\thirdwidth]{figures/dagma_img/d=1000_ER4_5000_gumbel_iters_dagma.pdf}
        \caption{$d=1000$ vertices}
    \end{subfigure}

     \begin{subfigure}{\textwidth}
        \includegraphics[width=\thirdwidth]{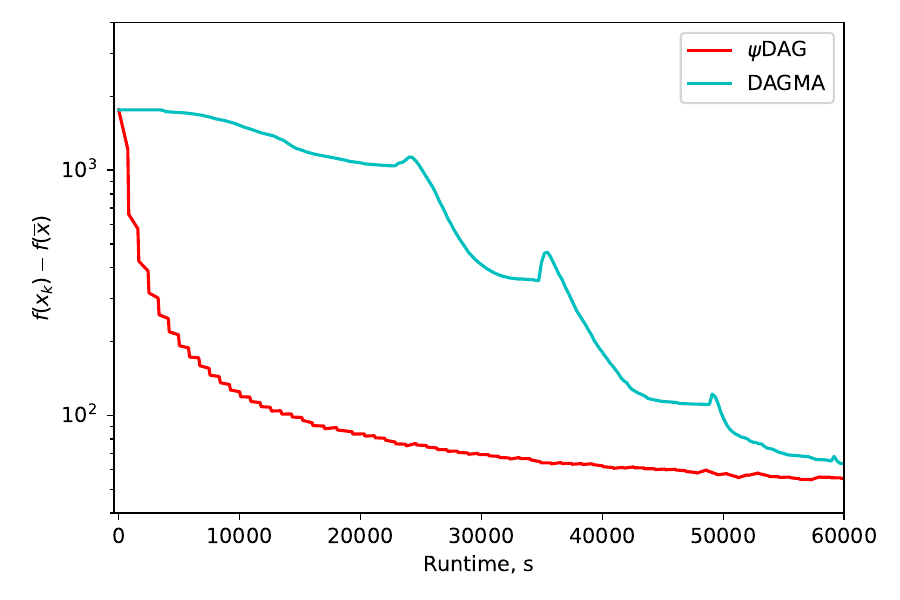}
        \includegraphics[width=\thirdwidth]{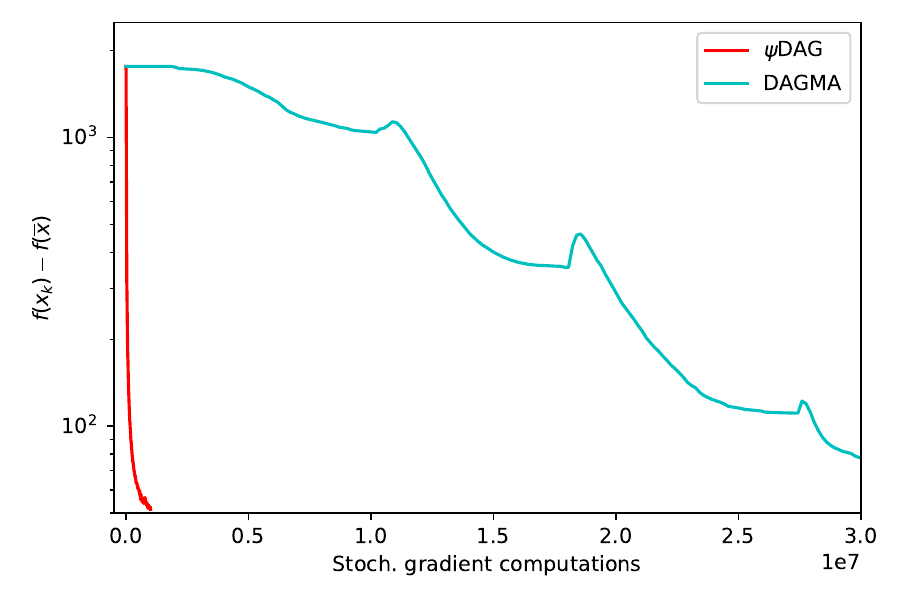}
        \caption{$d=3000$ vertices}
    \end{subfigure}

    \caption{Linear SEM methods on graphs of type ER4 with different noise distributions: Gaussian (first), exponential (second), Gumbel (third).}
    \label{fig:er4_medium}
\end{figure*}

\textbf{ER graph types:} Figures \ref{fig:er2_small} and \ref{fig:er2_medium} show performance on ER2 graphs, Figures \ref{fig:er4_small} and \ref{fig:er4_medium} show performance on ER4 graphs,
\Cref{fig:er6} shows performance on ER6 graphs.

\begin{figure*}
    \centering
    \begin{subfigure}{\textwidth}
        \centering
        \includegraphics[width=\thirdwidth]{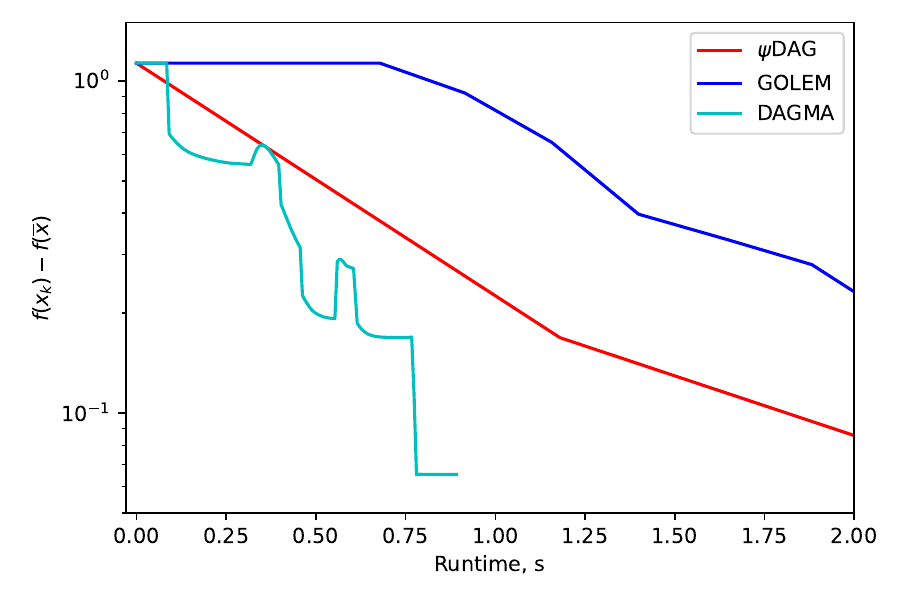}
        \includegraphics[width=\thirdwidth]{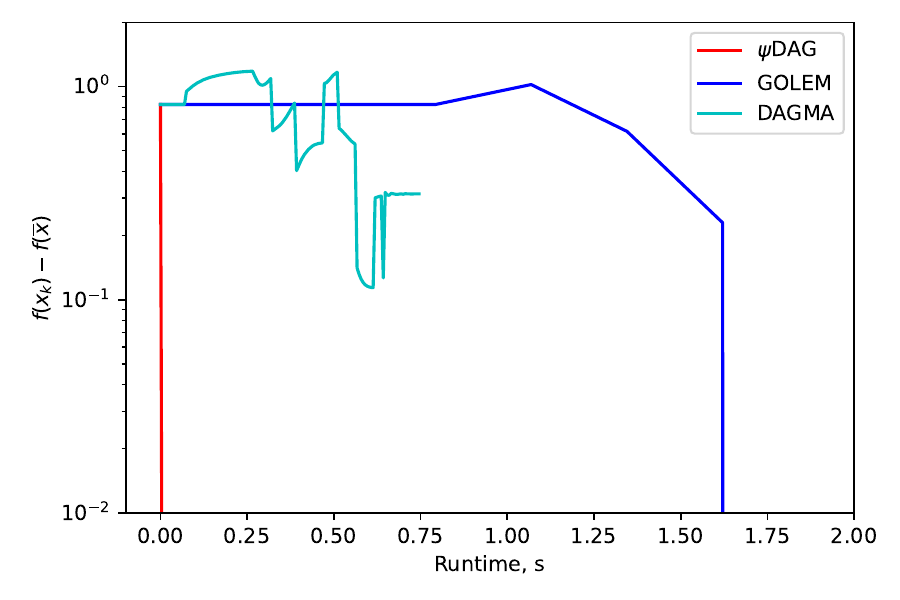}
        \includegraphics[width=\thirdwidth]{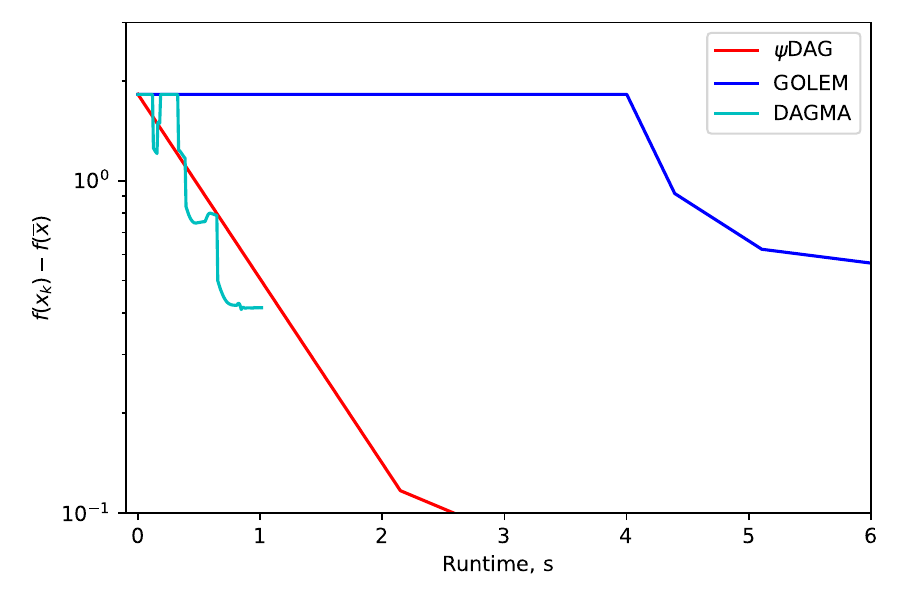}
        \includegraphics[width=\thirdwidth]{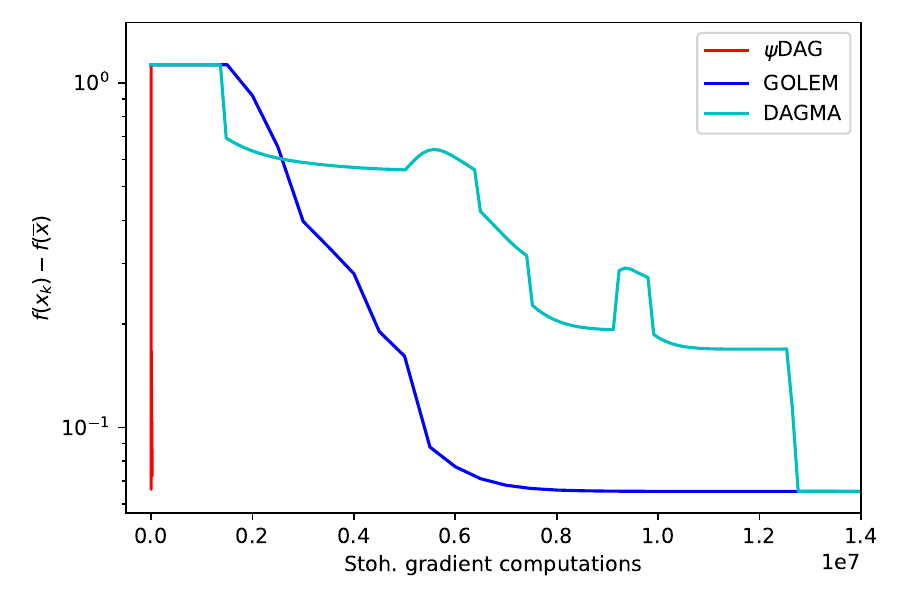}
        \includegraphics[width=\thirdwidth]{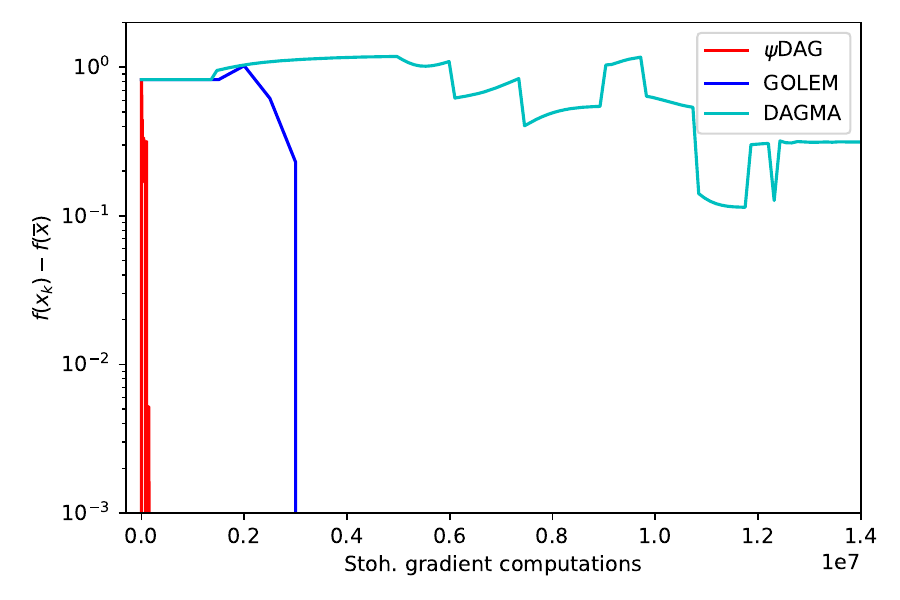}
        \includegraphics[width=\thirdwidth]{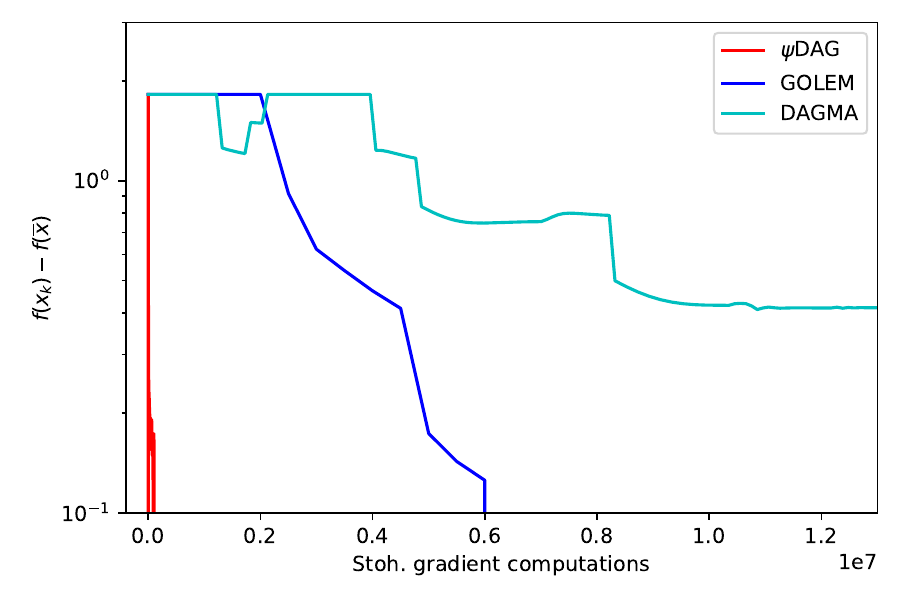}
        \caption{$d=10$ vertices}
    \end{subfigure}

    \begin{subfigure}{\textwidth}
        \centering
        \includegraphics[width=\thirdwidth]{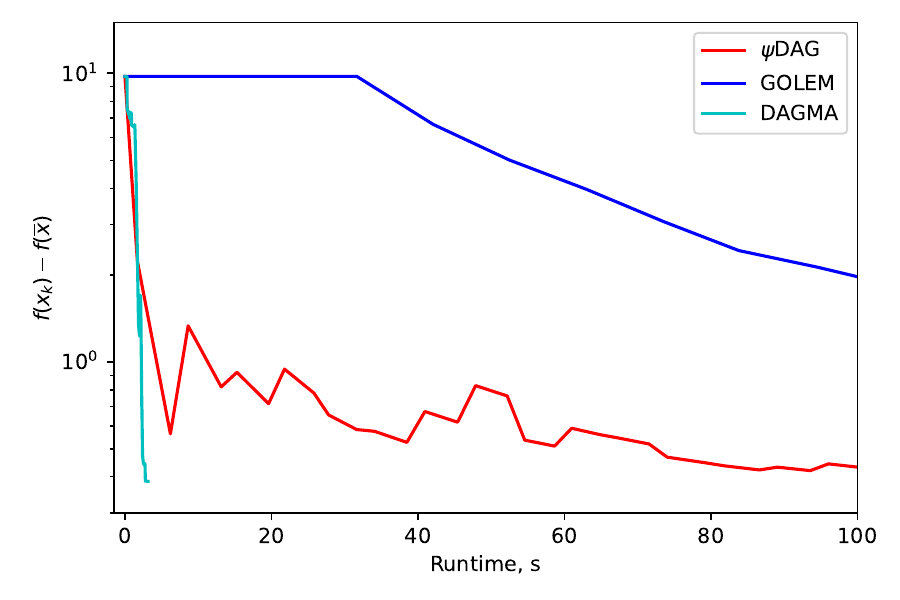}
        \includegraphics[width=\thirdwidth]{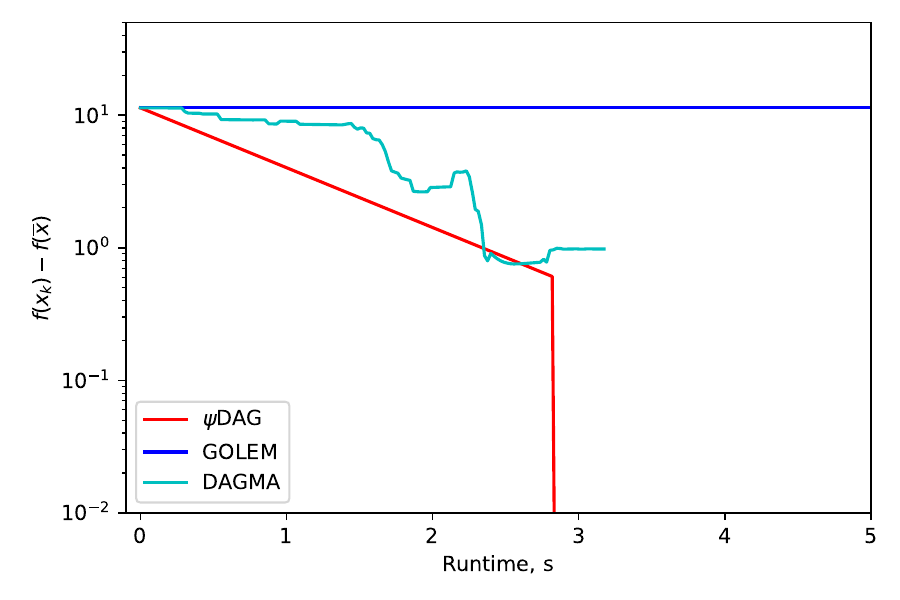}
        \includegraphics[width=\thirdwidth]{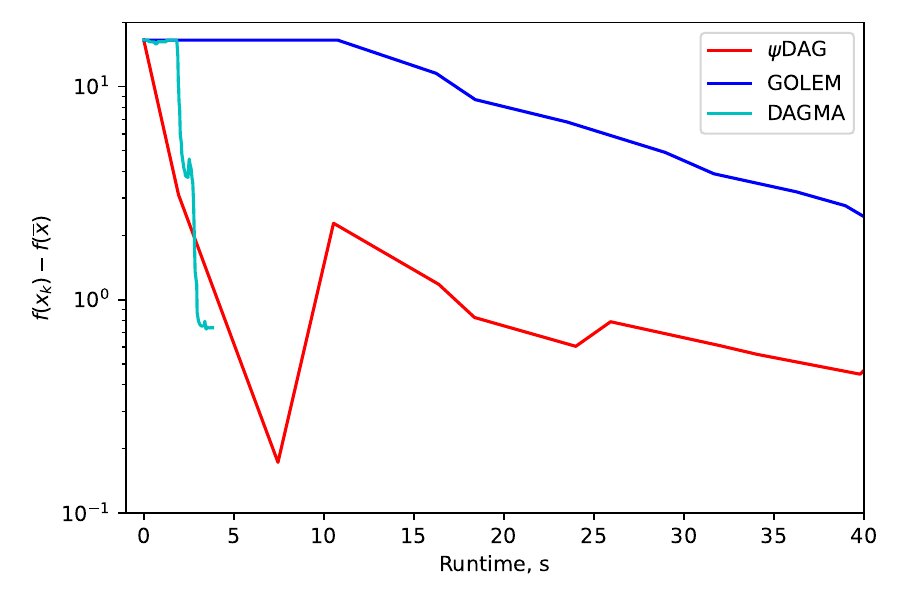}
        \includegraphics[width=\thirdwidth]{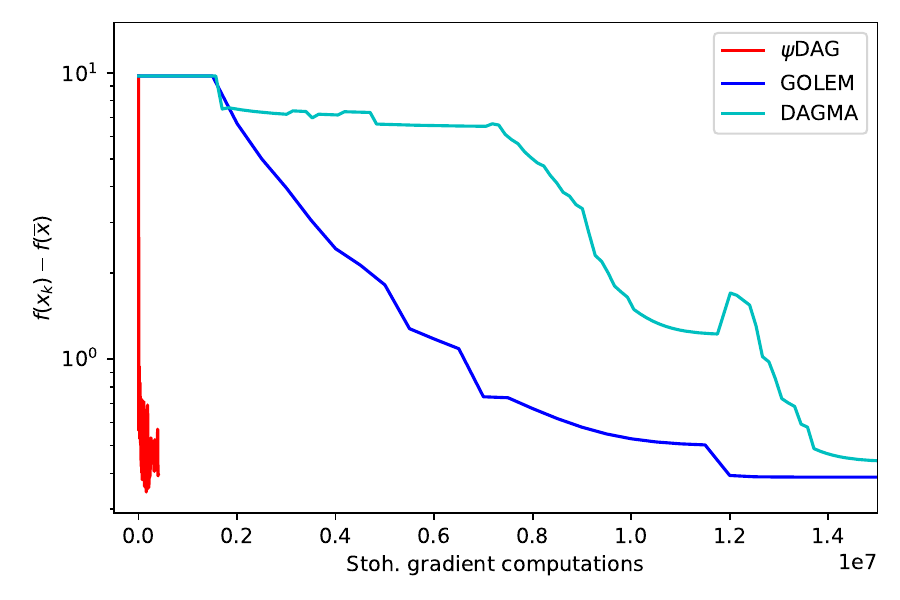}
        \includegraphics[width=\thirdwidth]{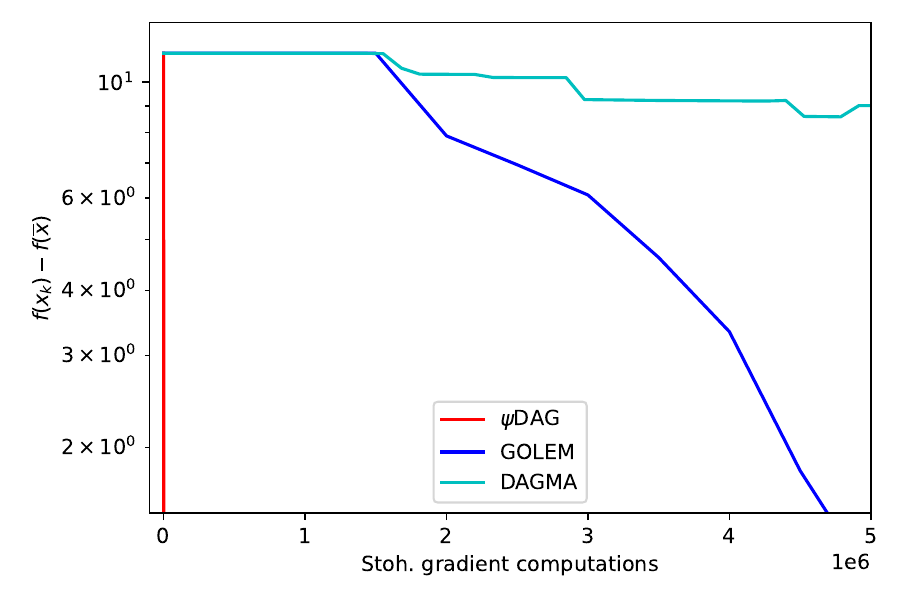}
        \includegraphics[width=\thirdwidth]{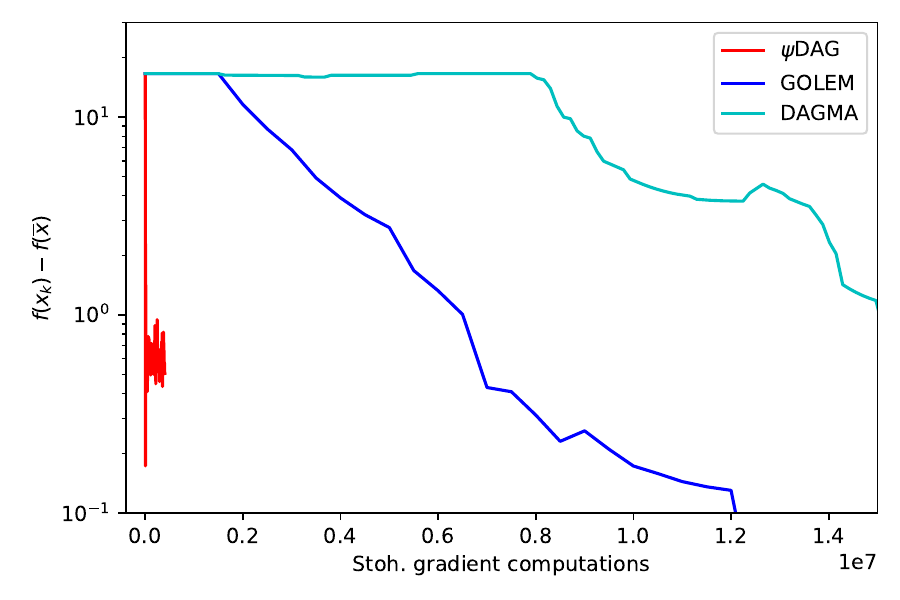}
        \caption{$d=50$ vertices}
    \end{subfigure}

    \begin{subfigure}{\textwidth}
        \centering
        \includegraphics[width=\thirdwidth]{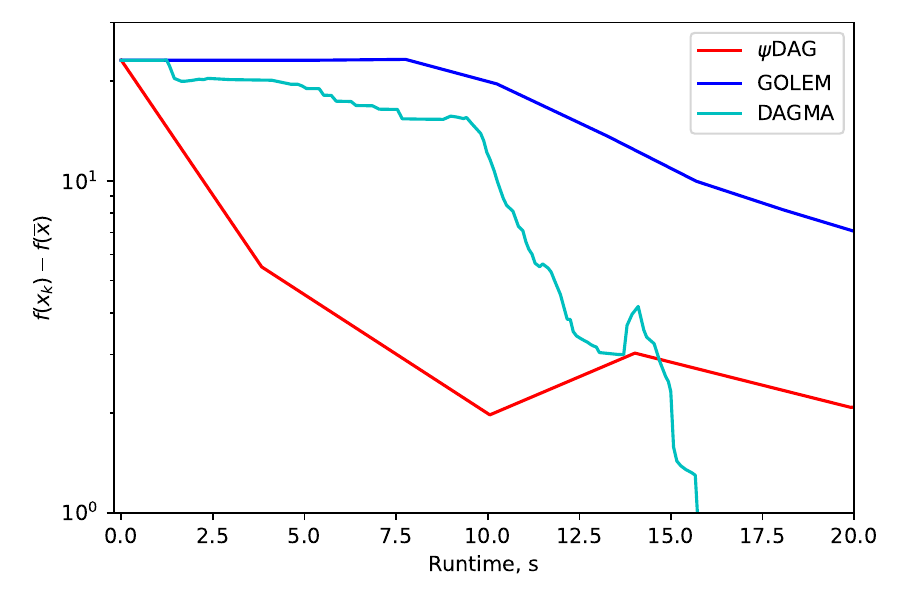}
        \includegraphics[width=\thirdwidth]{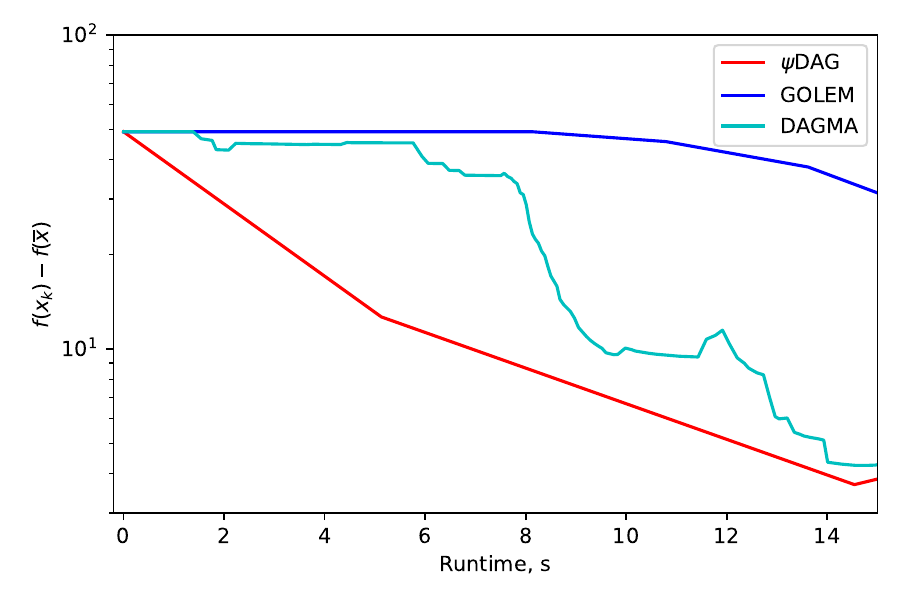}
        \includegraphics[width=\thirdwidth]{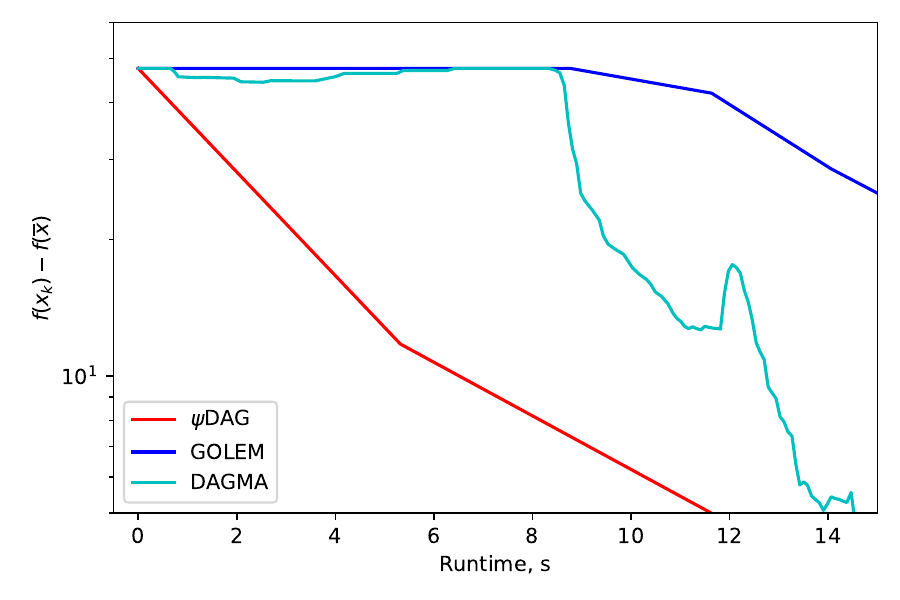}
        \includegraphics[width=\thirdwidth]{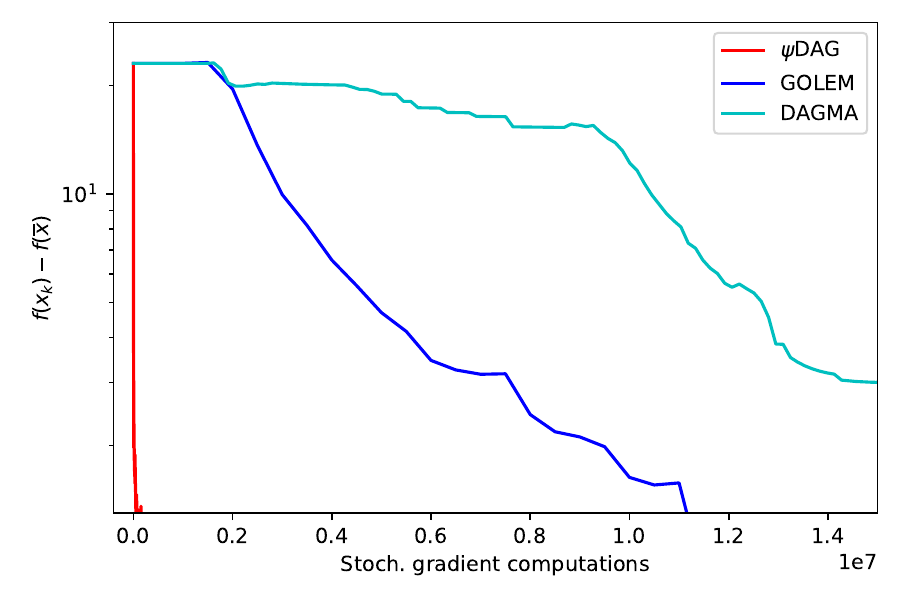}
        \includegraphics[width=\thirdwidth]{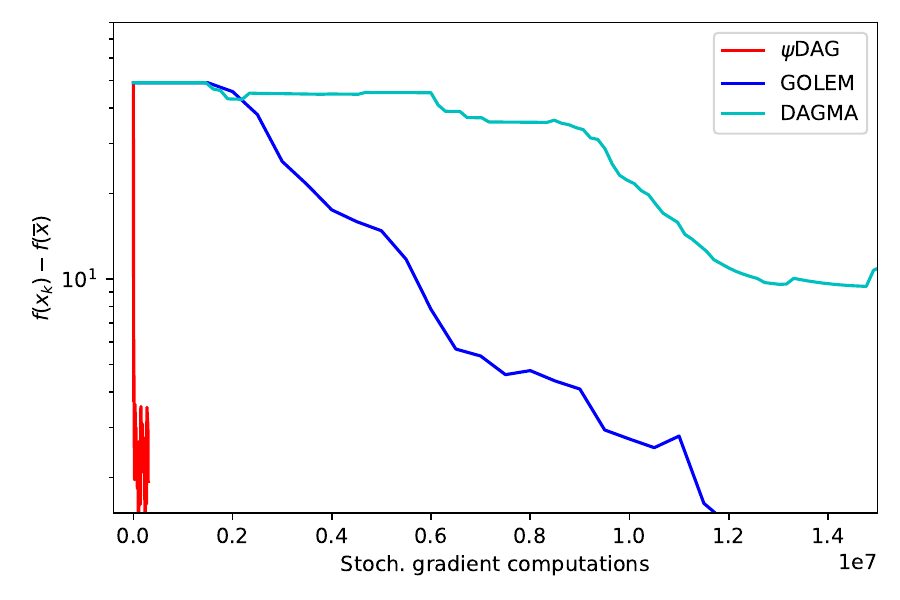}
        \includegraphics[width=\thirdwidth]{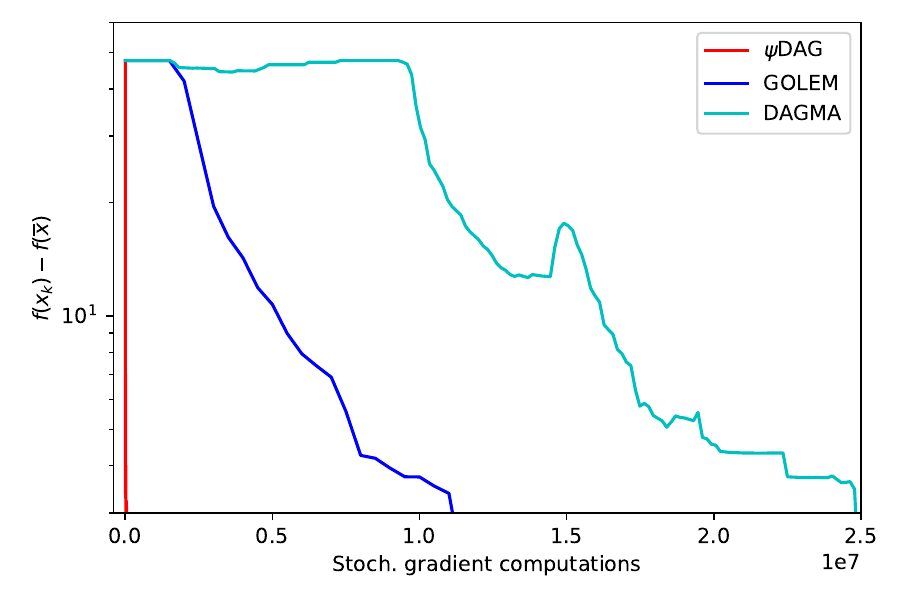}
        \caption{$d=100$ vertices}
    \end{subfigure}

    \caption{Linear SEM methods on graphs of type SF2 with different noise distributions: Gaussian (first), exponential (second), Gumbel (third).}
    \label{fig:sf2_small}
\end{figure*}

\begin{figure*}
    \centering
    
    \begin{subfigure}{\textwidth}
        \centering
        \includegraphics[width=\thirdwidth]{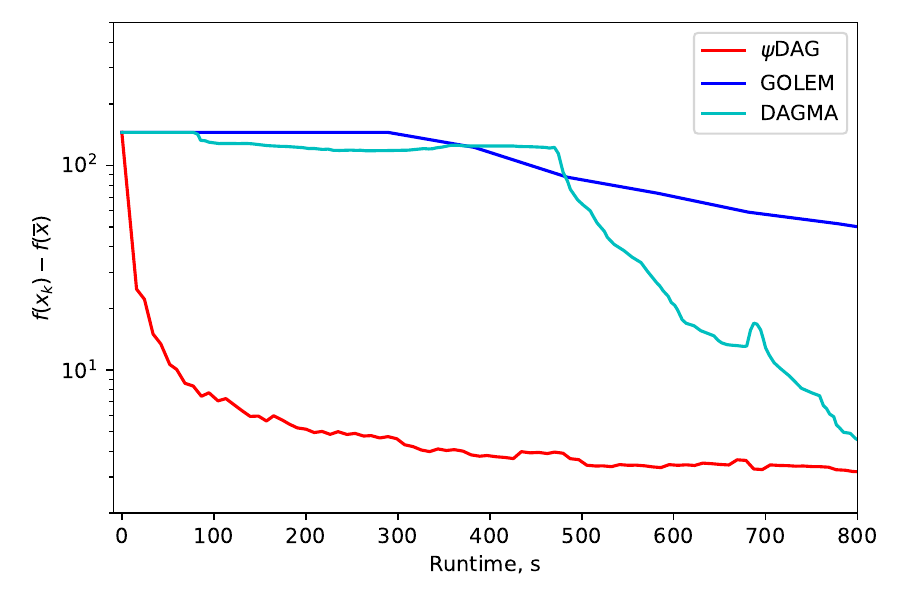}
        \includegraphics[width=\thirdwidth]{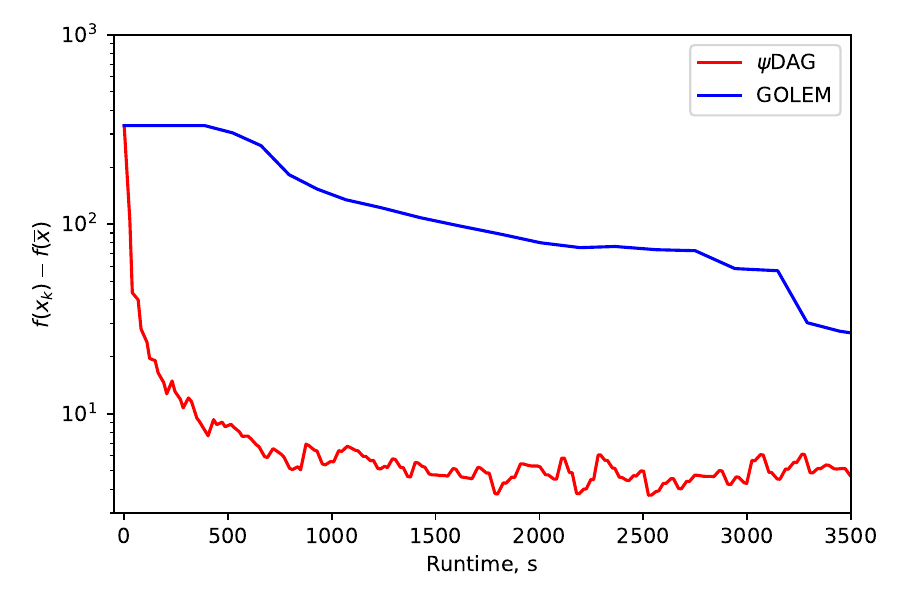}
        \includegraphics[width=\thirdwidth]{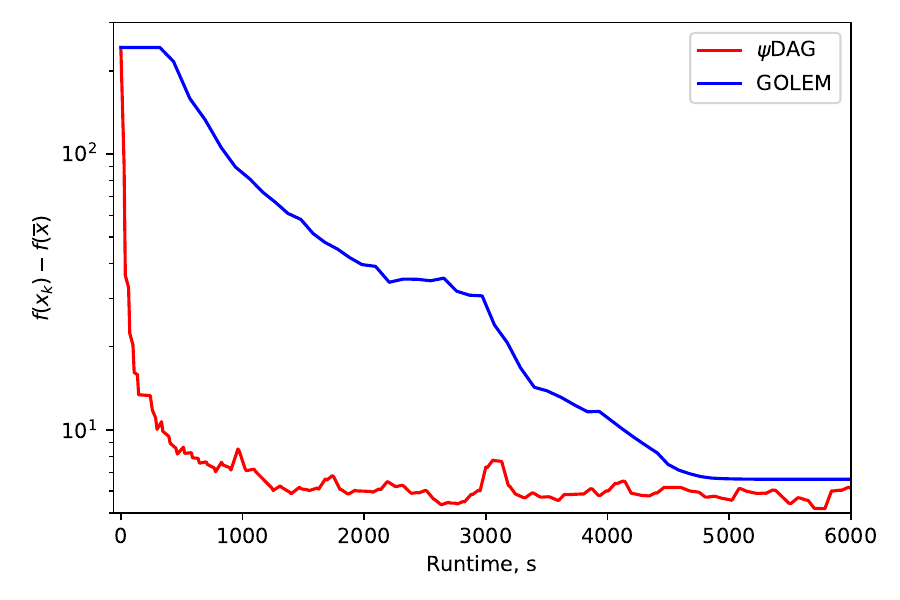}
        \includegraphics[width=\thirdwidth]{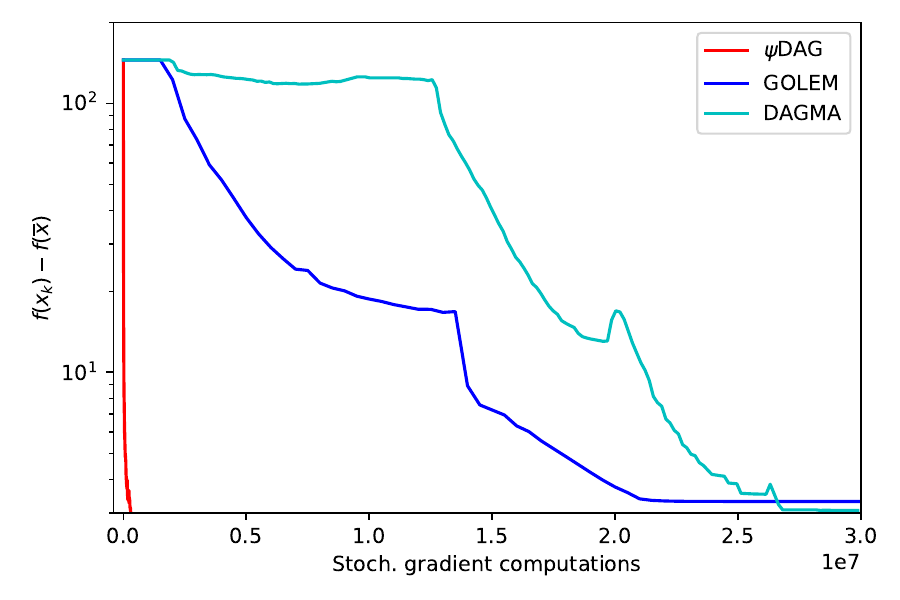}
        \includegraphics[width=\thirdwidth]{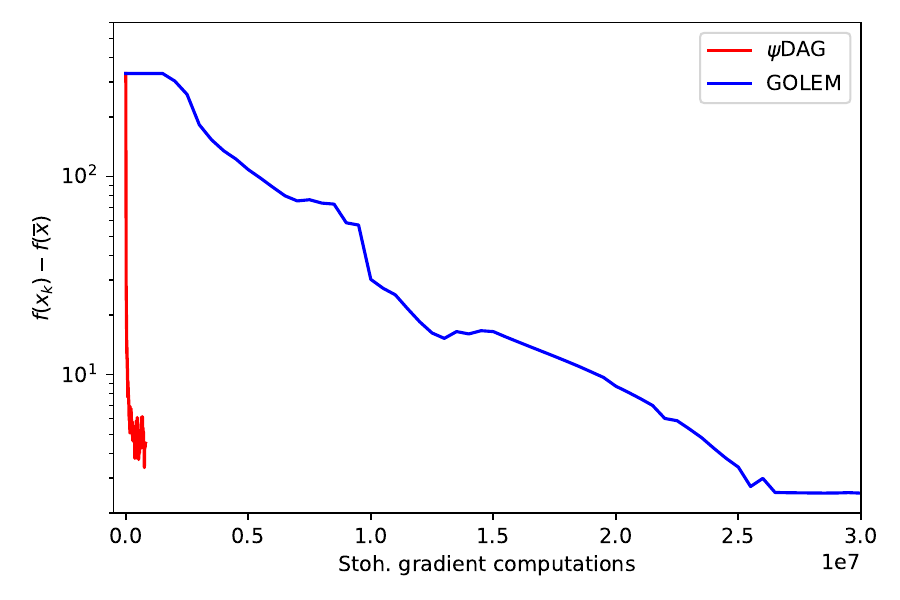}
        \includegraphics[width=\thirdwidth]{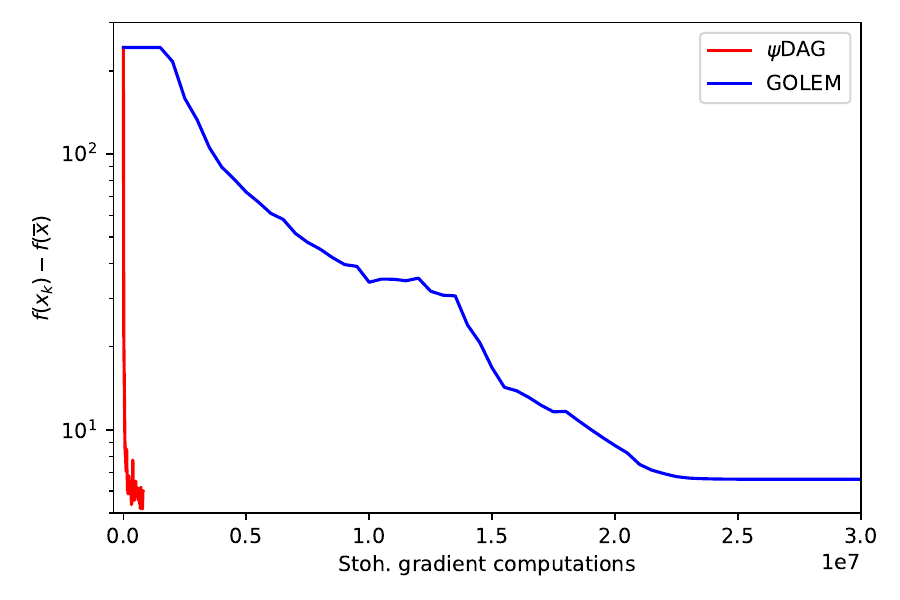}
        \caption{$d=500$ vertices}
    \end{subfigure}

    \begin{subfigure}{\textwidth}
        \centering
        \includegraphics[width=\thirdwidth]{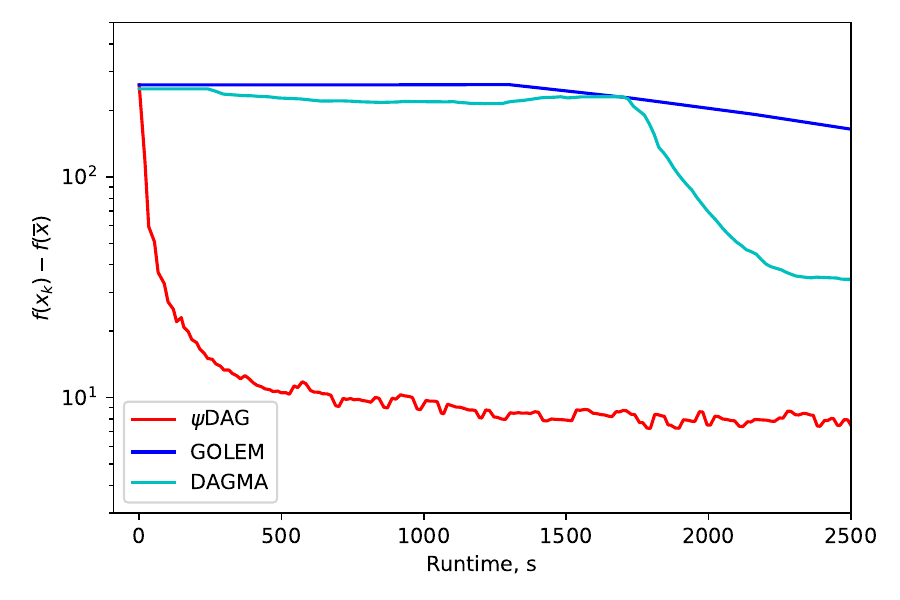}
        \includegraphics[width=\thirdwidth]{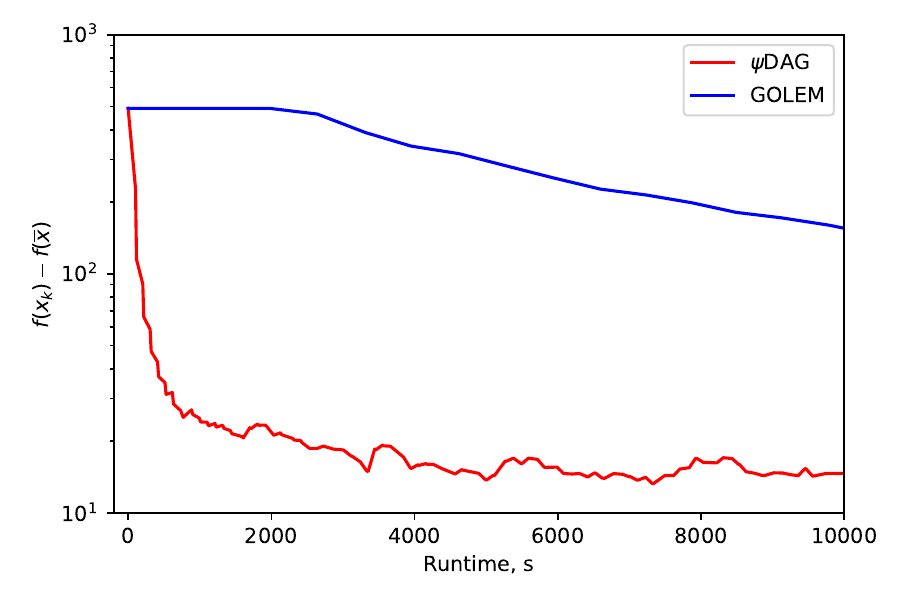}
        \includegraphics[width=\thirdwidth]{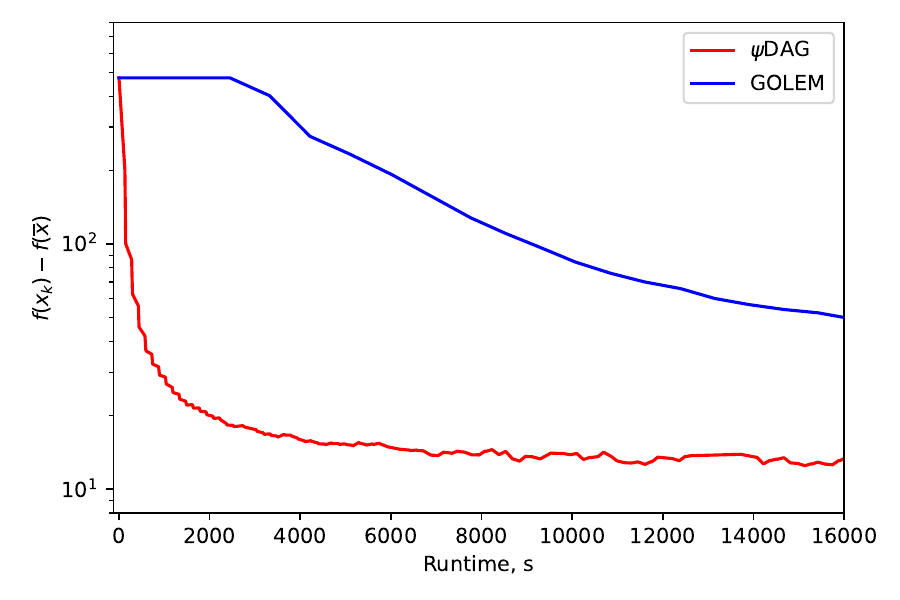}
        \includegraphics[width=\thirdwidth]{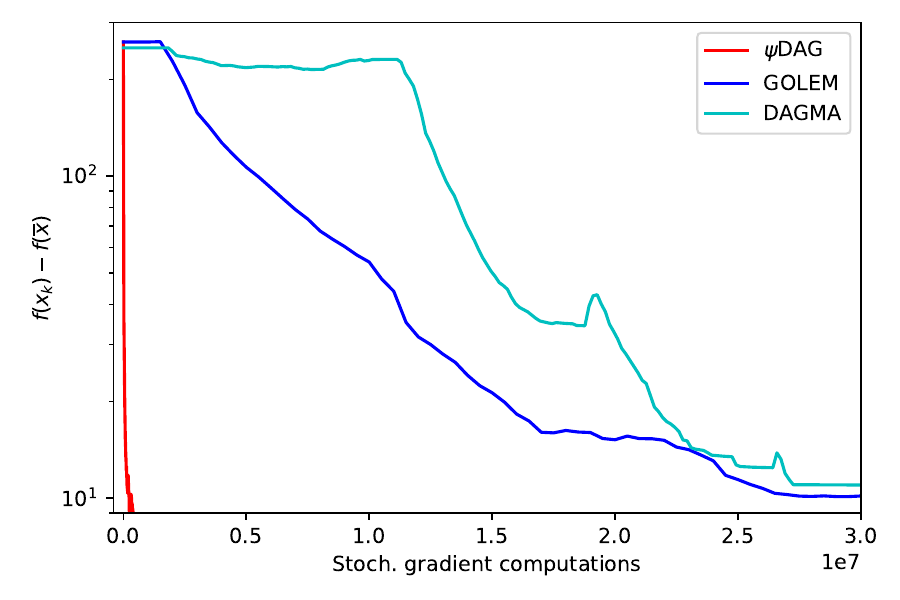}
        \includegraphics[width=\thirdwidth]{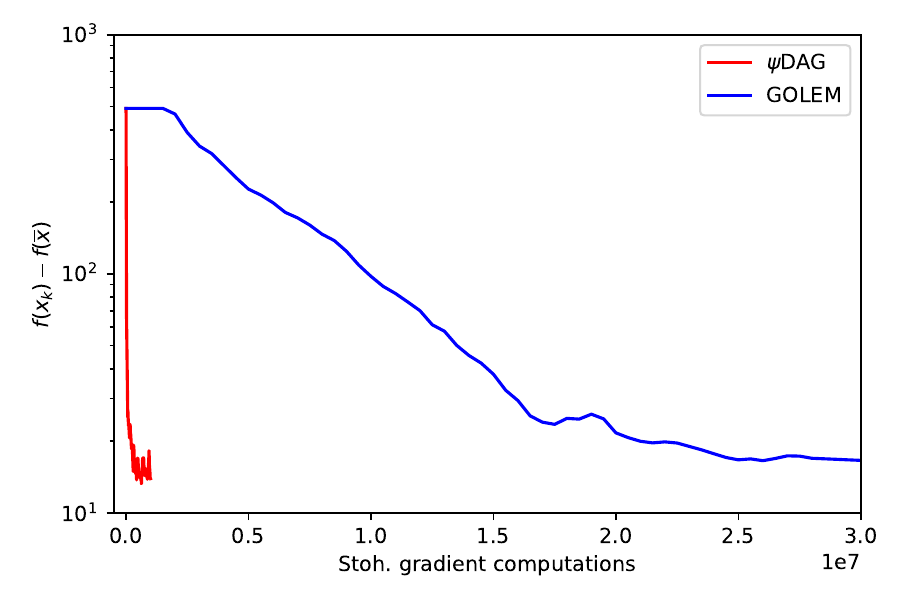}
        \includegraphics[width=\thirdwidth]{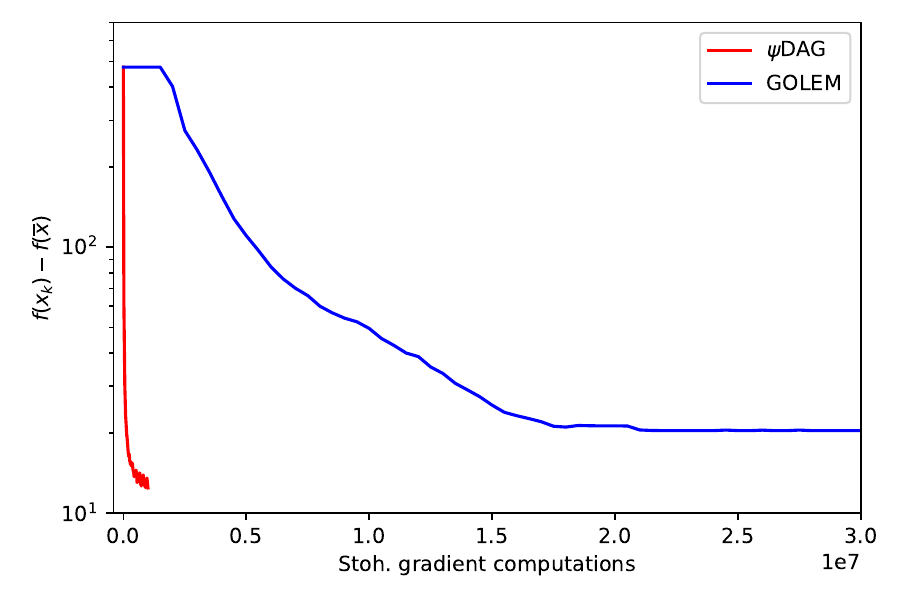}
        \caption{$d=1000$ vertices}
    \end{subfigure}

    \begin{subfigure}{\textwidth}
        \includegraphics[width=\thirdwidth]{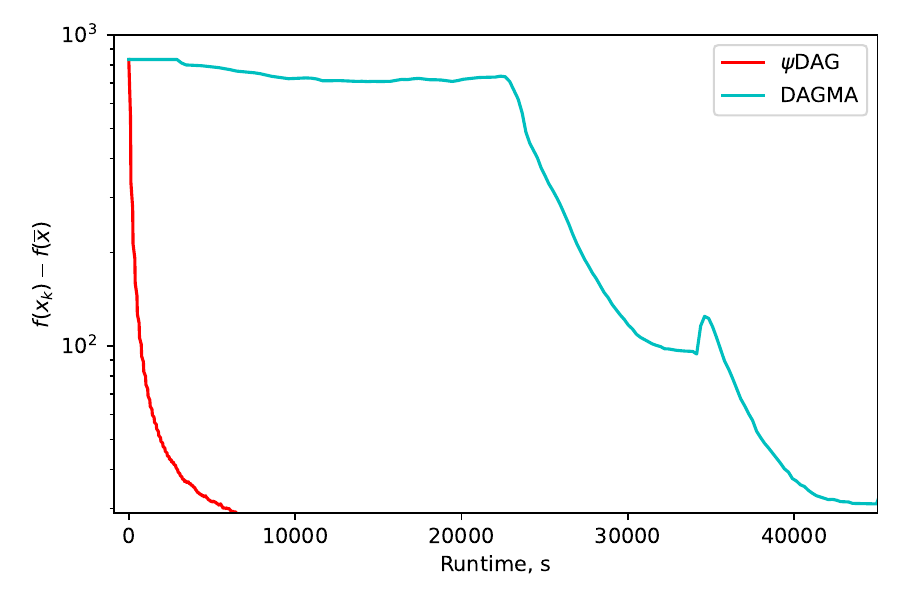}
        \includegraphics[width=\thirdwidth]{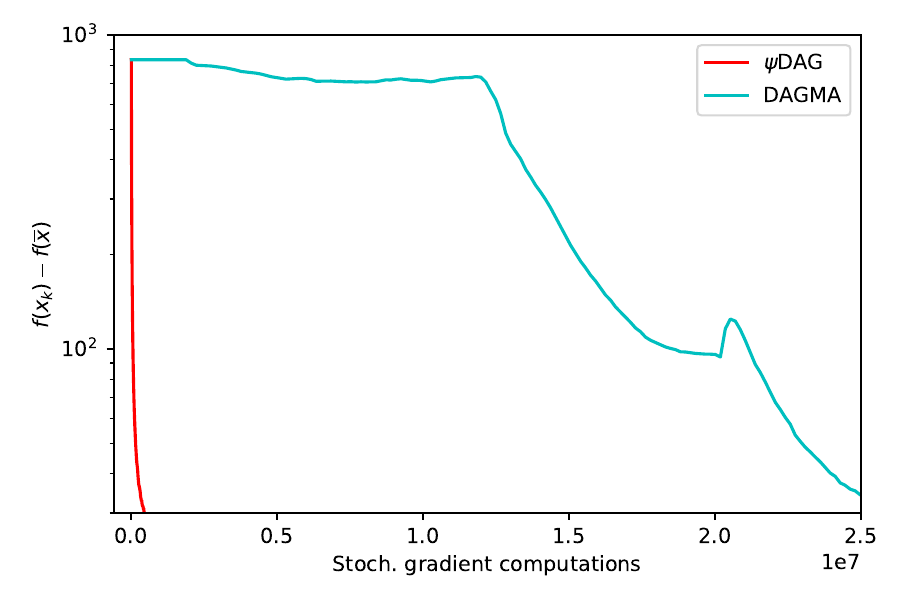}
        \caption{$d=3000$ vertices}
    \end{subfigure}

    \caption{Linear SEM methods on graphs of type SF2 with different noise distributions: Gaussian (first), exponential (second), Gumbel (third).}
    \label{fig:sf2_medium}
\end{figure*}

\begin{figure*}
    \centering
    \begin{subfigure}{\textwidth}
        \centering
        \includegraphics[width=\thirdwidth]{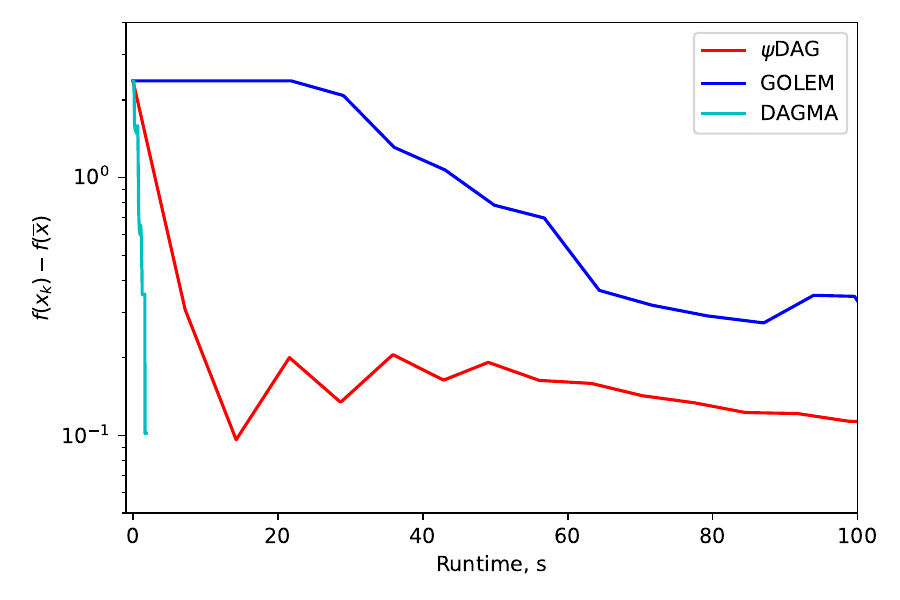}
        \includegraphics[width=\thirdwidth]{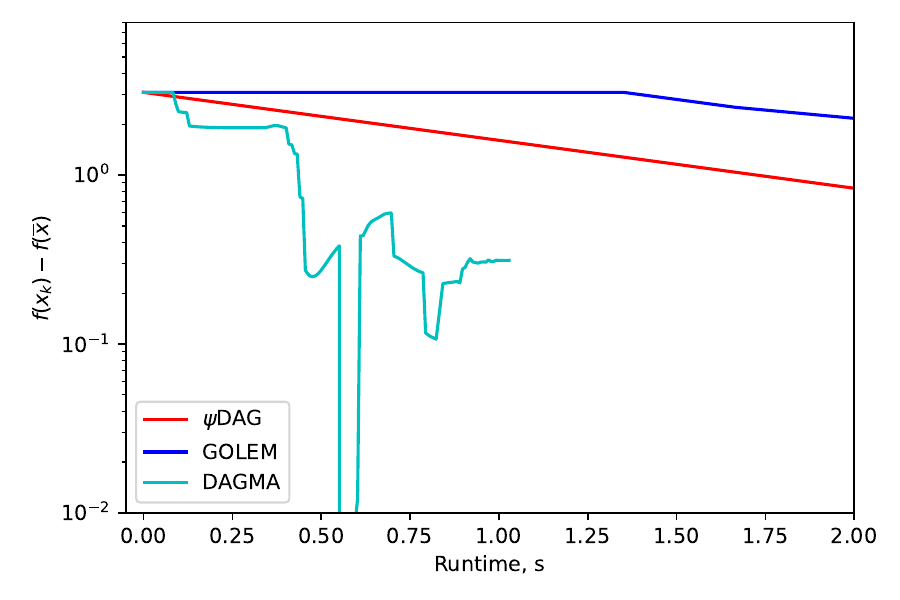}
        \includegraphics[width=\thirdwidth]{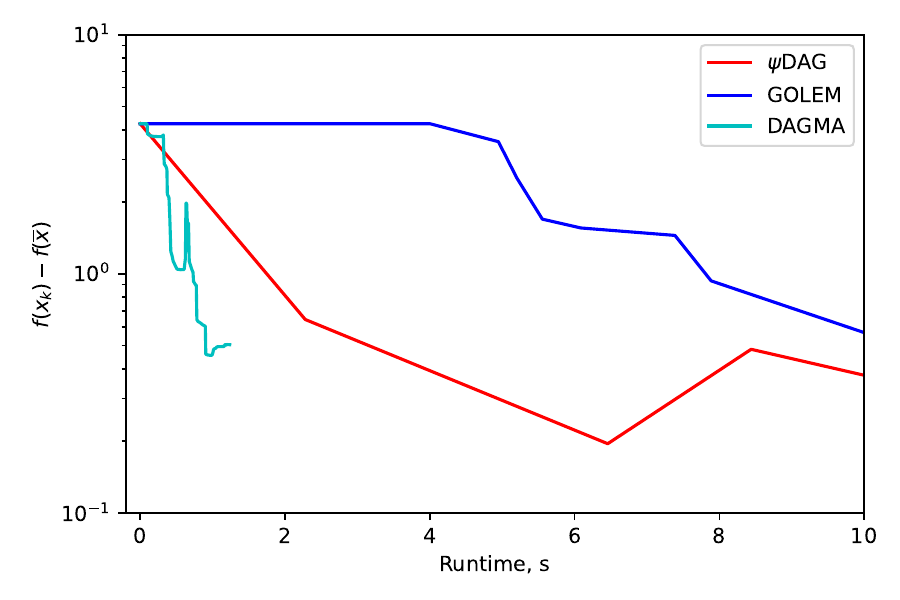}
        \includegraphics[width=\thirdwidth]{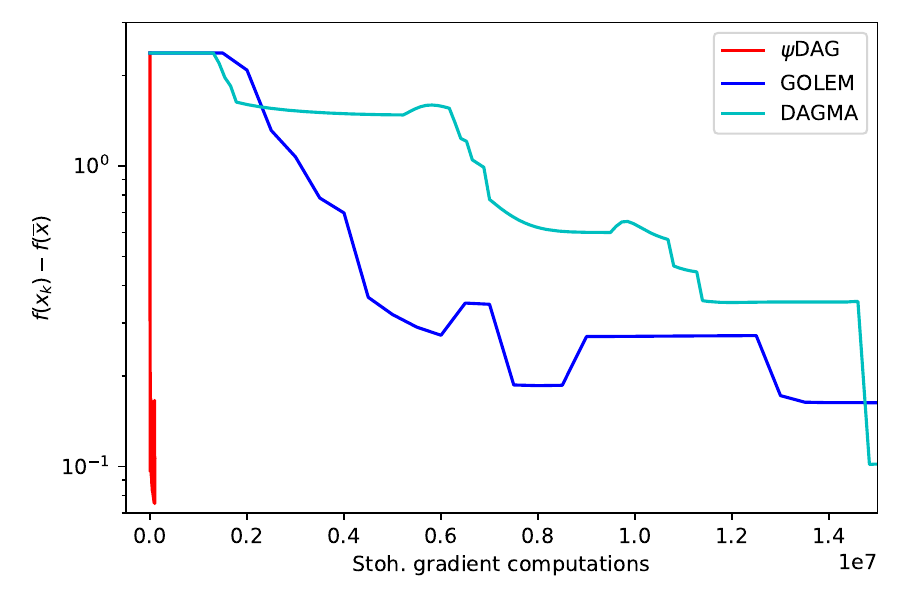}
        \includegraphics[width=\thirdwidth]{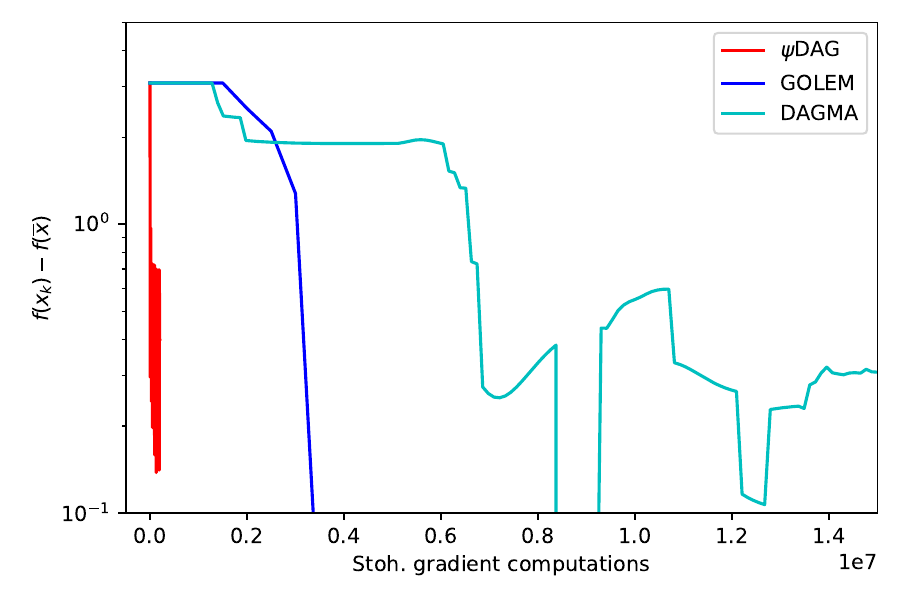}
        \includegraphics[width=\thirdwidth]{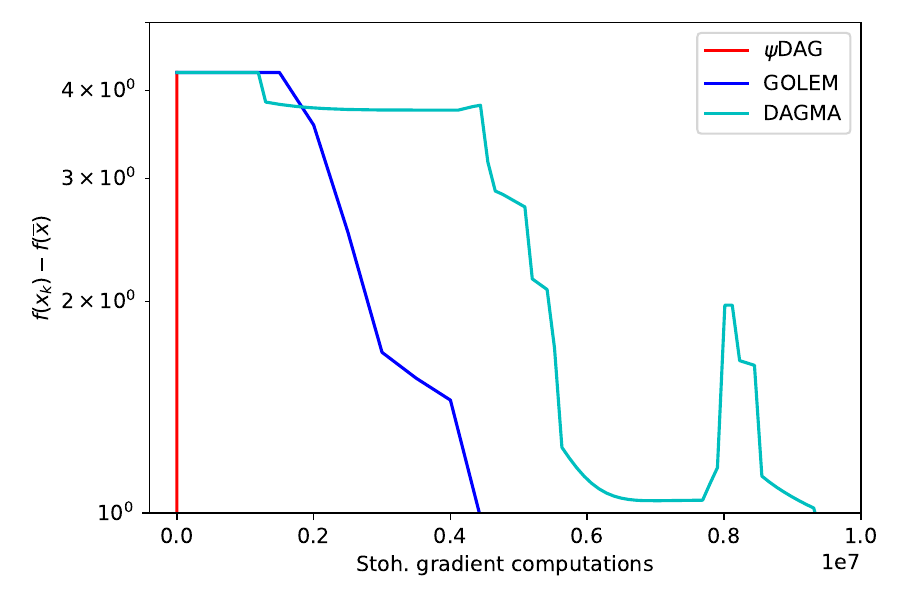}
        \caption{$d=10$ vertices}
    \end{subfigure}

    \begin{subfigure}{\textwidth}
        \centering
        \includegraphics[width=\thirdwidth]{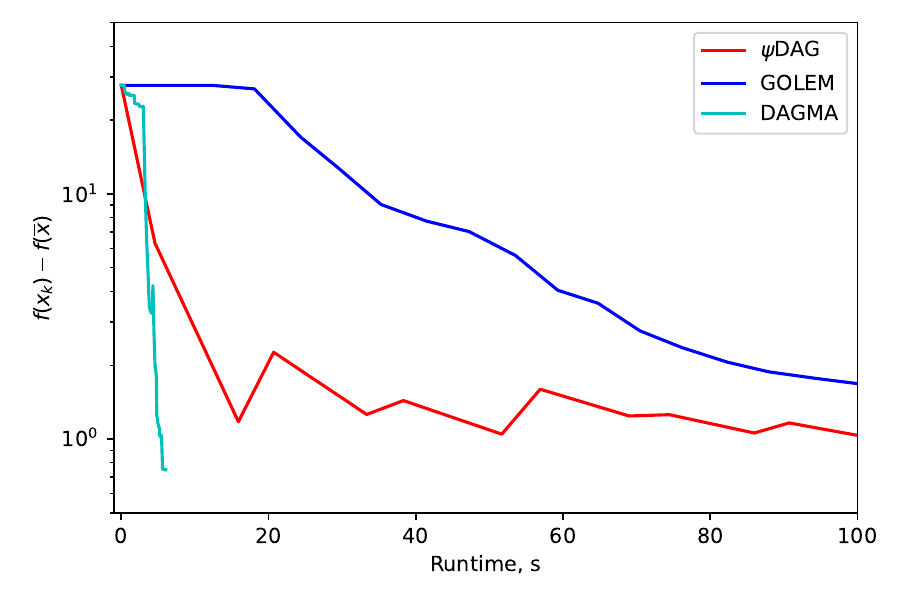}
        \includegraphics[width=\thirdwidth]{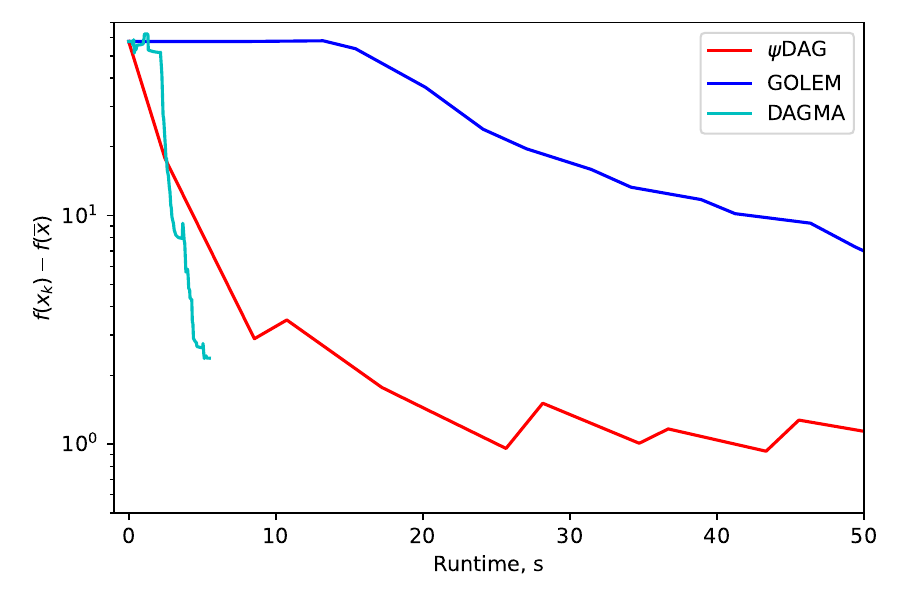}
        \includegraphics[width=\thirdwidth]{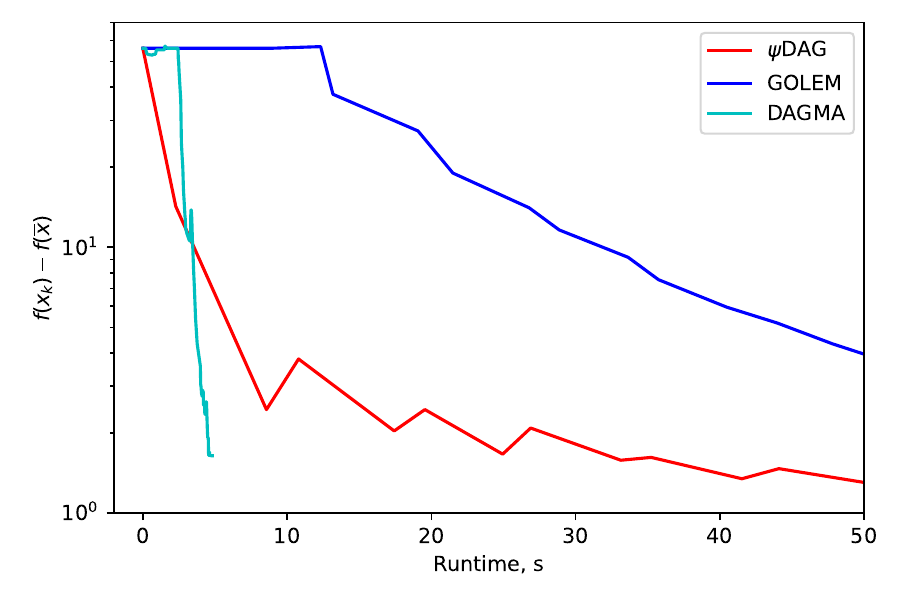}
        \includegraphics[width=\thirdwidth]{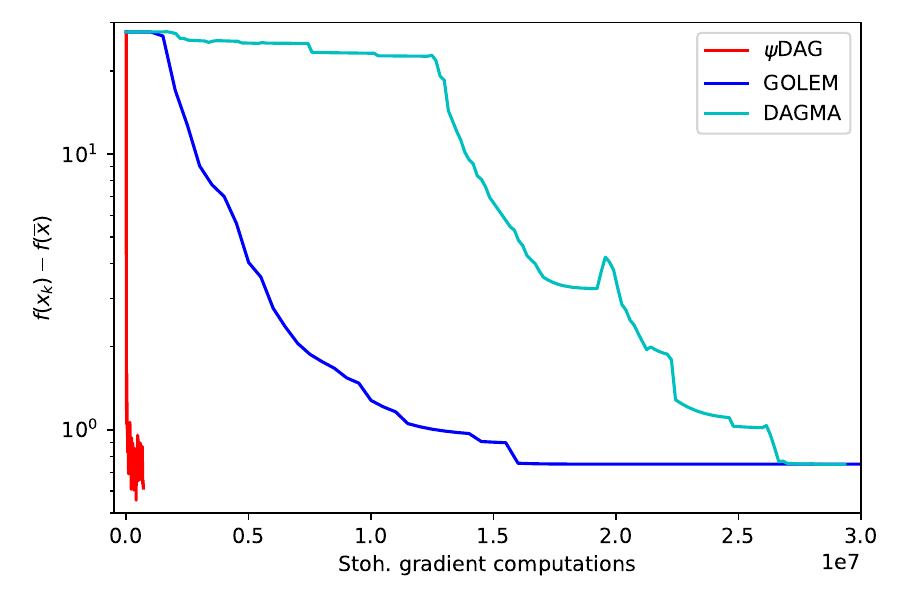}
        \includegraphics[width=\thirdwidth]{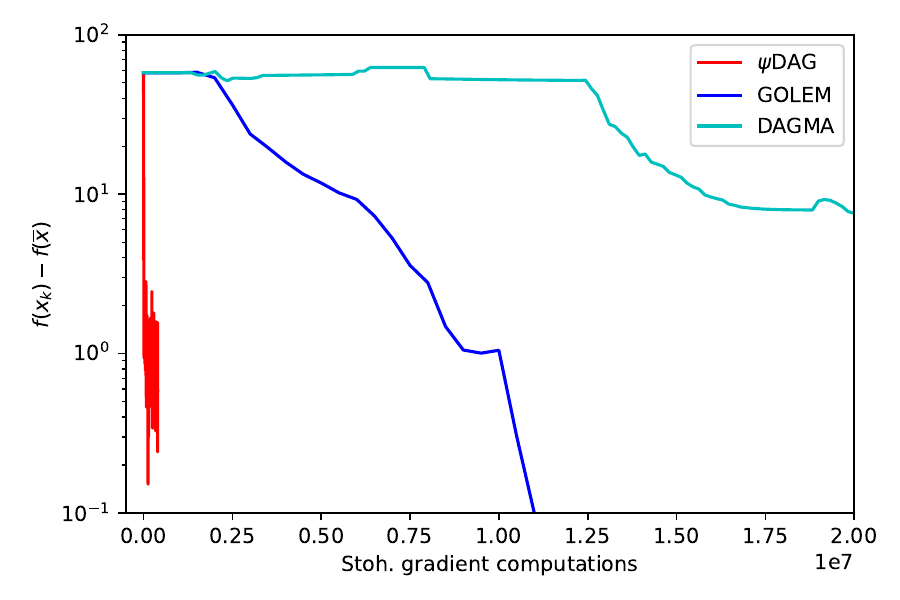}
        \includegraphics[width=\thirdwidth]{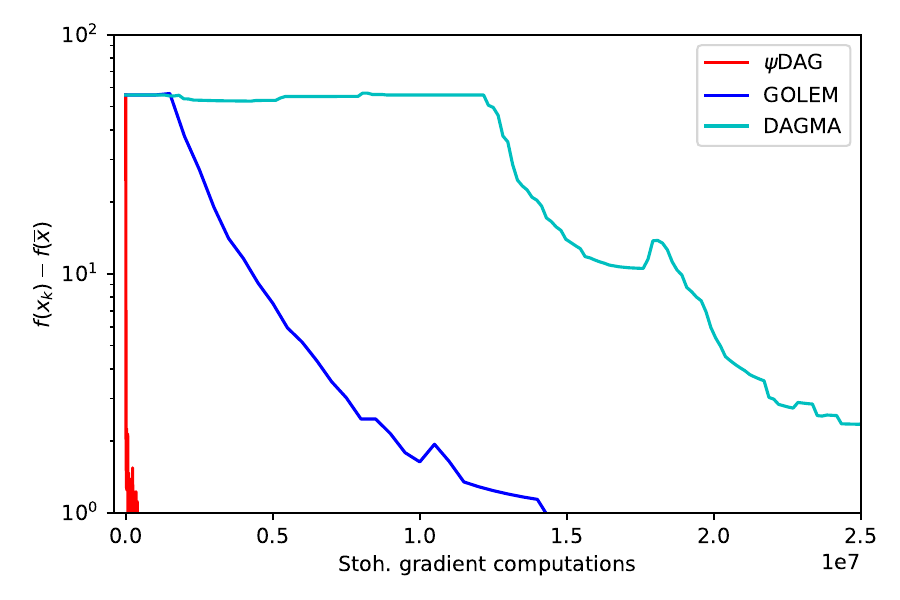}
        \caption{$d=50$ vertices}
    \end{subfigure}

    \begin{subfigure}{\textwidth}
        \centering
        \includegraphics[width=\thirdwidth]{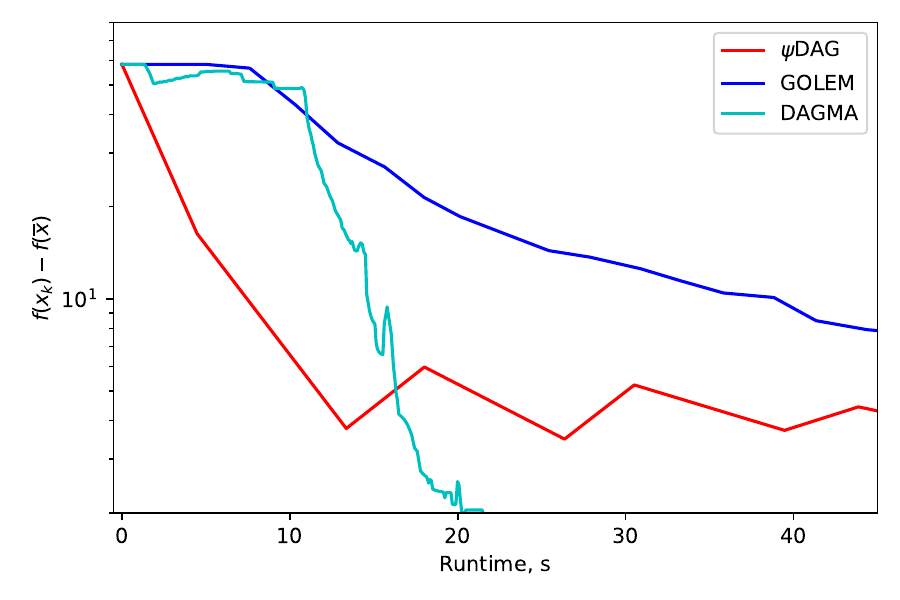}
        \includegraphics[width=\thirdwidth]{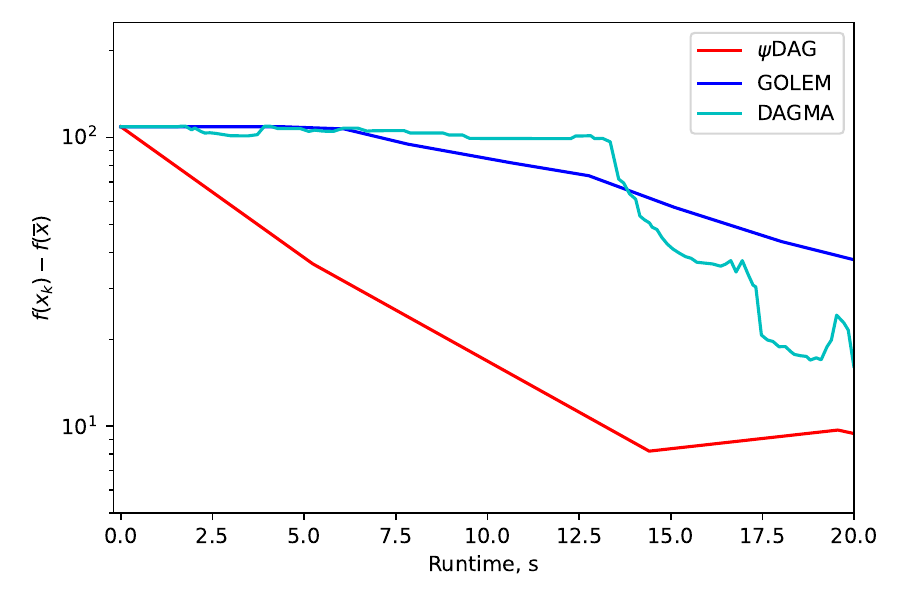}
        \includegraphics[width=\thirdwidth]{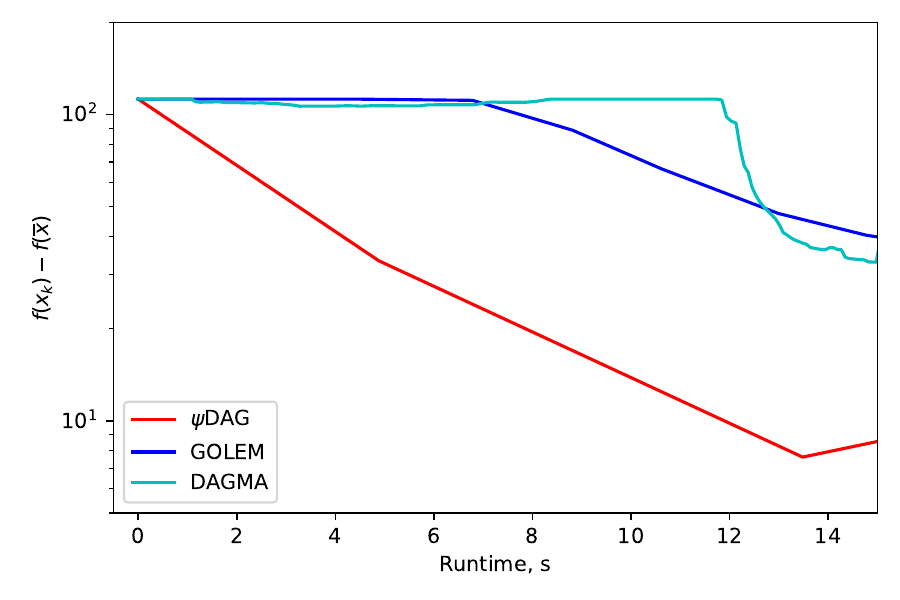}
        \includegraphics[width=\thirdwidth]{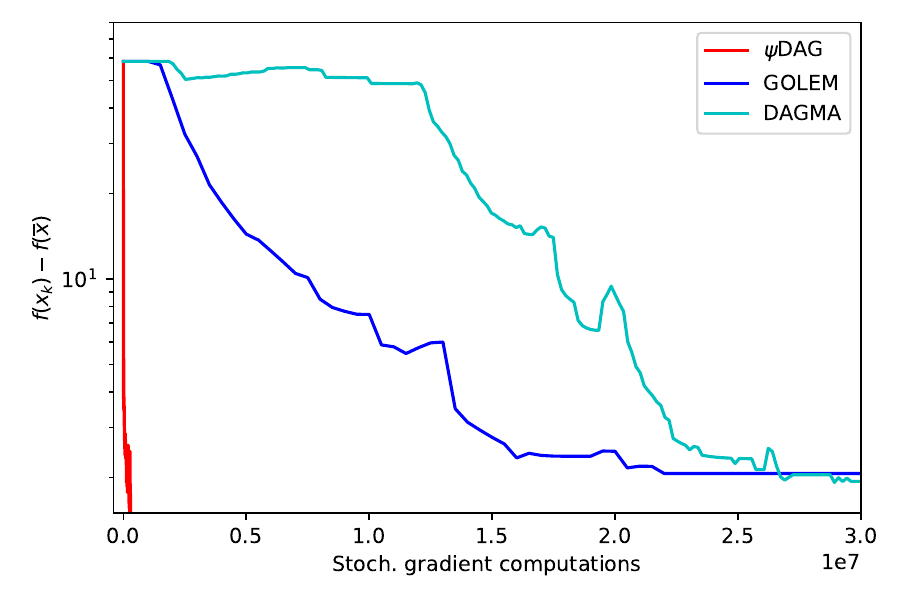}
        \includegraphics[width=\thirdwidth]{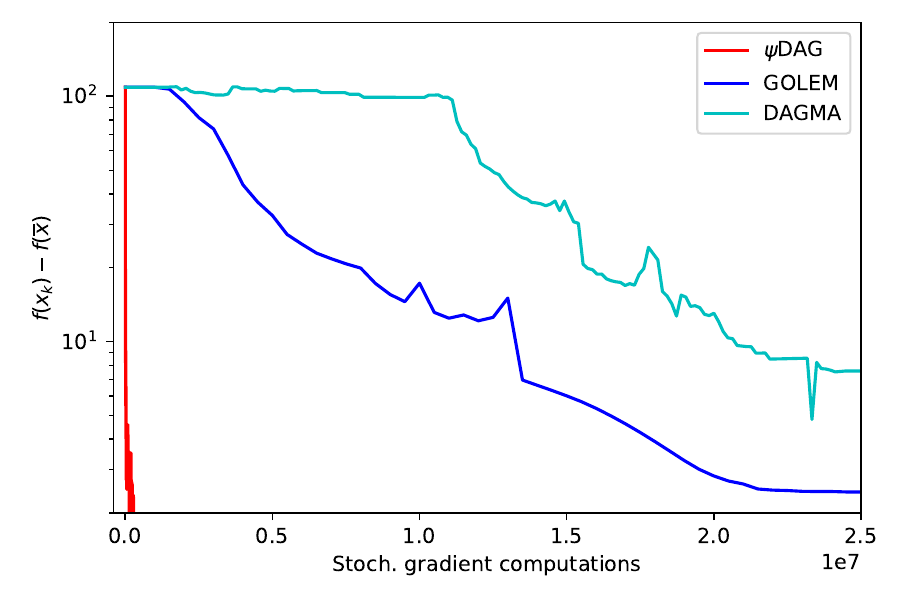}
        \includegraphics[width=\thirdwidth]{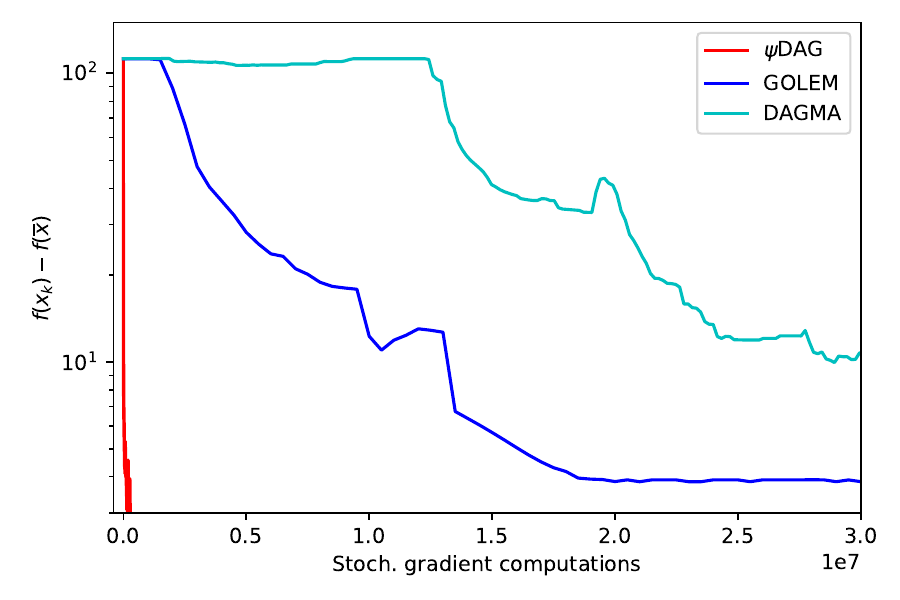}
        \caption{$d=100$ vertices}
    \end{subfigure}

    \caption{Linear SEM methods on graphs of type SF4 with different noise distributions: Gaussian (first), exponential (second), Gumbel (third).}
    \label{fig:sf4_small}
\end{figure*}

\begin{figure*}
    \centering
    
    \begin{subfigure}{\textwidth}
        \centering
        \includegraphics[width=\thirdwidth]{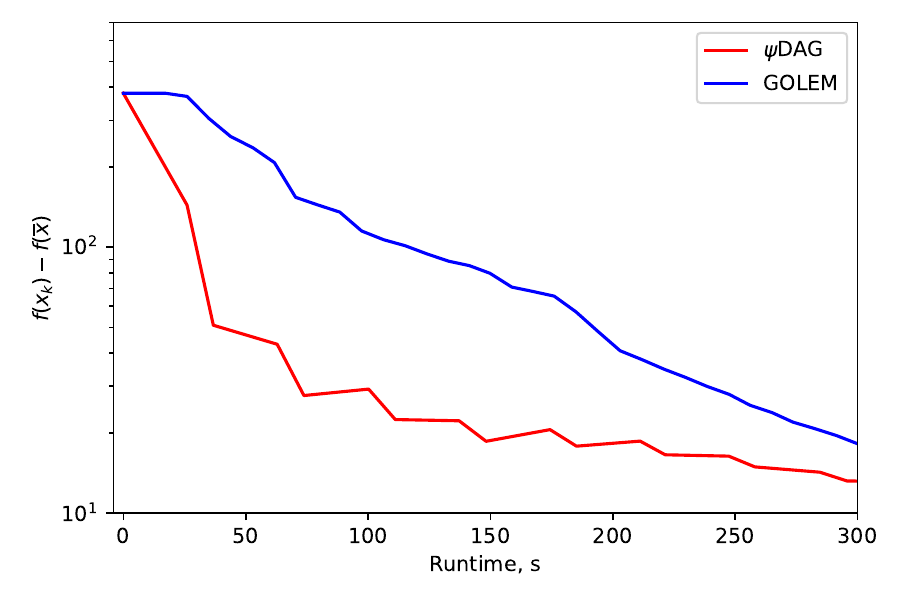}
        \includegraphics[width=\thirdwidth]{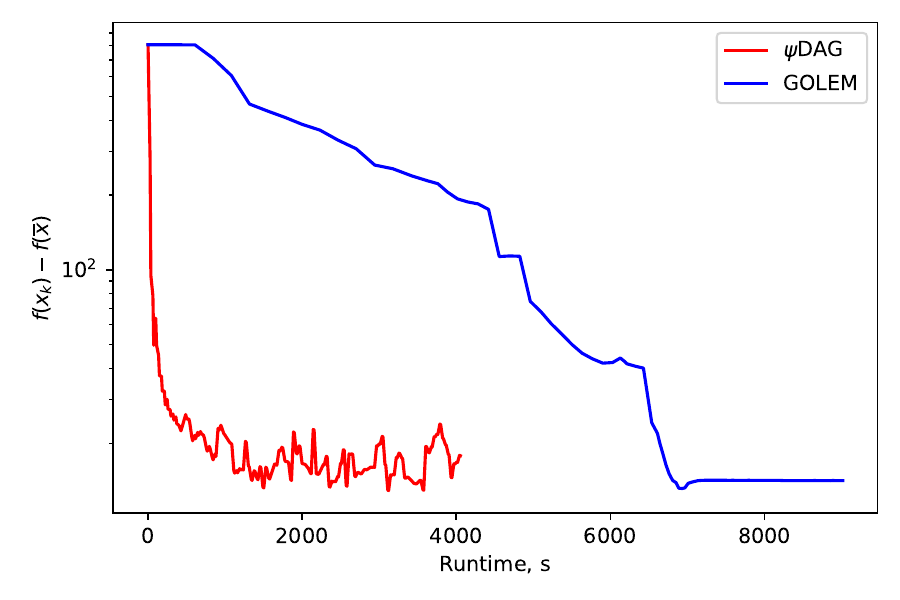}
        \includegraphics[width=\thirdwidth]{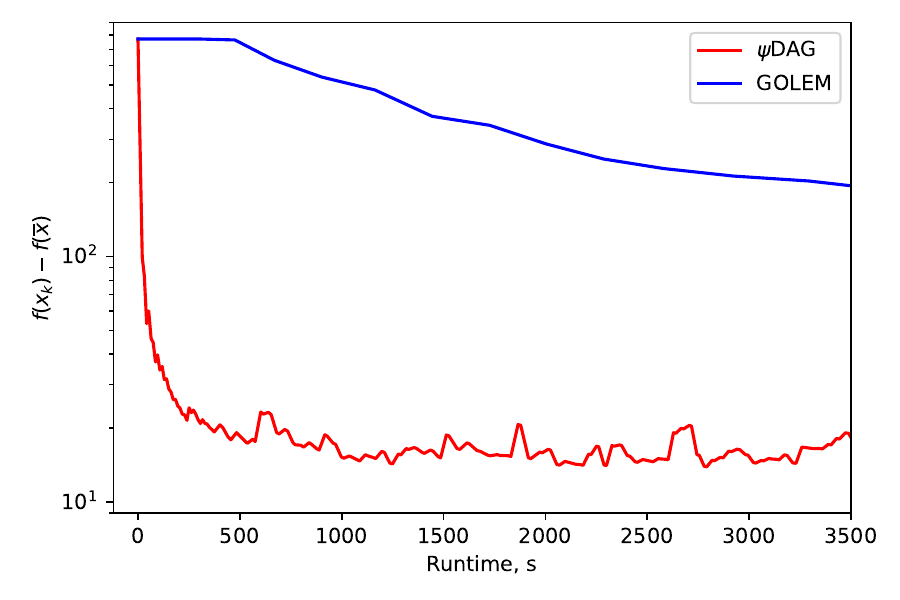}
        \includegraphics[width=\thirdwidth]{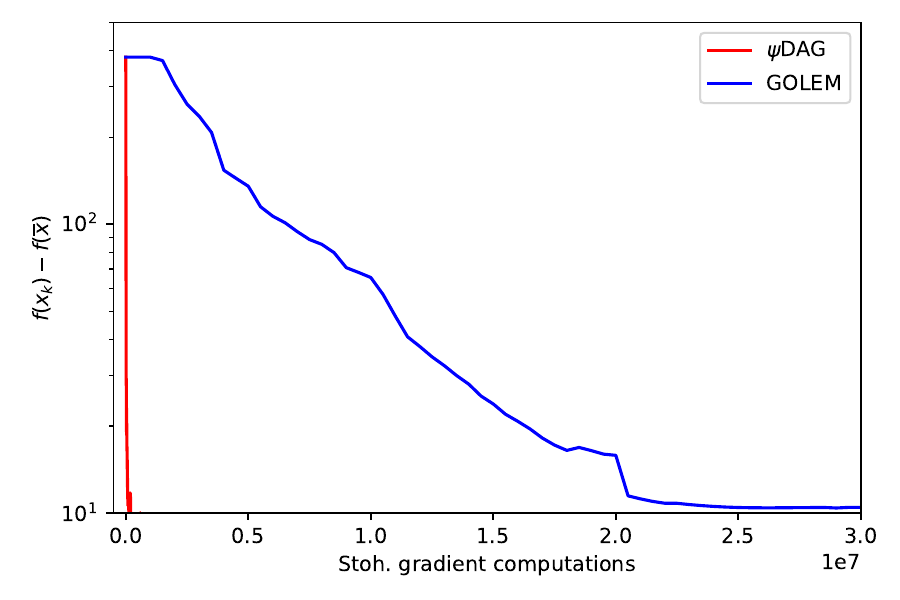}
        \includegraphics[width=\thirdwidth]{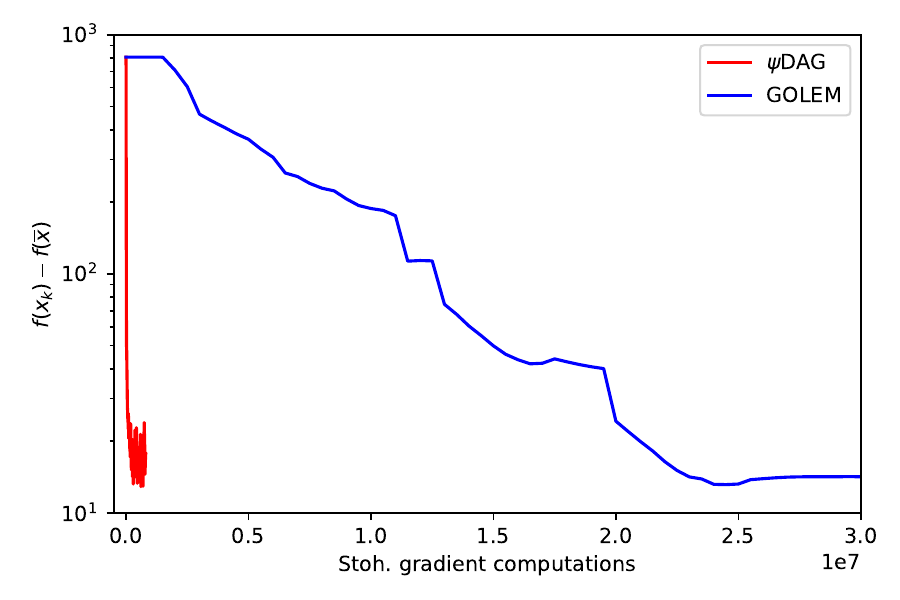}
        \includegraphics[width=\thirdwidth]{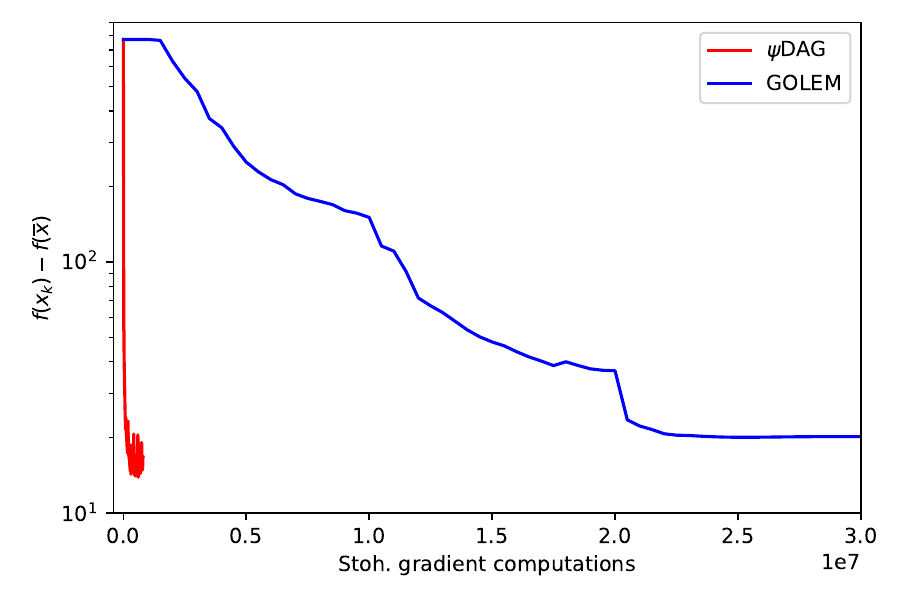}
        \caption{$d=500$ vertices}
    \end{subfigure}

    \begin{subfigure}{\textwidth}
        \centering
        \includegraphics[width=\thirdwidth]{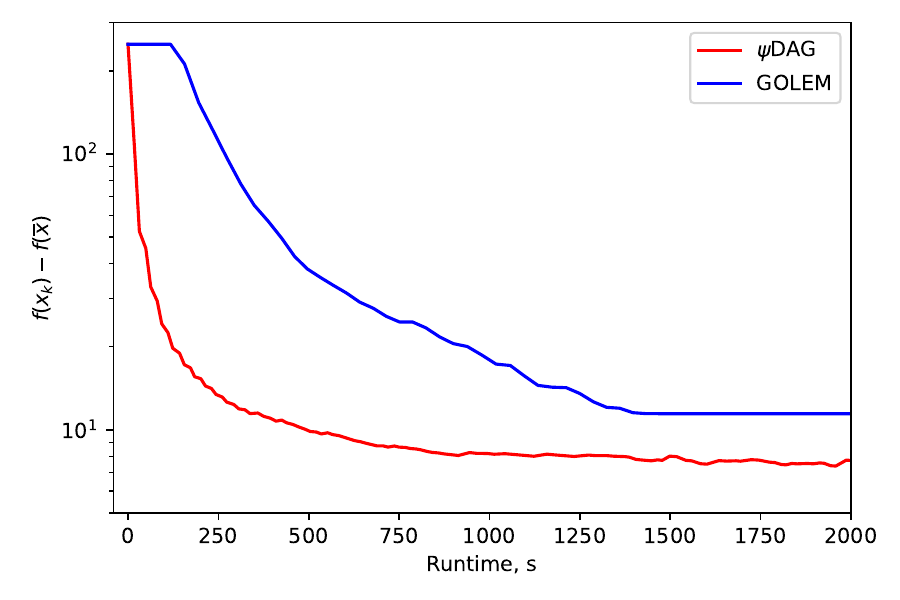}
        \includegraphics[width=\thirdwidth]{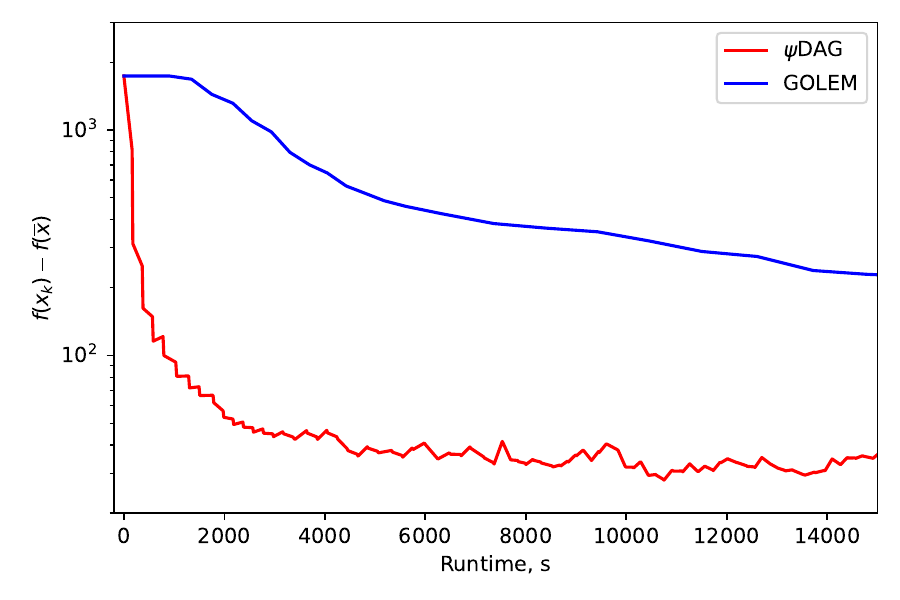}
        \includegraphics[width=\thirdwidth]{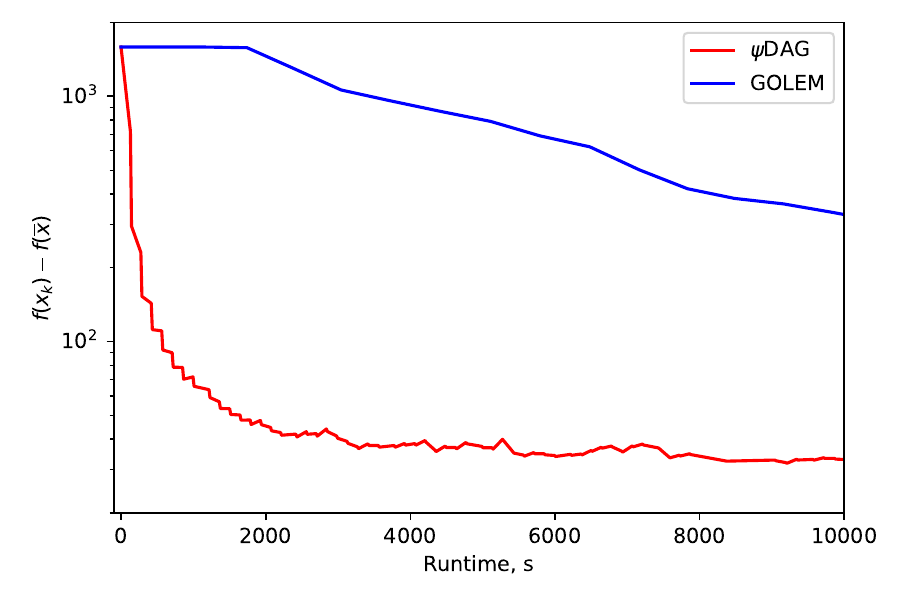}
        \includegraphics[width=\thirdwidth]{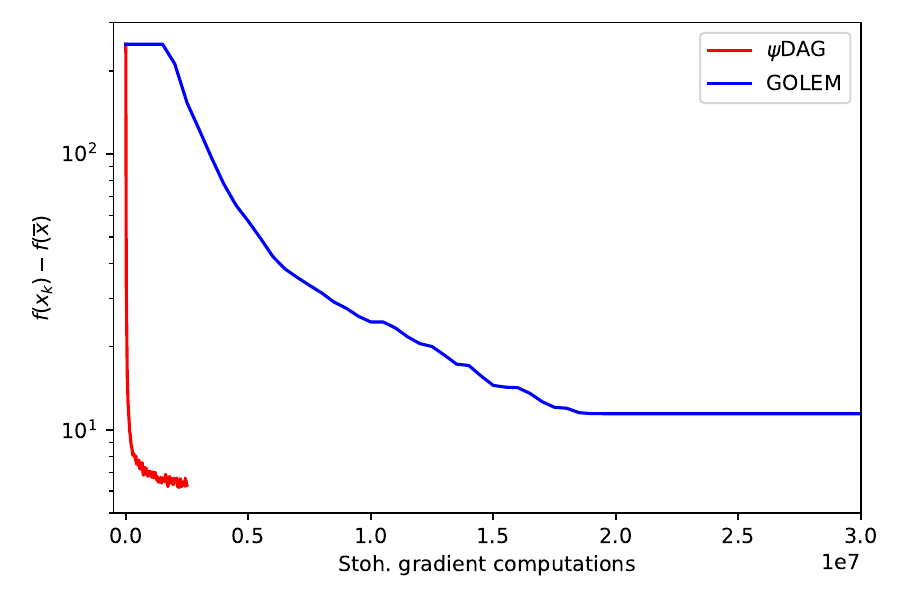}
        \includegraphics[width=\thirdwidth]{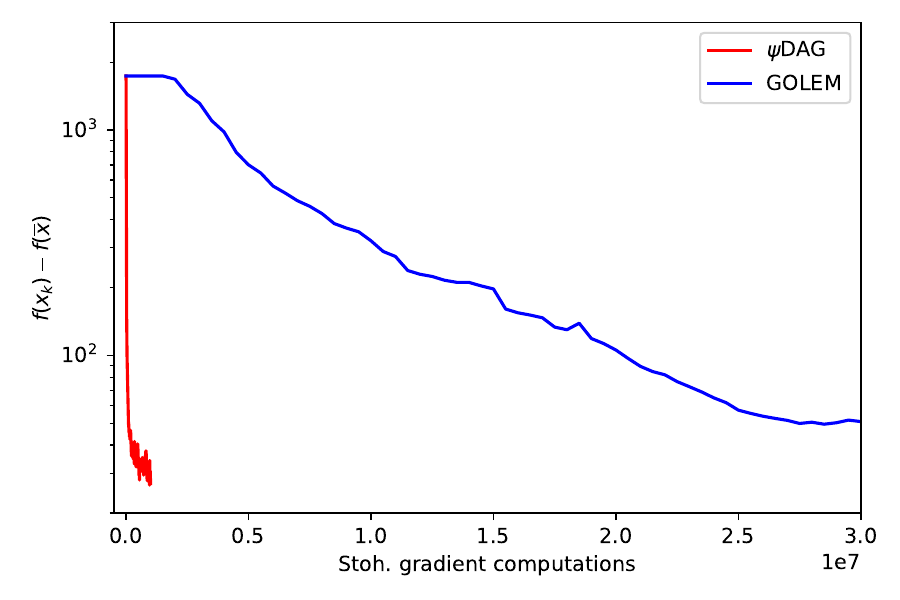}
        \includegraphics[width=\thirdwidth]{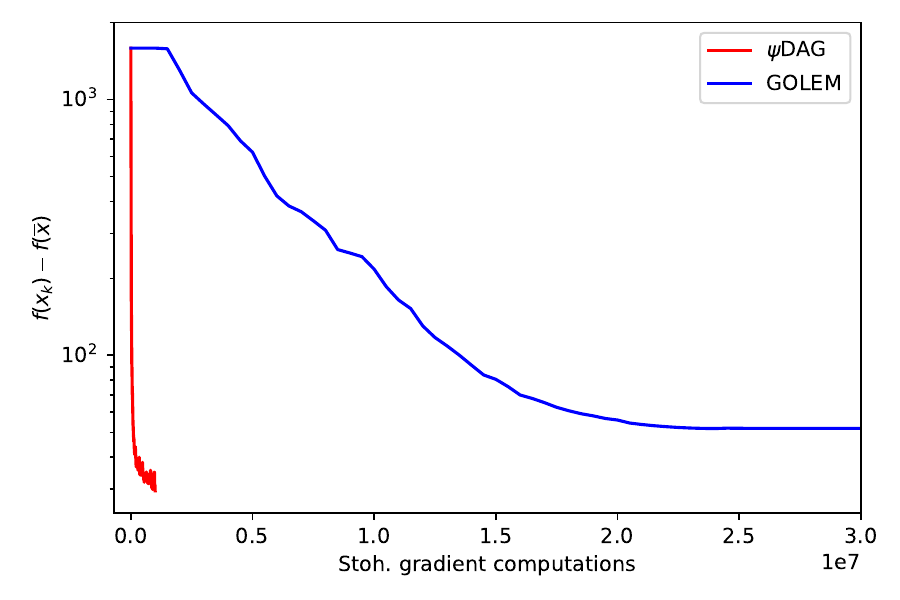}
        \caption{$d=1000$ vertices}
    \end{subfigure}
    \caption{Linear SEM methods on graphs of type SF4 with different noise distributions: Gaussian (first), exponential (second), Gumbel (third).}
    \label{fig:sf4_medium}
\end{figure*}

\textbf{SF graph types:} Figures \ref{fig:sf2_small} and \ref{fig:sf2_medium} show performance on SF2 graphs, Figures \ref{fig:sf4_small} and \ref{fig:sf4_medium} show performance on SF4 graphs,
\Cref{fig:sf6} shows performance on SF6 graphs.

We report a functional value decrease compared to \textbf{i)} time elapsed and \textbf{ii)} number of gradients computed, which also serves as a proxy of time.

\Cref{fig:1000_er2} shows that \dagma{} requires a significantly larger amount of gradient computations compared to both \ours{} and \golem{}. 

\subsection{Small to Moderate Number of Nodes}
Our experiments demonstrate that while number of nodes is small, $d< 100$, \golem{} is more stable than \dagma{}, and \ours{} method is the most stable.
While \dagma{} shows impressive speed for smaller node sets, the number of iterations required is still higher than both \golem{}  and our method. 
Across all scenarios, \ours{} consistently demonstrates faster convergence compared to the other approaches, requiring fewer iterations to reach the desired solution.

\subsection{Large Number of Nodes }

\begin{figure*}[ht]
    \centering
    \begin{subfigure}{0.45\textwidth}
        \includegraphics[width=\textwidth]{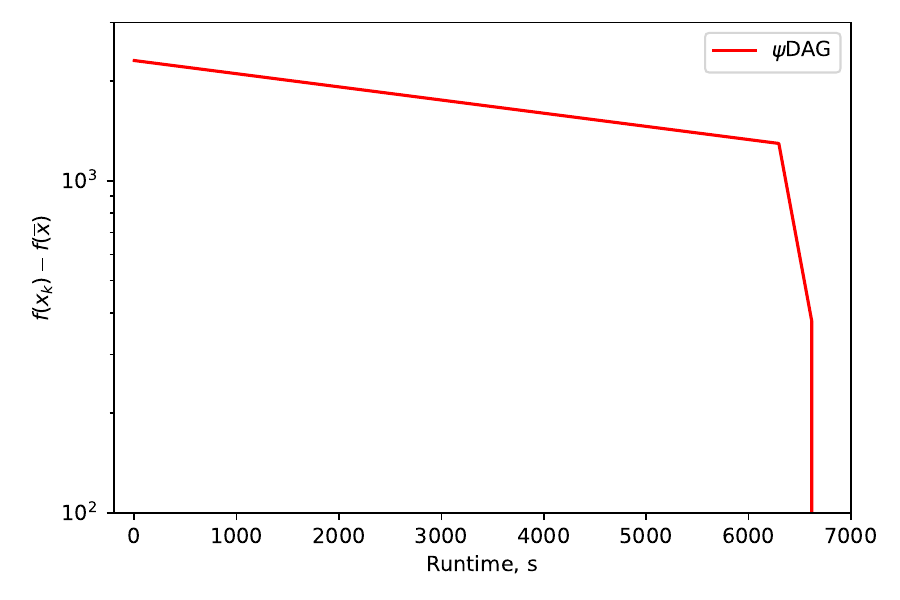}
    \end{subfigure}
    \begin{subfigure}{0.45\textwidth}
        \includegraphics[width=\textwidth]{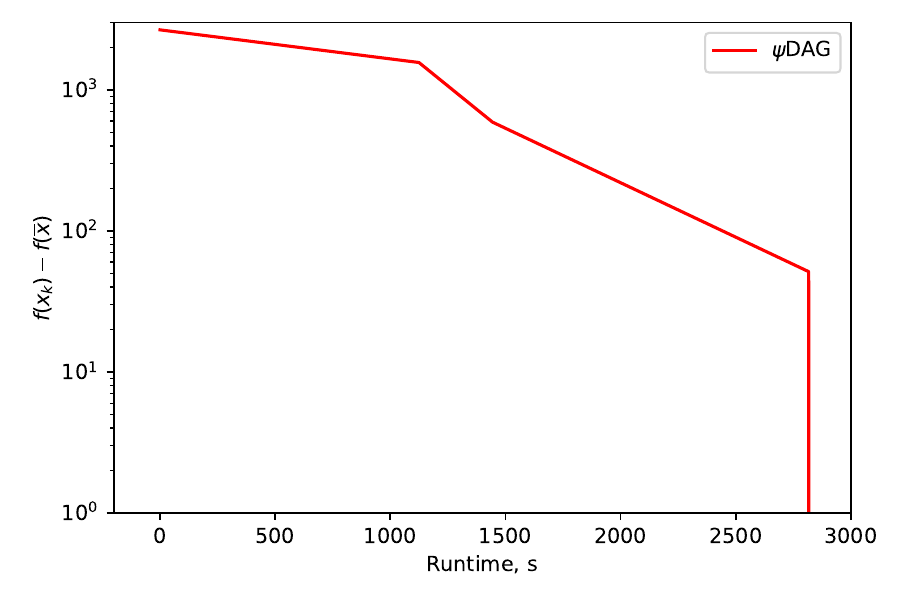}
    \end{subfigure}
    
    \begin{subfigure}{0.45\textwidth}
        \includegraphics[width=\textwidth]{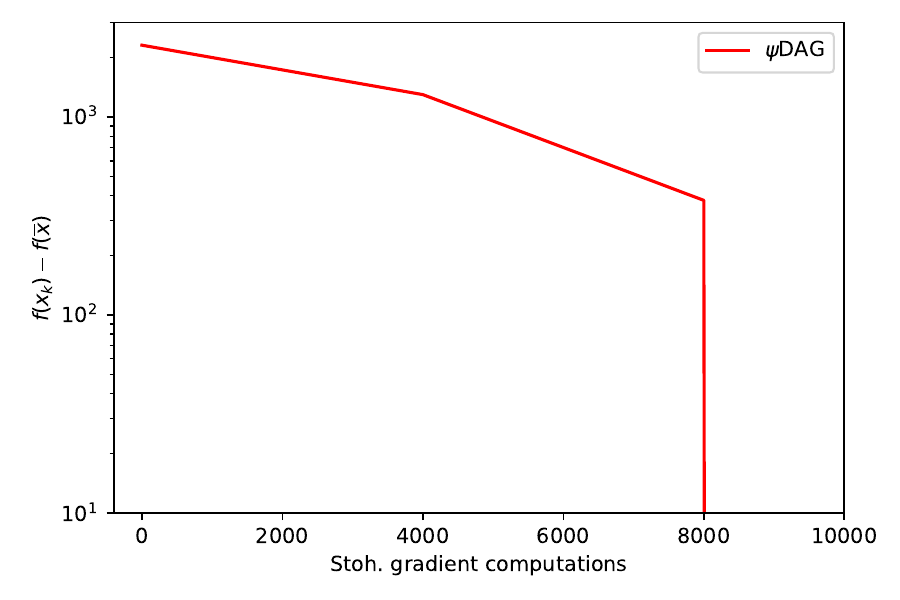}
        \caption{\texttt{ER2}}
        \label{fig:10000_ER2_gaus}
    \end{subfigure}
    \begin{subfigure}{0.45\textwidth}
        \includegraphics[width=\textwidth]{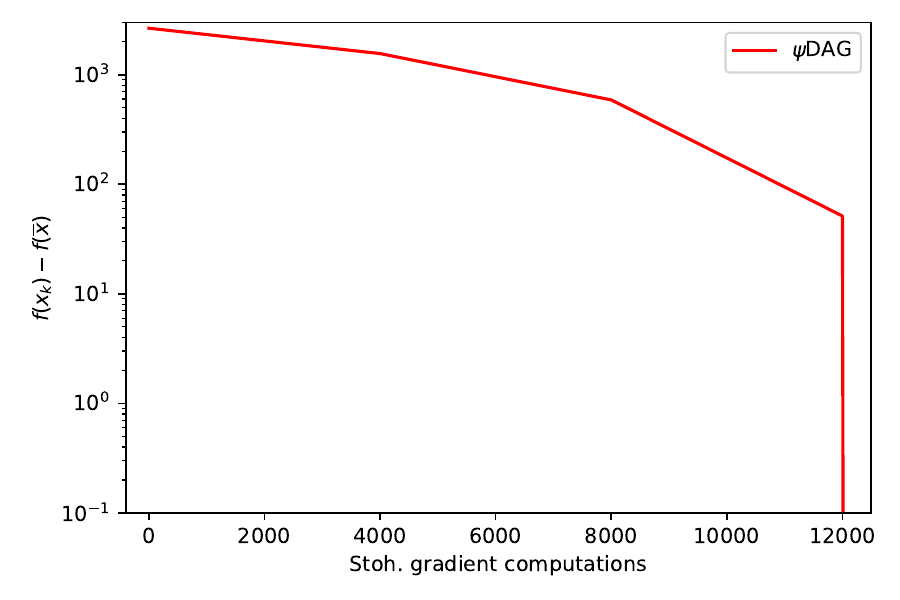}
        \caption{\texttt{SF2}}
        \label{fig:10000_SF2_gaus}
    \end{subfigure}
     \begin{subfigure}{0.45\textwidth}
        \includegraphics[width=\textwidth]{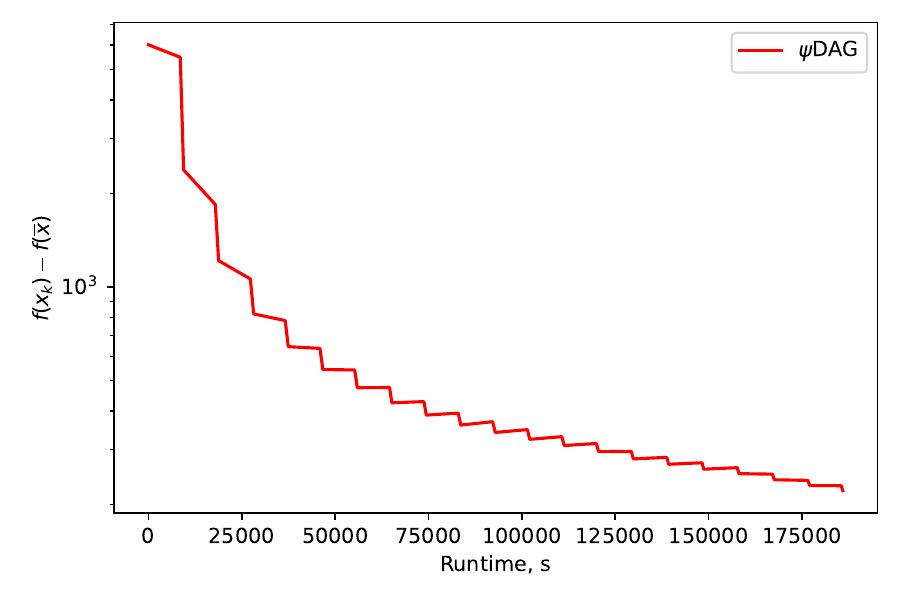}
    \end{subfigure}
    \begin{subfigure}{0.45\textwidth}
        \includegraphics[width=\textwidth]{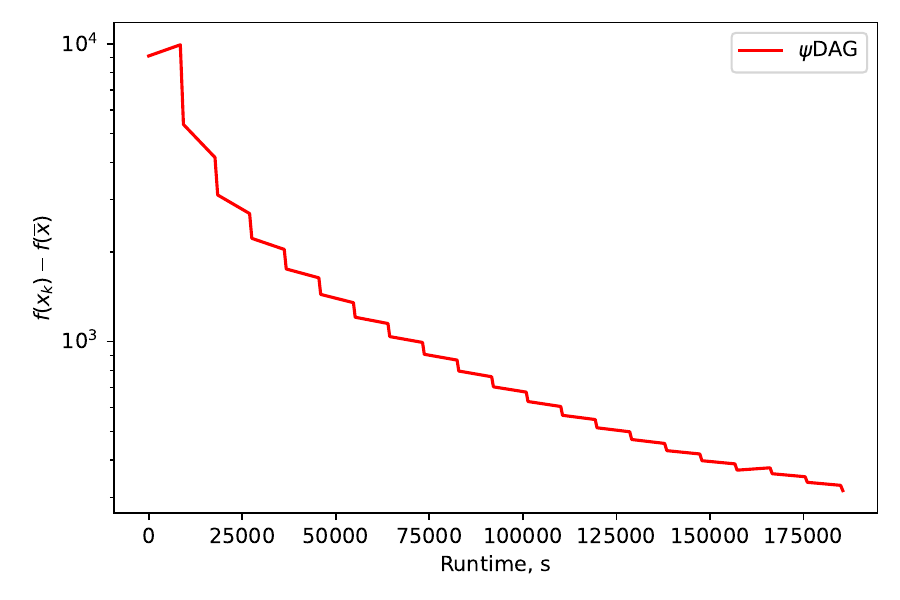}
    \end{subfigure}
    
      \begin{subfigure}{0.45\textwidth}
        \includegraphics[width=\textwidth]{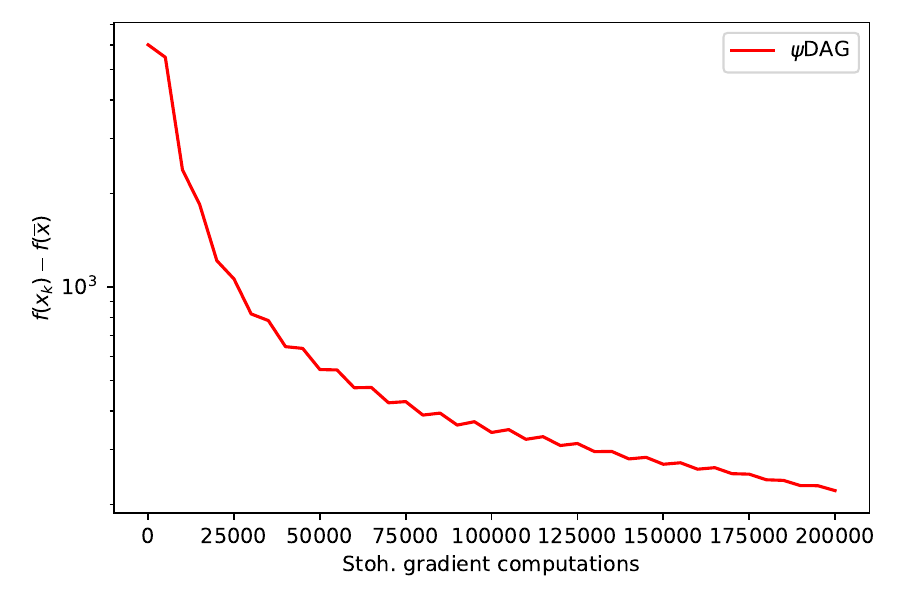}
        \caption{\texttt{ER4}}
        \label{10000_ER4_gaus}
    \end{subfigure}
    \begin{subfigure}{0.45\textwidth}
        \includegraphics[width=\textwidth]{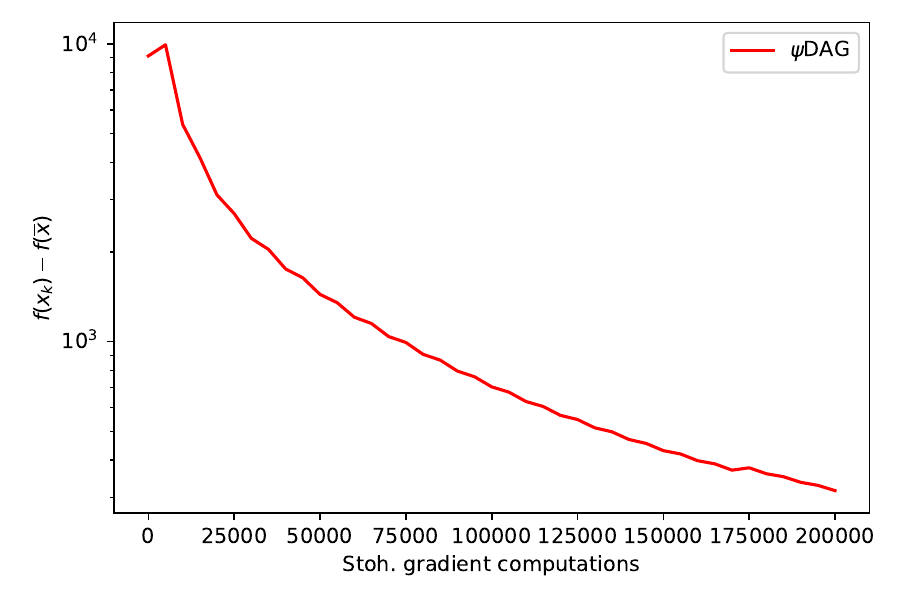}
        \caption{\texttt{SF4}}
        \label{fig:10000_SF4_gaus}
    \end{subfigure}
    
    \caption{\ours{} method for graph types ER2, ER4, SF2 and SF4 graphs with \(d =10000\) and Gaussian noise. Other linear SEM methods do not converge in less than $350$ hours.}
    \label{fig:4_10000_d_gauss}
\end{figure*}

For graphs with a large number of nodes \(d \in \{5000, 10000\}\), we were unable to run neither of the baselines, and consequently, \Cref{fig:4_10000_d_gauss} includes only one algorithm. \golem{} was not feasible due to its computation time exceeding $350$ hours. \dagma{} was impossible as its runs led to kernel crashes. In all cases, we utilized a training set of 5,000 samples and a validation set of 10,000 samples.

\subsection{Denser Graphs}
\begin{figure*}
    \centering
    \begin{subfigure}{\threesubfigwidth}
        \centering
        \includegraphics[width=\textwidth]{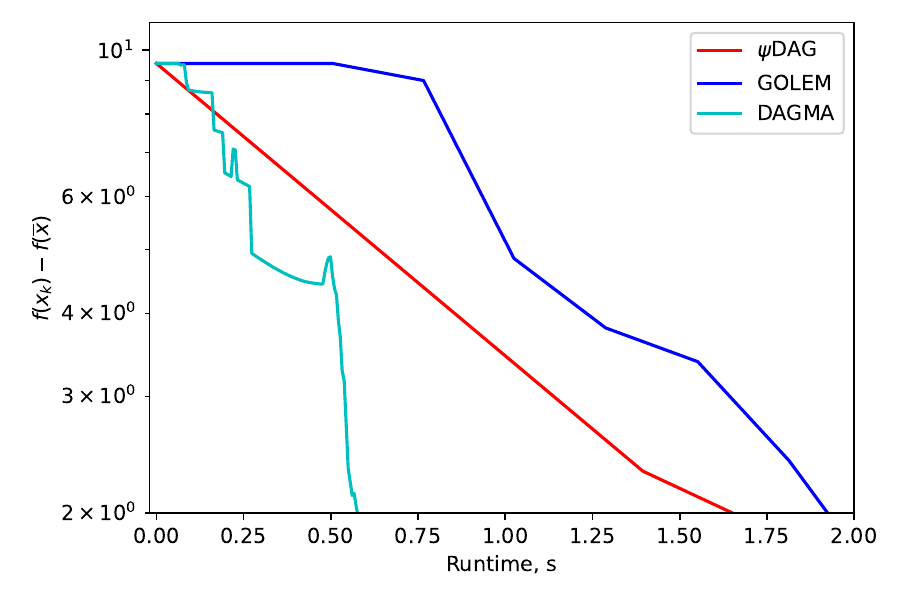}
        \\
        \includegraphics[width=\textwidth]{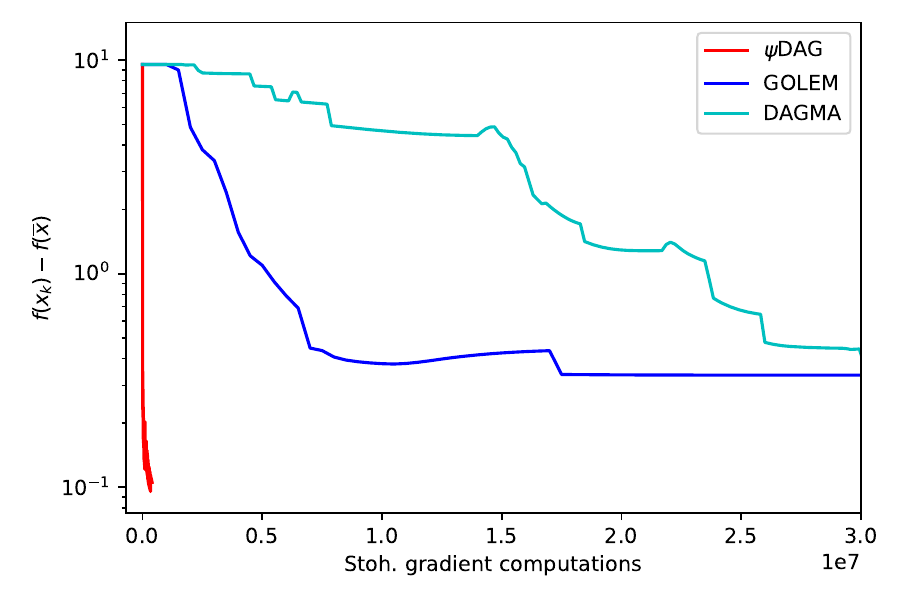}
        \caption{$d=10$ vertices}
    \end{subfigure}
    \begin{subfigure}{\threesubfigwidth}
        \centering
        \includegraphics[width=\textwidth]{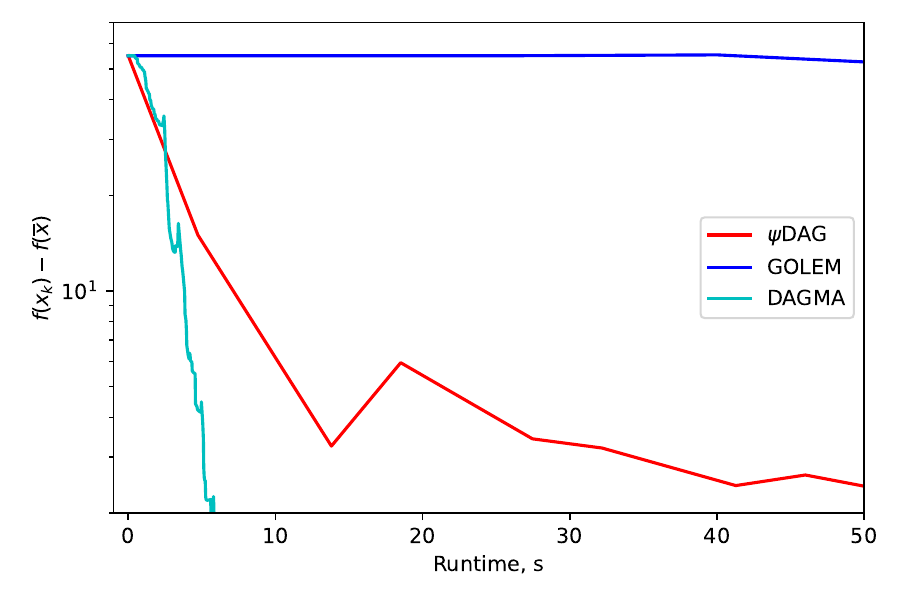}
        \\
        \includegraphics[width=\textwidth]{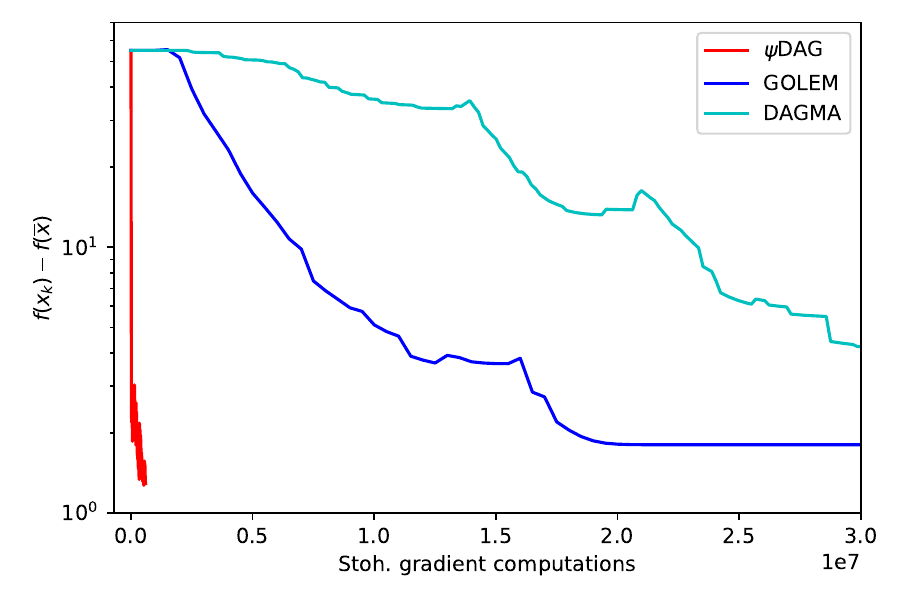}
        \caption{$d=50$ vertices}
    \end{subfigure}
    \begin{subfigure}{\threesubfigwidth}
        \centering
        \includegraphics[width=\textwidth]{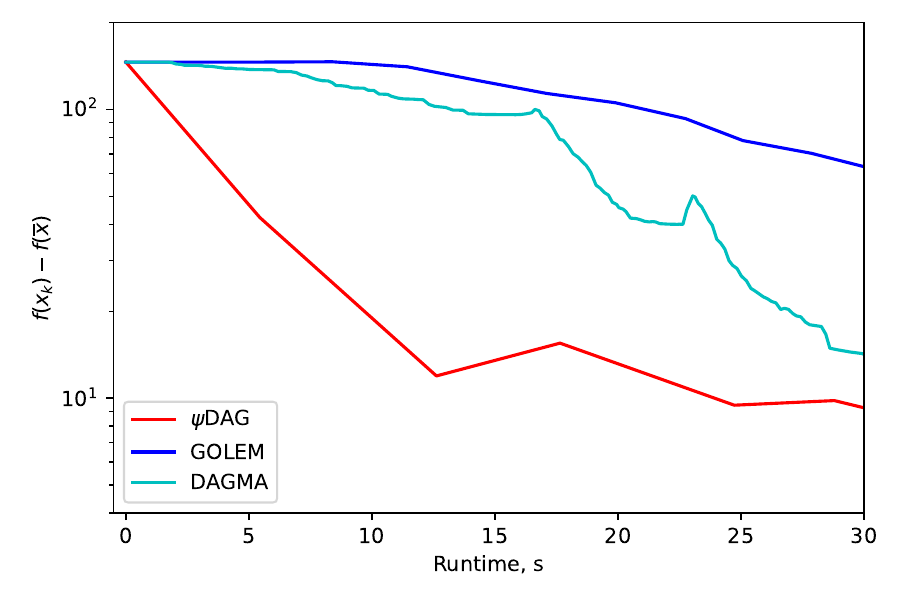}
        \\
        \includegraphics[width=\textwidth]{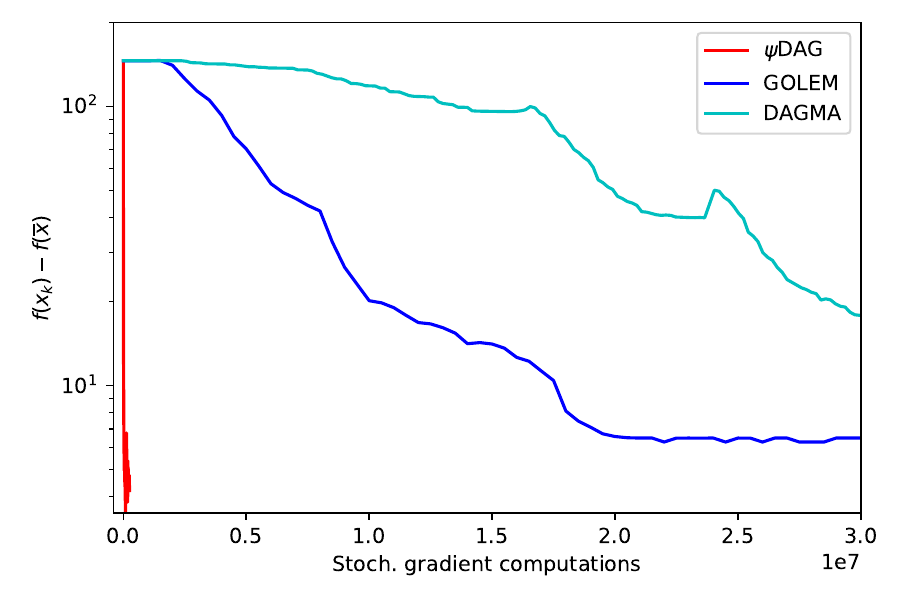}
        \caption{$d=100$ vertices}
    \end{subfigure}
    \begin{subfigure}{\threesubfigwidth}
        \centering
        \includegraphics[width=\textwidth]{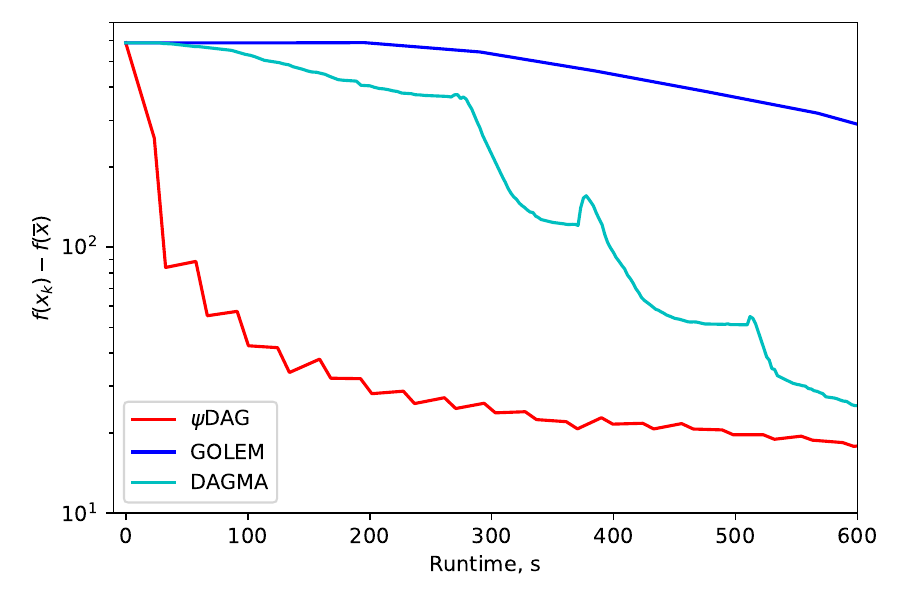}
        \\
        \includegraphics[width=\textwidth]{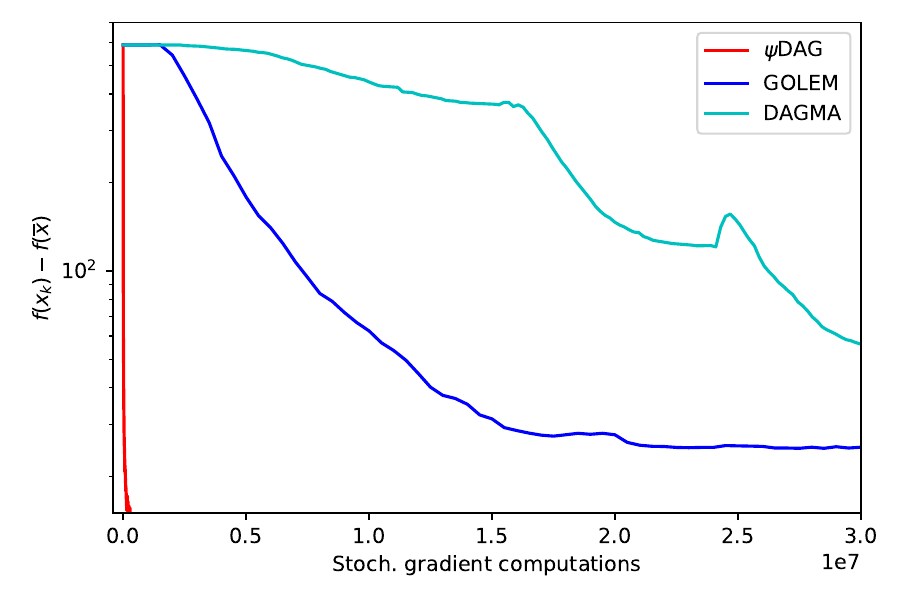}
        \caption{$d=500$ vertices}
    \end{subfigure}
    \begin{subfigure}{\threesubfigwidth}
        \centering
        \includegraphics[width=\textwidth]{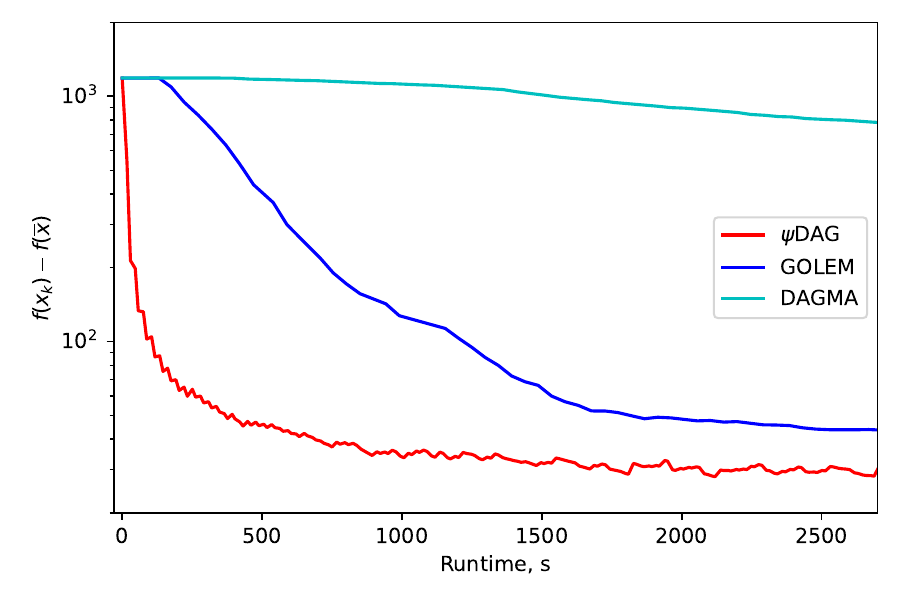}
        \\
        \includegraphics[width=\textwidth]{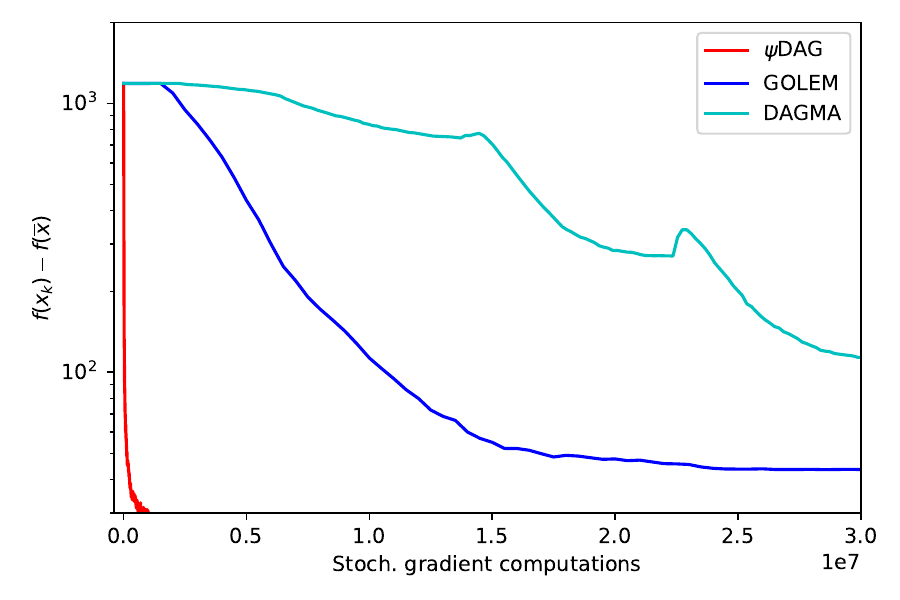}
        \caption{$d=1000$ vertices}
    \end{subfigure}
    \begin{subfigure}{\threesubfigwidth}
        \centering
        \includegraphics[width=\textwidth]{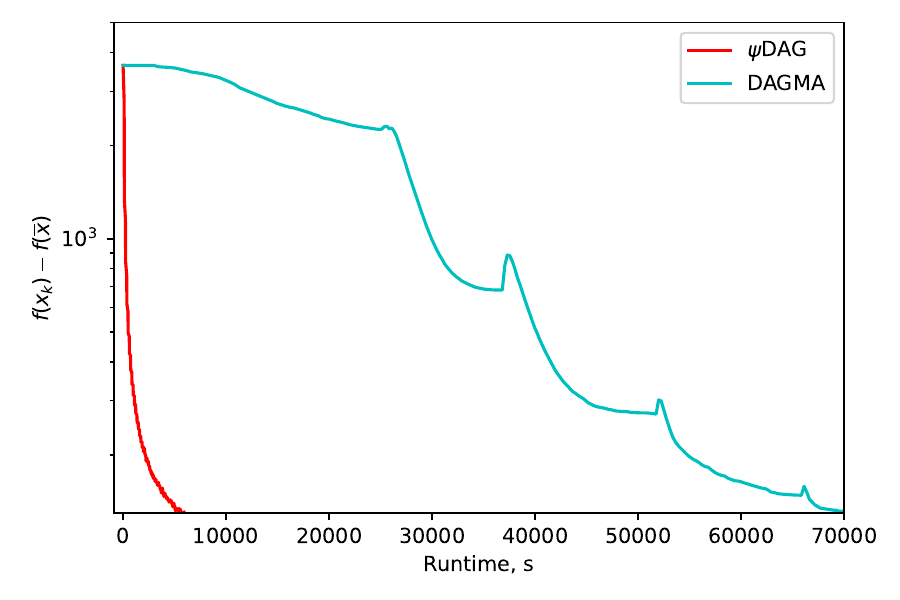}
        \\
        \includegraphics[width=\textwidth]{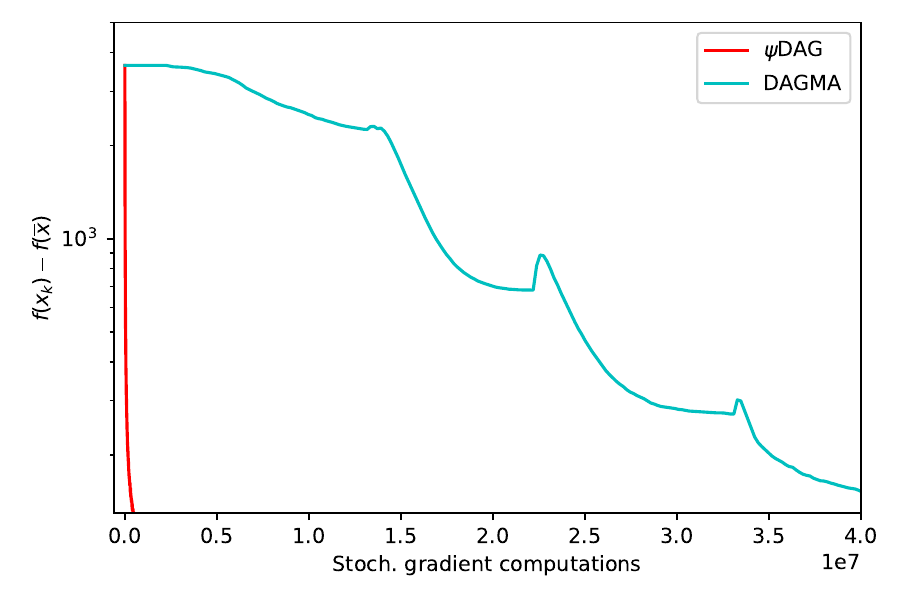}
        \caption{$d=3000$ vertices}
    \end{subfigure}

    \caption{Linear SEM methods on graphs of type ER6 with the Gaussian noise distribution.}
    \label{fig:er6}
\end{figure*}

\begin{figure*}
    \centering
    \begin{subfigure}{\threesubfigwidth}
        \centering
        \includegraphics[width=\textwidth]{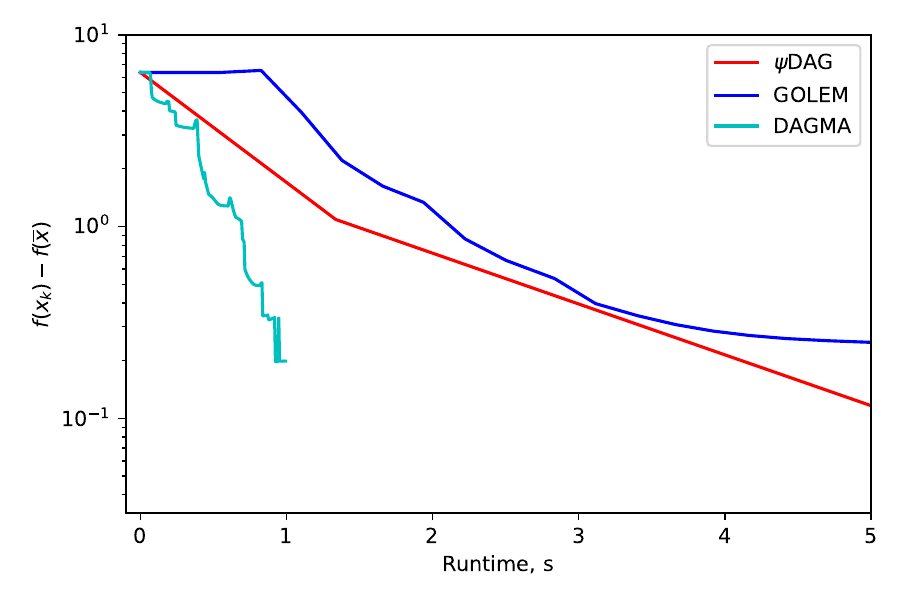}
        \\
        \includegraphics[width=\textwidth]{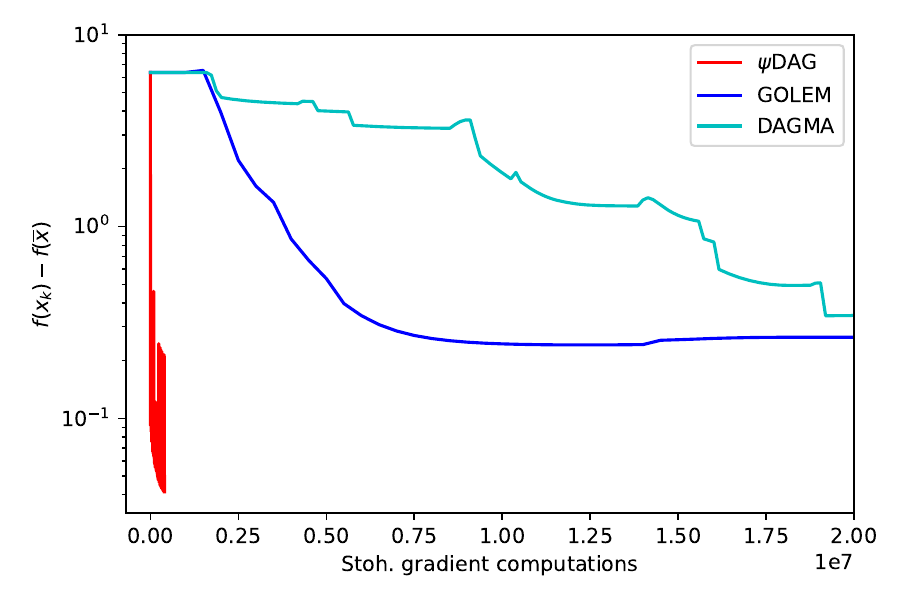}
        \caption{$d=10$ vertices}
    \end{subfigure}
    \begin{subfigure}{\threesubfigwidth}
        \centering
        \includegraphics[width=\textwidth]{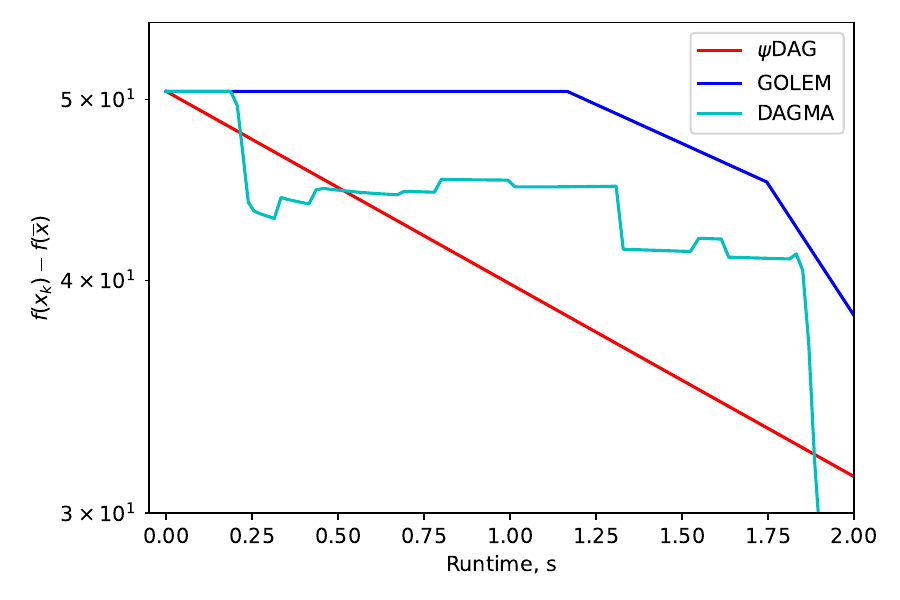}
        \\
        \includegraphics[width=\textwidth]{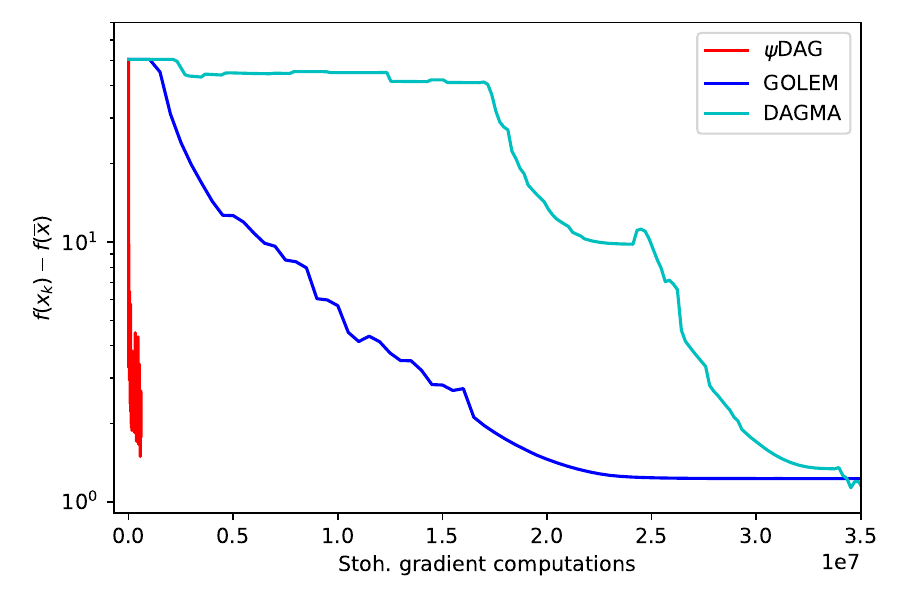}
        \caption{$d=50$ vertices}
    \end{subfigure}
    \begin{subfigure}{\threesubfigwidth}
        \centering
        \includegraphics[width=\textwidth]{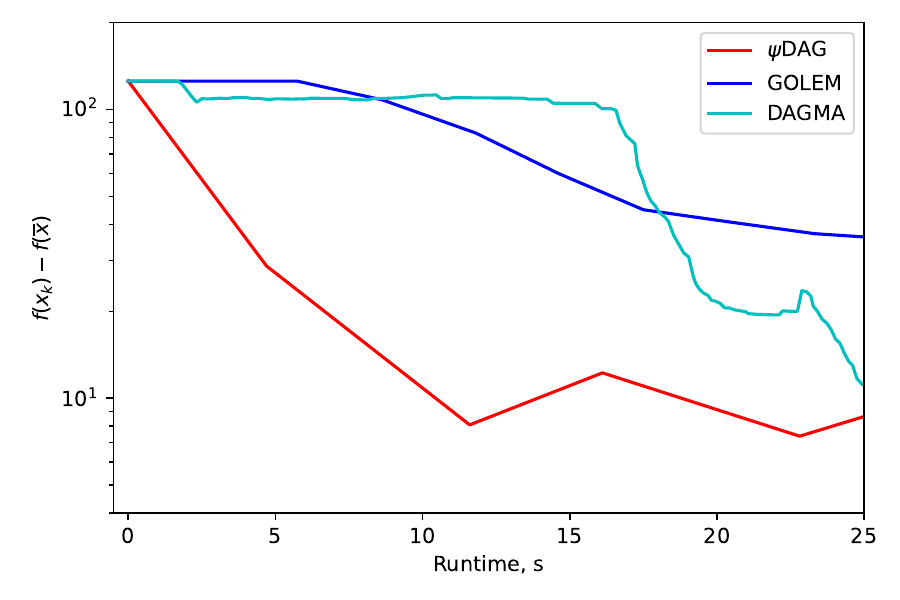}
        \\
        \includegraphics[width=\textwidth]{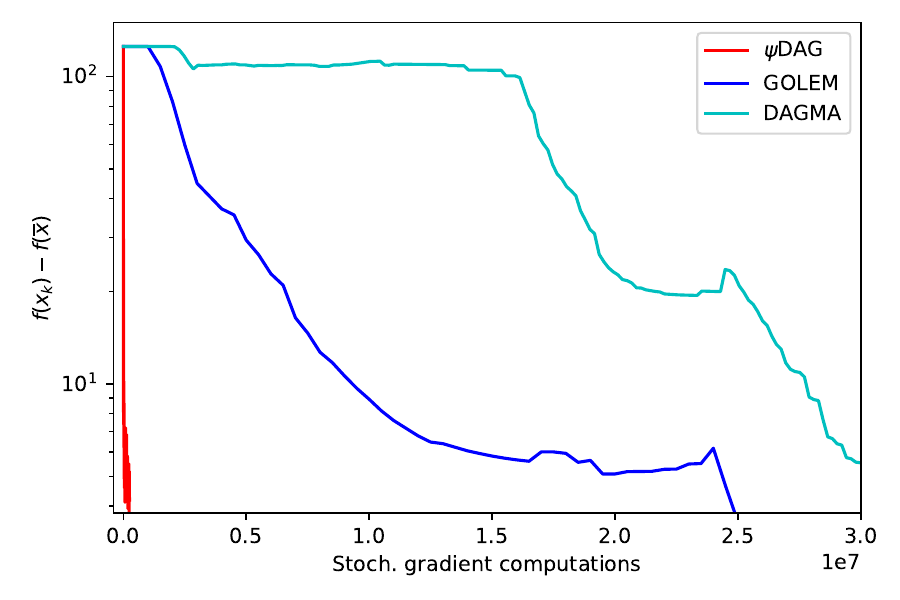}
        \caption{$d=100$ vertices}
    \end{subfigure}
    \begin{subfigure}{\threesubfigwidth}
        \centering
        \includegraphics[width=\textwidth]{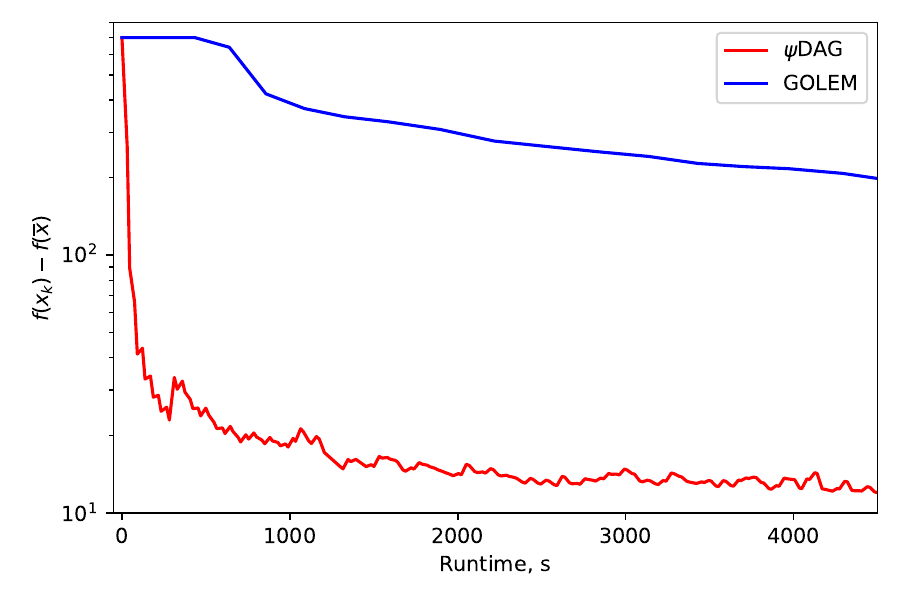}
        \\
        \includegraphics[width=\textwidth]{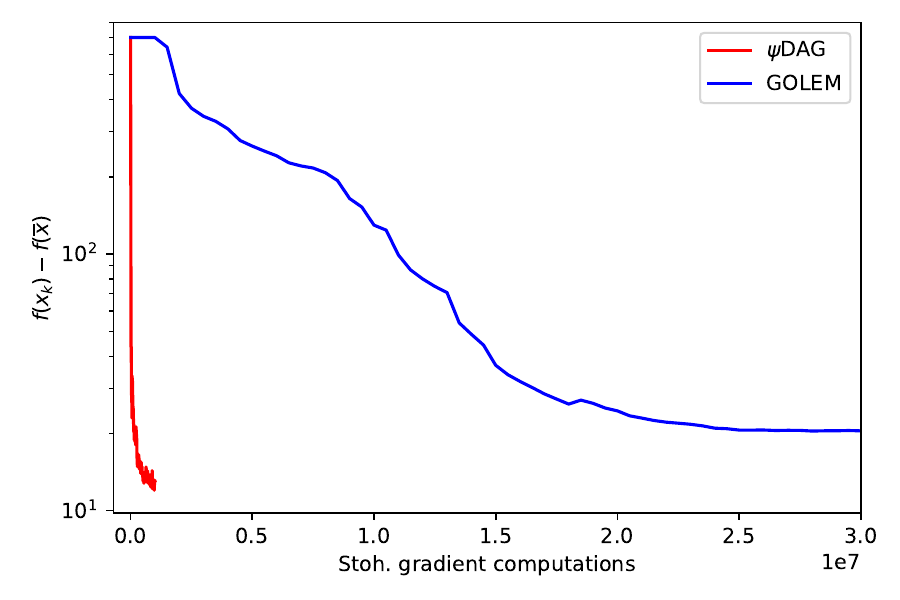}
        \caption{$d=500$ vertices}
    \end{subfigure}
    \begin{subfigure}{\threesubfigwidth}
        \centering
        \includegraphics[width=\textwidth]{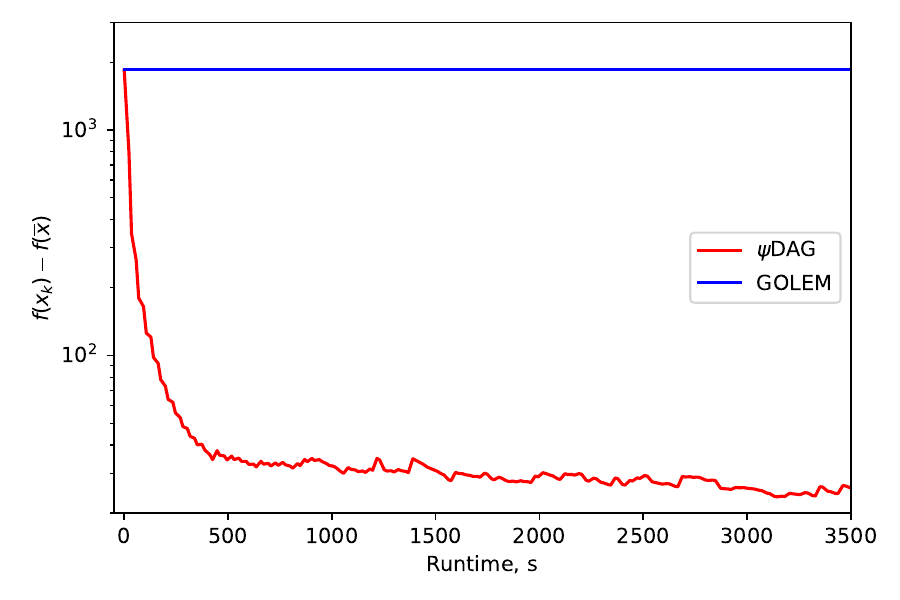}
        \includegraphics[width=\textwidth]{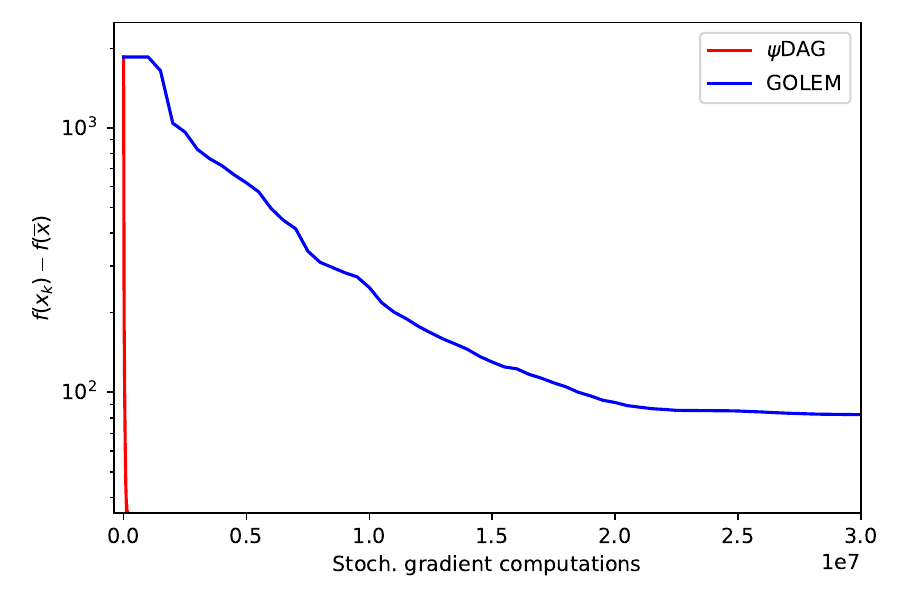}
        \caption{$d=1000$ vertices}
    \end{subfigure}
    \begin{subfigure}{\threesubfigwidth}
        \centering
        \includegraphics[width=\textwidth]{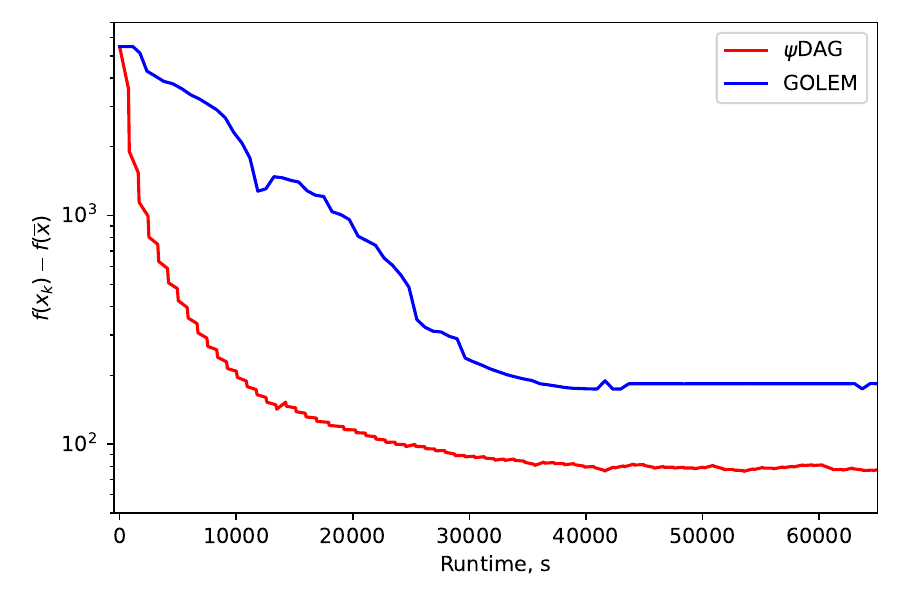}
        \\
        \includegraphics[width=\textwidth]{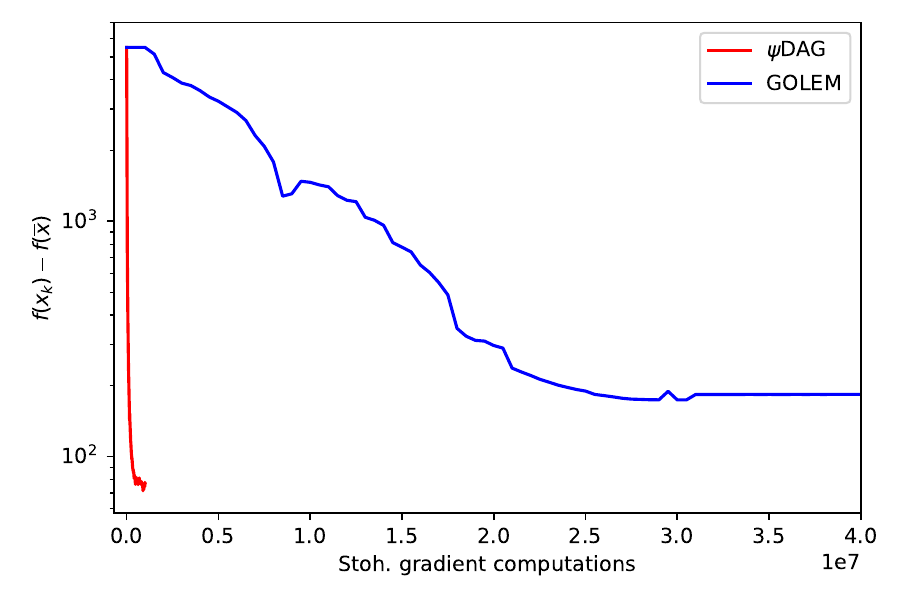}
        \caption{$d=3000$ vertices}
    \end{subfigure}

    \caption{Linear SEM methods on graphs of type SF6 with the Gaussian noise distribution.}
    \label{fig:sf6}
\end{figure*}

For a thorough comparison, in Figures \ref{fig:er6} and \ref{fig:sf6}, we compare graph structures ER6 and SF6 under the Gaussian noise type. 
Plots indicate that while \dagma{} exhibits a fast runtime when the number of nodes is small, $d<100$, it requires more iterations to achieve convergence. Algorithm \ours{} consistently outperforms \golem{} and \dagma{} in both training time and a number of stochastic gradient computations, and the difference is more pronounced for lager number of nodes and denser graphs.

\clearpage
\begin{figure}
  \centering
  \includegraphics[width=0.45\linewidth]{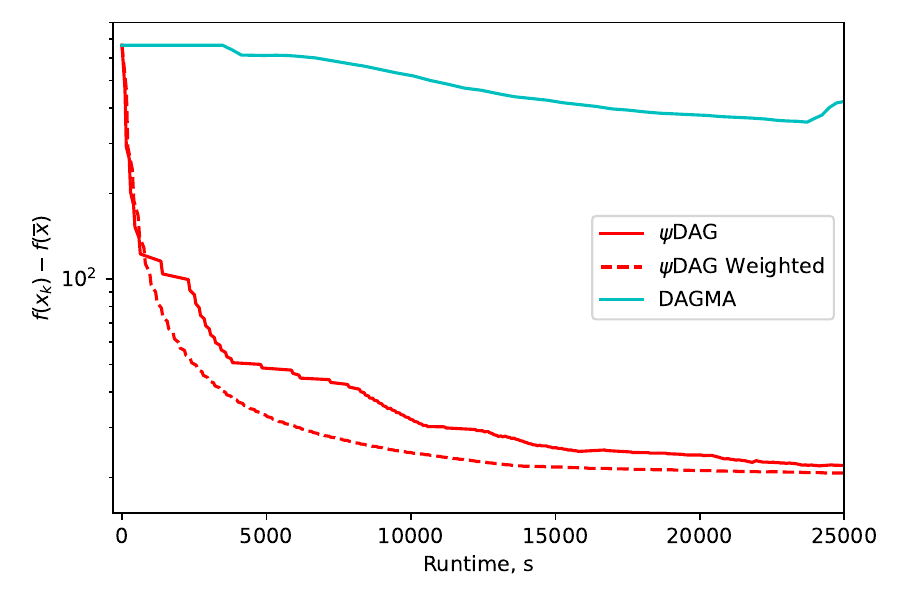}
  \includegraphics[width=0.45\linewidth]{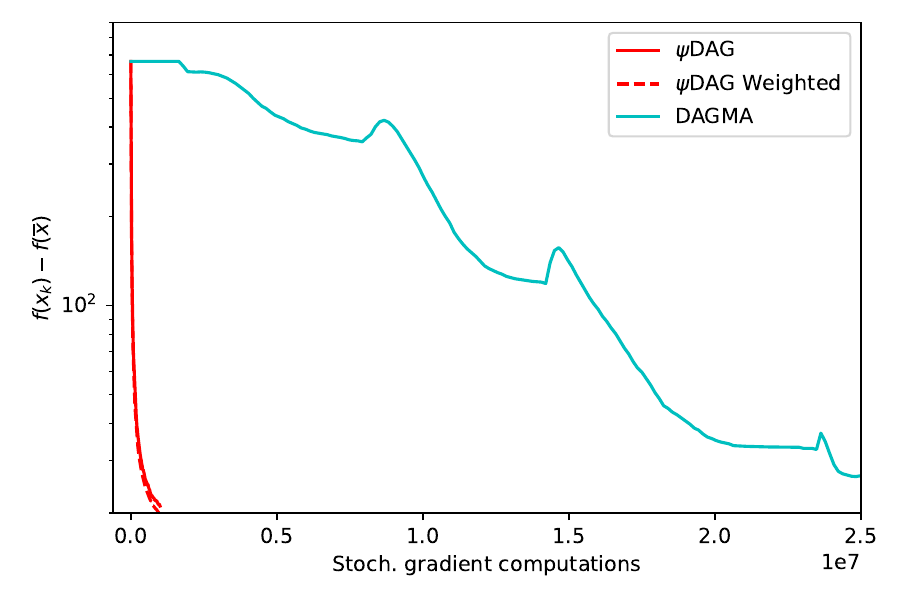}
  \caption{Comparison of \ours{}, \ours{} weighted and \dagma{} for ER2 graph with $d = 3000$ nodes and Gaussian noise.}
  \label{fig:d=3000_ER2_5000_gaussian_ev_w}
\end{figure}
\section{Weighted Projection} \label{sec:importance_weights}
Inspired by the importance of sampling, we considered adjustment of the projection method by weights. Specifically, we considered the elements of the $\model{}$ to be weighted element-wisely by the second directional derivatives of the objective function, $\weights[i][j]\eqdef\left( \frac d{d\model[i][j]} \right)^2 \Edist{X \sim \cD}{l(\model;X)} $.
As we don't have access to the whole distribution $\cD$, we approximate it by the mean of already seen samples, 
\begin{equation}\label{eq:weights}
    \weights_k[i][j]\eqdef\left( \frac d{d\model[i][j]} \right)^2 \frac 1k \sum_{k=0}^{k-1} {l \left(\model;X_k \right)} = \frac 1k \sum_{t=0}^{k-1} \left( X_k[j]\right)^2.
\end{equation}
Weights \eqref{eq:weights} are identical for whole columns; hence, they impose storing only one vector. Updating them requires a few element-wise vector operations.

\begin{figure*}[ht]
    \centering
     \begin{subfigure}{0.45\textwidth}
        \includegraphics[width=\textwidth]{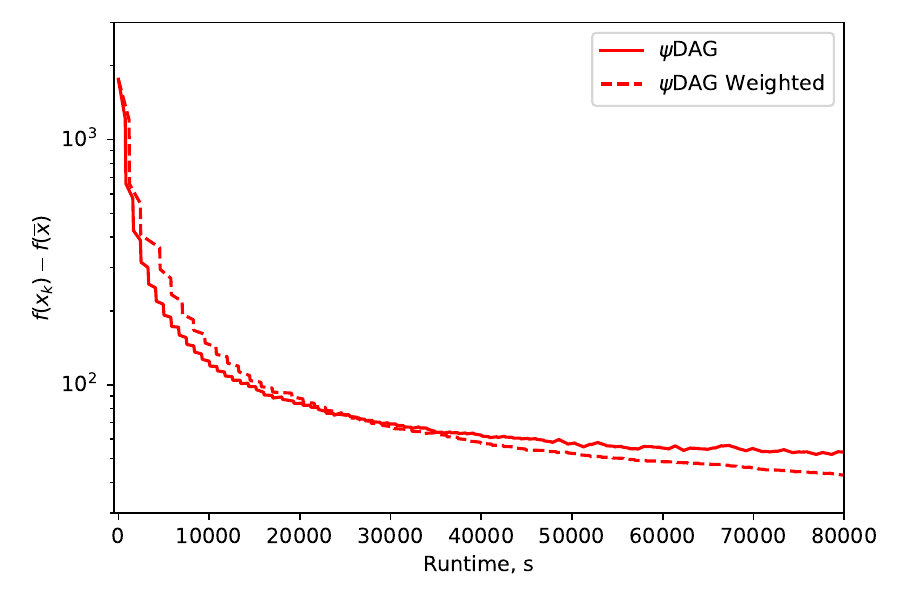}
    \end{subfigure}
    \begin{subfigure}{0.45\textwidth}
        \includegraphics[width=\textwidth]{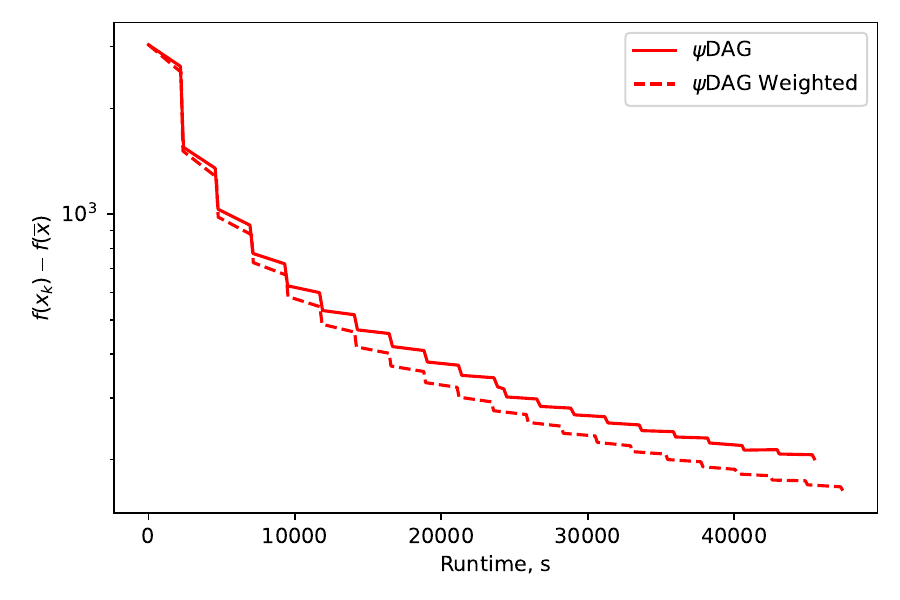}
    \end{subfigure}
    
      \begin{subfigure}{0.45\textwidth}
        \includegraphics[width=\textwidth]{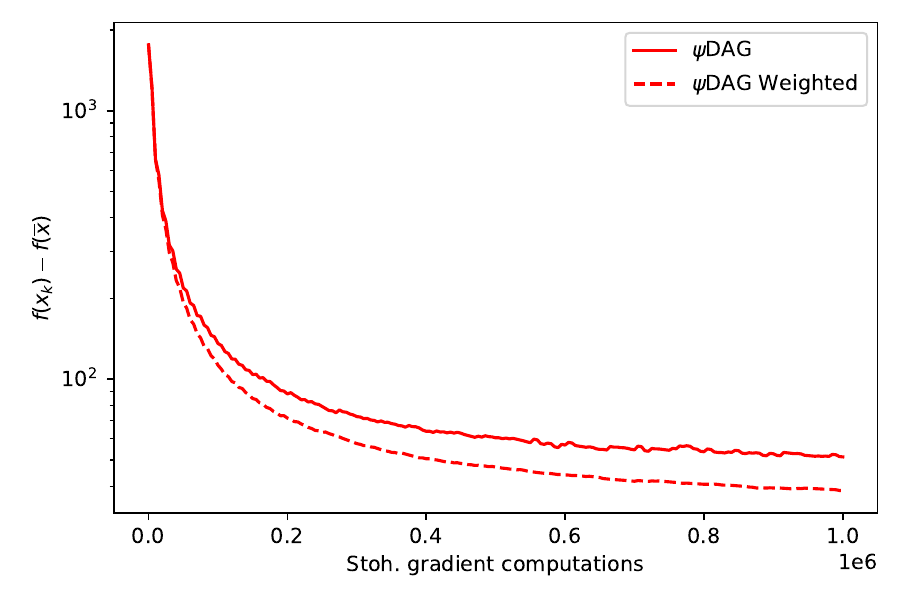}
        \caption{$d=3000$ vertices}
        \label{3000_ER4_gaus}
    \end{subfigure}
    \begin{subfigure}{0.45\textwidth}
        \includegraphics[width=\textwidth]{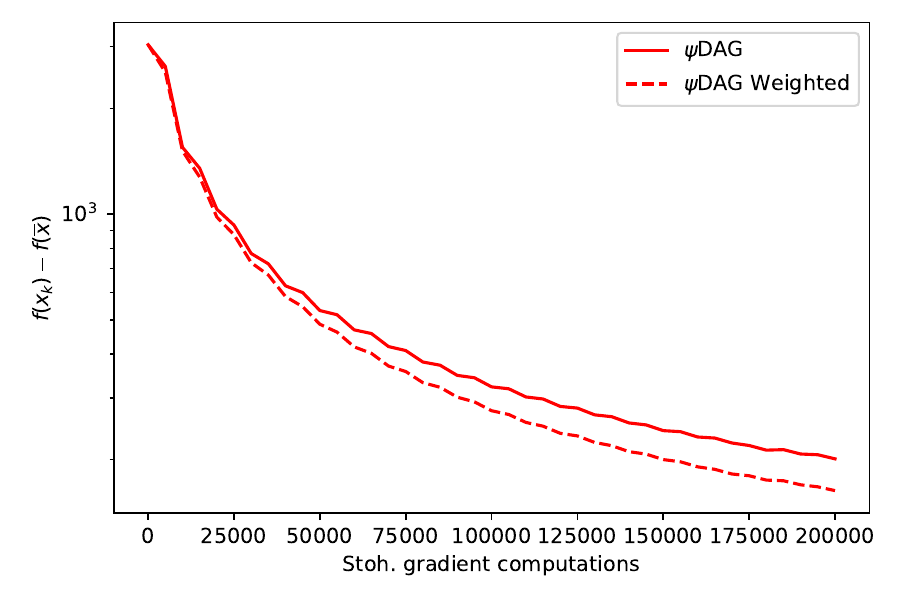}
        \caption{$d=5000$ vertices}
        \label{fig:5000_ER4_gaus}
    \end{subfigure}
    
    \caption{\ours{} method for graph types ER4 with \(d =\{3000, 5000\}\) respectively and Gaussian noise.}
    \label{fig:4_3000_w_gauss}
\end{figure*}

Figures~\ref{fig:d=3000_ER2_5000_gaussian_ev_w} and \ref{fig:4_3000_w_gauss} show that this weighting can lead to an improved convergence (slightly faster convergence to a slightly lower functional value) without imposing extra cost. However, this improvement over runtime is not consistent; hence, for simplicity, we deferred this to the appendix.

\end{document}